\documentclass[a4paper, 11pt,hidelinks,oneside,svgnames]{report}
\usepackage[pdftex]{graphicx}
\usepackage{longtable}
\usepackage{wrapfig}
\usepackage{rotating}
\usepackage[normalem]{ulem}
\usepackage{amsmath}
\usepackage{amssymb}
\usepackage{capt-of}
\usepackage{hyperref}
\usepackage{bbm, stmaryrd}
\usepackage{tikz}
\usepackage{pgfplots}
\usetikzlibrary{positioning, decorations.pathreplacing, backgrounds, decorations.pathmorphing}
\usetikzlibrary{matrix, plotmarks, shapes.geometric, calc}

\usepackage{standalone}
\usepackage{suthesis-2e}                   % modified Stanford thesis layout
\usepackage{geometry}
\usepackage{amsmath, amsfonts, amssymb}    % a bunch of math environments and commands
\usepackage{booktabs}                      % pretty tables
\usepackage{xcolor}                        % reasonable color names
\usepackage{graphicx}                      % for figures
\usepackage{pdfpages}                      % for the declaration of originality
\usepackage{subcaption}                    % multiple figures side-by-side
\usepackage{mdframed}                      % put a nice frame around listings
\usepackage{enumitem}                      % use a), b), c) and so on
\usepackage[capitalise]{cleveref}          % reasonable referencing

\usepackage[backend=biber, style=authoryear, doi=false, isbn=false, url=false, date=year, giveninits=true, maxbibnames=99, uniquelist=false, maxcitenames=2, dashed=false]{biblatex}
\DeclareNameAlias{sortname}{family-given}    % undo the weird default by biblatex
\defbibenvironment{bibliography} {\begin{enumerate}} {\end{enumerate}} {\item}  % number citations

\def\cind{\perp\!\!\!\perp}  % conditional independence
\def\dis{\ \tilde{\cind}\ }  % disentanglement

\author{Romeo Valentin}
\date{\today}
\title{Towards a Framework for Deep Learning Certification in Safety-Critical Applications Using Inherently Safe Design and Run-time Error Detection}
\hypersetup{
 pdfauthor={Romeo Valentin},
 pdftitle={Towards a Framework for Deep Learning Certification in Safety-Critical Applications Using Inherently Safe Design and Run-time Error Detection},
 pdfkeywords={},
 pdfsubject={},
 pdfcreator={Emacs 28.1.50 (Org mode 9.6)}, 
 pdflang={English}}

\usepackage{fvextra}

\fvset{%
  commandchars=\\\{\},
  highlightcolor=white!95!black!80!blue,
  breaklines=false,
  breaksymbol=\color{white!60!black}\tiny\ensuremath{\hookrightarrow}}

\usepackage{xcolor}

\providecolor{EfD}{HTML}{f7f7f7}
\providecolor{EFD}{HTML}{28292e}

\usepackage[xparse]{tcolorbox}
\DeclareTColorBox[]{Code}{o}%
{colback=EfD!98!EFD, colframe=EfD!95!EFD,
  fontupper=\footnotesize\setlength{\fboxsep}{0pt},
  colupper=EFD,
  IfNoValueTF={#1}%
  {boxsep=2pt, arc=2.5pt, outer arc=2.5pt,
    boxrule=0.5pt, left=2pt}%
  {boxsep=2.5pt, arc=0pt, outer arc=0pt,
    boxrule=0pt, leftrule=1.5pt, left=0.5pt},
  right=2pt, top=1pt, bottom=0.5pt,
  }

\usepackage{float}
\floatstyle{plaintop}
\newfloat{listing}{htbp}{lst}
\newcommand{\listingsname}{Listing}
\floatname{listing}{\listingsname}
\newcommand{\listoflistingsname}{List of Listings}
\providecommand{\listoflistings}{\listof{listing}{\listoflistingsname}}

\definecolor{EFD}{HTML}{000000}
\definecolor{EfD}{HTML}{ffffff}
\definecolor{EFvp}{HTML}{000000}
\definecolor{EFh}{HTML}{7f7f7f}
\definecolor{EFsc}{HTML}{228b22}
\definecolor{EFw}{HTML}{ff8e00}
\definecolor{EFe}{HTML}{ff0000}
\definecolor{EFl}{HTML}{ff0000}
\definecolor{EFlv}{HTML}{ff0000}
\definecolor{EFhi}{HTML}{ff0000}
\definecolor{EFc}{HTML}{b22222}
\newcommand{\EFc}[1]{\textcolor{EFc}{#1}} % font-lock-comment-face
\definecolor{EFcd}{HTML}{b22222}
\newcommand{\EFcd}[1]{\textcolor{EFcd}{#1}} % font-lock-comment-delimiter-face
\definecolor{EFs}{HTML}{8b2252}
\definecolor{EFd}{HTML}{8b2252}
\definecolor{EFm}{HTML}{008b8b}
\definecolor{EFk}{HTML}{9370db}
\newcommand{\EFk}[1]{\textcolor{EFk}{#1}} % font-lock-keyword-face
\definecolor{EFb}{HTML}{483d8b}
\definecolor{EFf}{HTML}{0000ff}
\newcommand{\EFf}[1]{\textcolor{EFf}{#1}} % font-lock-function-name-face
\definecolor{EFv}{HTML}{a0522d}
\definecolor{EFt}{HTML}{228b22}
\newcommand{\EFt}[1]{\textcolor{EFt}{#1}} % font-lock-type-face
\definecolor{EFo}{HTML}{008b8b}
\newcommand{\EFo}[1]{\textcolor{EFo}{#1}} % font-lock-constant-face
\definecolor{EFwr}{HTML}{ff0000}
\definecolor{EFpp}{HTML}{483d8b}
\definecolor{EFOa}{HTML}{0000ff}
\definecolor{EFOb}{HTML}{a0522d}
\definecolor{EFOc}{HTML}{a020f0}
\definecolor{EFOd}{HTML}{b22222}
\definecolor{EFOe}{HTML}{228b22}
\definecolor{EFOf}{HTML}{008b8b}
\definecolor{EFOg}{HTML}{483d8b}
\definecolor{EFOh}{HTML}{8b2252}
\definecolor{EFhn}{HTML}{008b8b}
\newcommand{\EFhn}[1]{\textcolor{EFhn}{#1}} % highlight-numbers-number
\definecolor{EFhq}{HTML}{9370db}
\definecolor{EFhs}{HTML}{008b8b}
\definecolor{EFrda}{HTML}{707183}
\definecolor{EFrdb}{HTML}{7388d6}
\definecolor{EFrdc}{HTML}{909183}
\definecolor{EFrdd}{HTML}{709870}
\definecolor{EFrde}{HTML}{907373}
\definecolor{EFrdf}{HTML}{6276ba}
\definecolor{EFrdg}{HTML}{858580}
\definecolor{EFrdh}{HTML}{80a880}
\definecolor{EFrdi}{HTML}{887070}
\usepackage{biblatex}

\begin{document}

\dept{Computer Science}
\principaladvisor{Andreas Krause}
\coprincipaladvisor{Mykel Kochenderfer}
\firstreader{Jonas Rothfuss}

\submitdate{August 2022}
\copyrightfalse
\tablespagefalse
\beforepreface

\prefacesection{Abstract}
Although an ever-growing number of applications employ deep learning based systems for prediction, decision-making, or state estimation, almost no certification processes have been established that would allow such systems to be deployed in \emph{safety-critical applications}.
In this work we consider, among others, real-world problems arising in aviation, medical decision-making, and industrial control, and investigate their requirements for a certified model.
To this end, we investigate methodologies from the machine learning research community aimed towards verifying robustness and reliability of deep learning systems, and evaluate these methodologies with regard to their applicability to real-world problems.
Then, we establish a new framework towards deep learning certification based on (i) inherently safe design, and (ii) run-time error detection.
Using a concrete use case from aviation, we show how deep learning models can recover \emph{disentangled variables} through the use of weakly-supervised representation learning.
We argue that such a system design is inherently less prone to common model failures, and can be verified to encode underlying mechanisms governing the data.
Then, we investigate four techniques related to the run-time safety of a model, namely (i) uncertainty quantification, (ii) out-of-distribution detection, (iii) feature collapse, and (iv) adversarial attacks.
We evaluate each for their applicability and formulate a set of desiderata that a certified model should fulfill.
Finally, we propose a novel model structure that exhibits all desired properties discussed in this work, and is able to make regression and uncertainty predictions, as well as detect out-of-distribution inputs, while requiring no regression labels to train.
We conclude with a discussion of the current state and expected future progress of deep learning certification, and its industrial and social implications.

\afterpreface

\chapter{Introduction}
\label{sec:introduction}
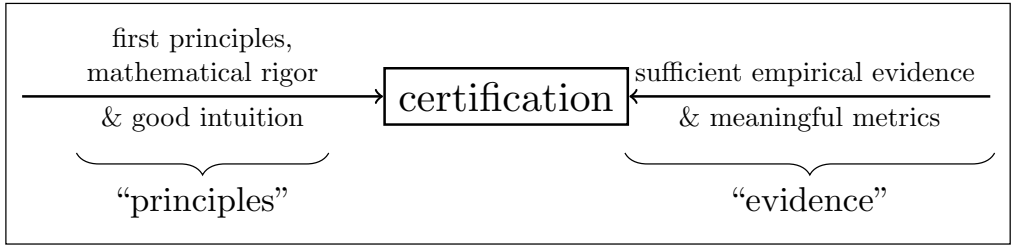
\begin{figure}[t]
\centering
\begin{tikzpicture}[framed]
    \node[draw, scale=1.2, thick] (center) at (0,0) {certification};
    \coordinate[left=4.0cm of center] (lhs);
    \coordinate[right=4.0cm of center] (rhs);
    \draw[->, align=center, thick] (lhs) to
    node[above, scale=0.8] (text11) {first principles, \\ mathematical rigor}
    node[below, scale=0.8] (text12) {\& good intuition}
    (center);
    \draw[->, thick] (rhs) to
    node[above, align=center, scale=0.8] (text21) {sufficient empirical evidence }
    node[below, scale=0.8] (text22) {\& meaningful metrics}
    (center);

    \coordinate[below=0.3cm of center] (brace-row);
    \draw[decorate, decoration={brace, mirror, amplitude=7pt}] (text11.west |- brace-row) to node[below=7pt] {{``principles''}} (text11.east |- brace-row);
    \draw[decorate, decoration={brace, mirror, amplitude=7pt}] (text21.west |- brace-row) to node[below=7pt] {{``evidence''}} (text21.east |- brace-row);
\end{tikzpicture}
\caption{\label{fig:how-to-certify}The two fundamental directions towards building a certification protocol.}
\end{figure}

\noindent In recent years, deep learning based systems have become of ever-increasing importance in \emph{safety critical} domains such as medical decision-making, autonomous driving, aviation, finance, and power plant control.
In such settings, deep learning approaches often instantiate as autonomous perception and decision-making systems.
Decisions are made in real-time, often without additional supervision by a human.
Due to the safety-critical nature of these fields, any errors can have catastrophic consequences.
Certifying functional safety of these models is therefore of vital importance for regulators, engineers, and consumers.
Additionally, certification can help to answer questions of liability and help consumers make informed choices about their products.

Traditional certification procedures commonly combine mathematical principles with empirical evidence, see \cref{fig:how-to-certify}.
For example, they may rely on formal analysis, path based analysis, or event testing.
Such certification procedures analyze all possible execution paths of the system for correctness and consider a large but finite number of inputs and events.
Further, certification often relies on mathematical proofs about the correctness of the certified system.
These proofs may be supported by assumptions about the physical system which can be constructed from first principles, or empirically validated.

Unfortunately, for learning based systems many of these approaches are not applicable.
In particular, modern deep learning based systems often encode little prior knowledge of the principles that govern the relationships between inputs \(x\) and outputs \(y\).
Instead, they represent a class of machine learning methods that capture statistical dependencies between \(x\) and \(y\) with high accuracy, but do not capture a concise set of underlying principles.
Moreover, while traditional models represent intermediate computations in a structured form, neural networks instead use a large number -- often millions -- of unstructured intermediate computations that are linked by simple functions.

In practice, this leads to substantial problems for certification.
Firstly, due to the lack of intermediate symbolic variables, it is generally hard or impossible to verify whether known mechanisms or invariants of the data are correctly captured by the model.
Secondly, due to the large number of intermediate variables, most certification approaches relying on enumeration of execution paths or inputs become computationally infeasible due to the combinatorial nature of the many variables.
Finally, despite large progress in recent years, it is still hard to leverage mathematical proofs in order to verify correctness of constructed models.
For instance, construction of deep learning models requires optimizing high-dimensional non-linear non-convex functions, and the outcome of the optimization is highly stochastic.
Thus, proofs about the convergence to a global minimum are generally inapplicable.

In light of these difficulties, several old and new domains of machine learning research have (re-)emerged in order to address the issue of certification, or general robustness verification, from a variety of perspectives.
For instance, results from statistical learning theory have been applied to deep learning in order to prove ``generalization bounds'', i.e. bounds on the error when a model is evaluated on new data;
representation learning methods have been proposed to recover low-dimensional representations of the data that satisfies fundamental properties about the data;
and uncertainty quantification is used to predict the magnitude of model prediction's deviation from the ground truth.

Despite these efforts, there exists a significant gap between achievements in the academic domain and certification requirements emerging for real-world safety-critical applications.
Several regulation agencies like the Food and Drug Administration (FDA), Federal Aviation Administration (FAA), and European Union Aviation Safety Agency (EASA) have recently proposed timelines and made efforts towards certification and regulation of deep learning based methods in safety-critical systems.
However, at this point it is not clear which methodologies both satisfy the strong safety standards of the domain, e.g. aviation, and are practically applicable at the same time.

The goal of this work is therefore to investigate the current gap between machine learning methodology aimed towards certification, and their applicability in real-world safety-critical applications.
To this end, we first briefly study current advances in artificial intelligence (AI) regulation for industry applications, including aviation, medicine, and autonomous driving.
We contrast these with current efforts in the machine learning research community to propose guidelines for certifiable AI.
Then we introduce a new taxonomy of current machine learning subfields useful for AI certification and regulation.
In this taxonomy, we informally evaluate each subfield with respect to its applicability, as well as mathematical rigor, and draw conclusions for their potential in a certification framework.
To aid this process, we limit the discussion to a set of substantially restricted problems which naturally arise in the field of robotics.
We introduce a specific use case involving pose estimation of an aircraft w.r.t. a runway, but argue that the assumptions made are applicable to a wide range of applications.

Having established the assumptions and example use case, we focus on two broad approaches to certification, namely (i) inherently safe design, and (ii) run-time error detection.

Regarding inherently safe design, we review recent studies connecting \emph{causality}, \emph{disentanglement}, and \emph{representation learning}, and present a novel framework for modeling real world systems using these principles.
In particular, we explore how regression predictions can be modeled through the numerical representation of semantic variables which govern the data generating process, and how representation learning can be used to recover these representations.

Regarding run-time error detection, we address a variety of methodologies useful for safety in the real-time setting, namely (a) uncertainty quantification, (b) out-of-distribution detection, (d) feature collapse, and (d) the risks of adversarial attacks and defenses.
We formulate a set of desiderata for each and provide principles and procedures aimed at empirically validating the satisfaction of the desiderata.
We formalize the procedures through executable code snippets that aim to reduce ambiguity and provide immediate applicability.

Having established the certification framework, we propose a novel deep learning based method that aims to fulfill all certification principles established in this work by combining several methods known from in literature in a novel way.

Finally, we conclude this work with a discussion of the future of certification for deep learning based systems in safety-critical settings, and highlight current efforts that we expect to have a significant impact in the coming years.

\section{Organization of this work}
\label{sec:organization}
\noindent \cref{sec:industry-needs} summarizes recent developments in industry pushing towards certification of machine learning and deep learning based systems.
\cref{subsec:proposed-taxonomy} presents our proposed taxonomy of machine learning approaches useful for ML certification, and reviews a set of recent publications that propose steps towards certification of deep learning based systems.
\cref{sec:fundamental-assumptions} introduces a concrete use case involving a 2D pose estimation for an aircraft on a runway, which will serve as the leading example throughout this work.
Subsequently, it defines a strongly restricted problem setting to certify by specifying a set of assumptions.
\cref{sec:deriving-a-framework} introduces a set of principles that can be used for the certification of deep learning based systems.
In particular, in \cref{subsec:inherently-safe-design} we discuss inherently safe design through the lens of disentanglement, and provide desiderata, architectural principles and methods for empirical evidence.
Conversely, in \cref{subsec:runtime-error-detection} we discuss run-time error detection from four different angles, and again provide desiderata, architectural principles and methods for empirical evidence.
\cref{sec:proposed-model-and-dataset} proposes a novel architecture which aims to fulfill the principles defined in the previous chapter.
Finally, \cref{sec:conclusion} reviews the results and their broader impact and discusses future research directions.

\section{Contributions}
\label{sec:org2176aa5}
The main contributions of this work are
\begin{enumerate}[label=\Roman*.]
\item a taxonomy of current subfields of machine learning research and their relation to deep learning certification in safety-critical fields;
\item a new perspective of how regression variables can be modeled through semantic features, which are directly recovered through a form of disentangled representation learning;
\item a set of principles and empirical tests for verifying critical properties of the disentangled semantic variables;
\item a set of principles and empirical tests for several methods supporting run-time error detection; specifically uncertainty quantification, out-of-distribution detection, feature collapse, and adversarial attacks; and
\item a novel weakly-supervised model structure that allows numerical prediction of disentangled regression variables and aims to fulfill the inherently safe design principles established in this work.
\end{enumerate}

\chapter[Current Progress in AI Certificaton]{Current Progress for AI Certification in Safety-critical Applications}
\label{sec:industry-needs}
\section{Industry applications}
\label{sec:orgfa27538}
\begin{figure}[t]
\centering
\includegraphics[width=\textwidth]{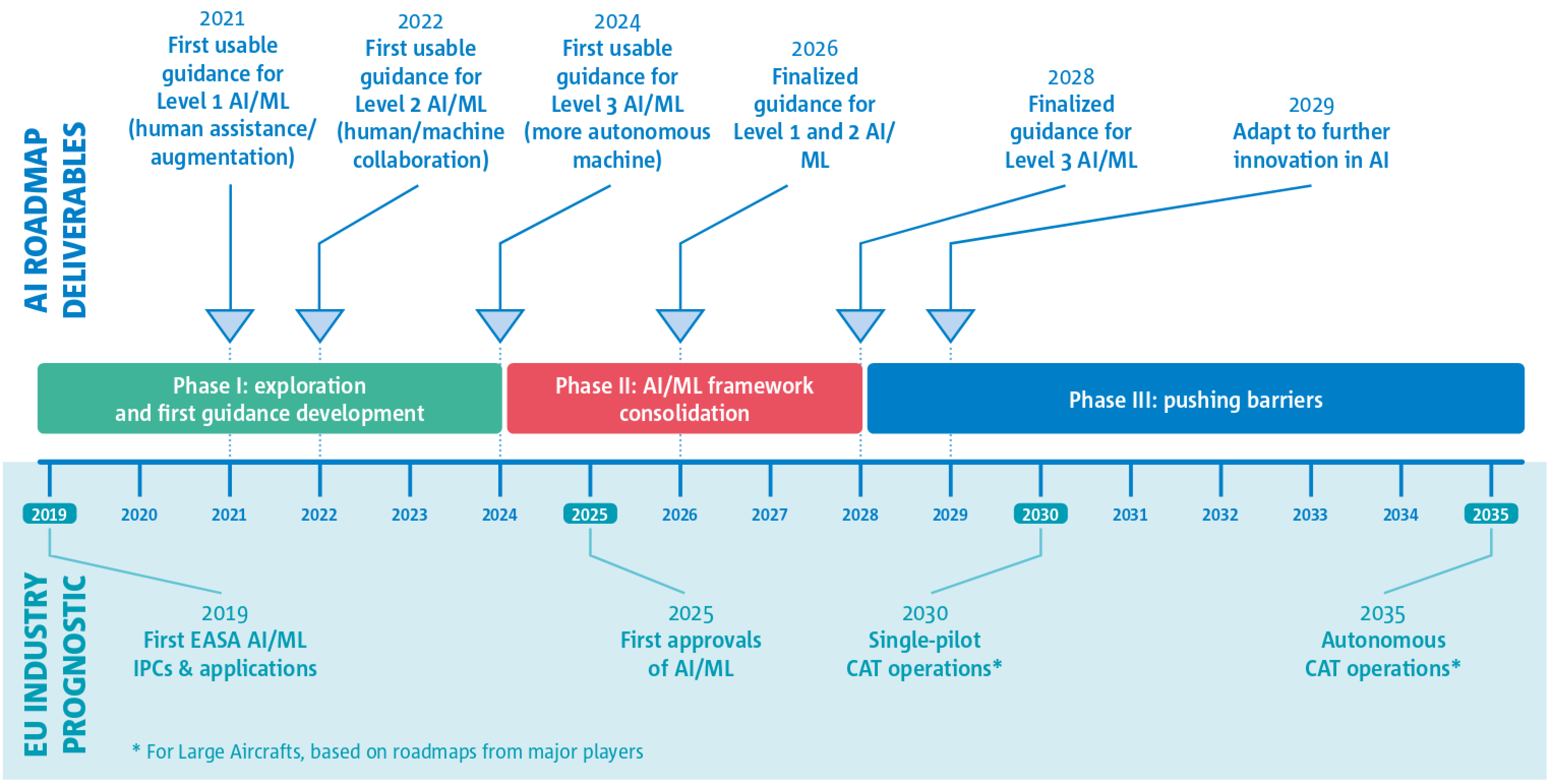}
\caption{\label{fig:easa-roadmap}The EASA artificial intelligence roadmap, proposed in 2020. Reprinted from \autocite{easaEASAArtificialIntelligence2020}.}
\end{figure}

One of the most ambitious public studies for machine learning certification is the progress made by the European Union Aviation Safety Association (EASA).
In 2020, the EASA published a ``roadmap'' for integrating AI based systems into the aviation sector, starting with by approving human assistance and human machine collaboration systems by 2025, and having fully autonomous commercial air transport operations by 2035 \autocite{easaEASAArtificialIntelligence2020}.
The roadmap is reprinted in \cref{fig:easa-roadmap}.

Following this roadmap, EASA developed and published two reports towards ``Concepts of Design Assurance for Neural Networks'' together with an industry partner \autocite{easaConceptsDesignAssurance2020,easaConceptsDesignAssurance2021}, and has additionally released a concept paper describing a ``First usable guidance for Level 1 machine learning applications'' \autocite{easaEASAConceptPaper2021}.
Key components of the design assurance and published guidance are (i) trustworthiness analysis, (ii) learning assurance \& dataset assurance, (iii) AI explainability, and (iv) AI safety risk mitigation.

Similarly, in 2022 the US-based Federal Aviation Administration (FAA) published a report on ``Neural Network Based Runway Landing Guidance for General Aviation Autoland'' \autocite{faaNeuralNetworkBased2022}, which has been developed with the same industry partner as the reports by EASA.
In this report the specific application of an aircraft autonomously landing on a runway using a single camera is investigated.
In particular, a deep learning based AI model is used for visual pose estimation and uncertainty quantification, and empirical performance is evaluated.

In the sector of autonomous driving, in 2019 a joint report on ``Safety first for automated driving'' was developed and published by eleven automotive companies \autocite{daimlerSafetyFirstAutomated2019}.
In this report, twelve principles of automated driving have been established, namely
(i) ensuring safe operation,
(ii) specifying and verifying a operational design domain,
(iii) allowing vehicle operator-initiated handover,
(iv) proving computer security,
(v) maintaining user responsibility,
(vi) allowing vehicle-initiated handover,
(vii) considering the interdependency between the vehicle operator and the autonomous driving system,
(ix) requiring safety assessment,
(x) requiring data recording,
(xi) ensuring passive safety, and
(xii) providing predictable behavior in traffic.
Additionally, the report also describes challenges when involving deep learning models into the driving system, and states concerns about (i) the operational design domain, (ii) dataset attributes and challenges, (iii) probabilistic predictions, (iv) performance indicators, and (v) hardware.

In the medical field, in 2018 a first device was approved making autonomous screening decisions and using machine learning (although not deep learning) \autocite{abramoffPivotalTrialAutonomous2018}.
Since then, discussion around certification for AI based systems in medical decision-making has substantially increased, and regulatory approval for such systems has slowed down.
Since 2019, there is an ongoing discussion about a ``Software as a Medical Device (SaMD) Action Plan'' \autocite{centerfordevicesandradiologicalhealthArtificialIntelligenceMachine2021}.
We note that, in the medial domain, the medical classification of a device has important implications on its regulation, and discussions about the classification seem to cause a delay in progress for certification overall \autocite{harveyHowFDARegulates2020a}.
Further, some authors argue that a ``system view'' needs to be adapted for AI certification, that involves understanding the implications of such autonomous decision-making systems when interfacing with humans, for example when considering the interaction with humans making medical decisions based on the system's outputs \autocite{gerkeNeedSystemView2020}.
Several ``checklists'' have been proposed involving the evaluation and design of AI systems in the medical domain \autocite{larsonRegulatoryFrameworksDevelopment2021,cabitzaNeedSeparateWheat2021}.
Finally, some work has been designated to fundamental principles of deep learning systems and data properties, including dataset shift, causality, and shift-stable models \autocite{subbaswamyEvaluatingModelRobustness2021}.

\section{Recent progress on deep learning certification from the research community}
\label{sec:org993d414}

\begin{figure}[t]
\centering
\begin{tikzpicture}[show background rectangle]

  \node[draw, minimum height=4.5cm, fill=green!100!red!10] (lhs) {
    \begin{tikzpicture}[minimum height=0cm]
      \node[align=center, text width=4.5cm] (title) {\textsc{Inherently Safe} \\ \textsc{Design} };
      \node[text width=5cm, below=0 of title] {
          \begin{itemize}
            \item Model Transparency
            \item Design Specification
            \item Model Verification and Testing \phantom{wow this is a longg textt}
          \end{itemize}
        };
    \end{tikzpicture}
  };

  \node[draw, right=of lhs, minimum height=4.5cm, fill=yellow!100!red!10] (mid) {
    \begin{tikzpicture}[minimum height=0cm]
      \node[align=center, text width=4.5cm] (title) {\textsc{Enhancing Performance} \\ \textsc{and Robustness}};
      \node[text width=5cm, below=0 of title] {
          \begin{itemize}
            \item Robust Network Architecture
            \item Robust Training
            \item Data Sampling and Augmentation
          \end{itemize}
        };
    \end{tikzpicture}
  };

  \node[draw, right=of mid, minimum height=4.5cm, fill=green!100!red!10] (rhs) {
    \begin{tikzpicture}[minimum height=0cm]
      \node[align=center, text width=4.5cm] (title) {\textsc{Run-time Error} \\ \textsc{Detection}};
      \node[text width=5cm, below=0 of title] (items) {
          \begin{itemize}
            \item Prediction Uncertainty
            \item Out-of-distribution Detection
            \item Adversarial Attack Detection and Guard
          \end{itemize}
        };
      \node[below=0 of items, inner sep=1pt] {\phantom{.}};
    \end{tikzpicture}
  };
  
\end{tikzpicture}
\caption{\label{fig:mohseni-taxonomy}A taxonomy of machine learning safety as developed by \textcite{mohseniTaxonomyMachineLearning2022}. In this work we focus on inherently safe design and run-time error prediction.}
\end{figure}
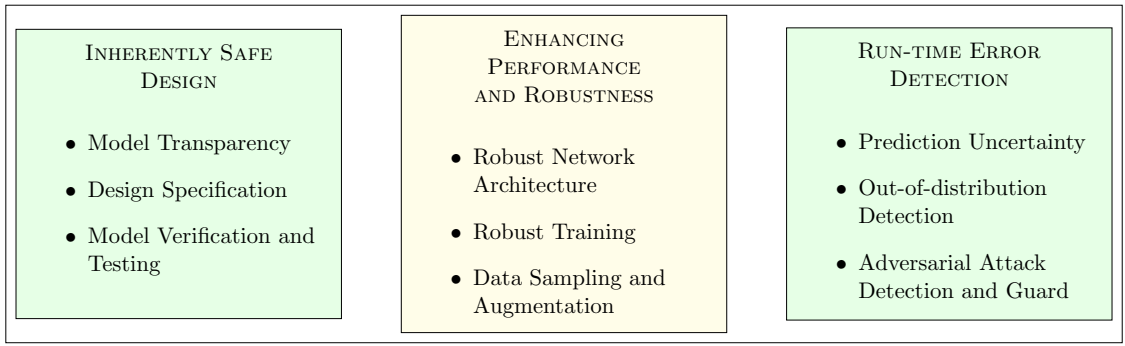

\noindent In 2022, \textcite{mohseniTaxonomyMachineLearning2022} presented a ``Taxonomy of Machine Learning Safety'', discussing key engineering safety principles commonly used for non-AI based systems,  and relating them to fundamental limitations inherent to machine learning systems.
Then, they propose a taxonomy of machine learning methodologies useful for robustness and certification, consisting of the categories (i) inherently safe design, (ii) enhancing performance \& robustness, and (iii) run-time error detection.
In this work, we adopt this classification and will consider especially categories (i) and (iii).

Another notable work has been published by \textcite{seshiaVerifiedArtificialIntelligence2020} concerning ``Verified Artificial Intelligence''.
The authors discuss the current gap between formal methods and their applicability to machine learning and deep learning systems, and propose a set of five verification perspectives, namely
(i) environment modeling,
(ii) formal specification,
(iii) modeling learned systems,
(iv) efficient and scalable design and verification of models and data, and
(v) Correct-by-Construction Intelligent Systems.

Finally we also mention recent work on a explainable AI \autocite{mohseniMultidisciplinarySurveyFramework2021}, trustworthy AI, \autocite{shneidermanHumanCenteredArtificialIntelligence2020}, and responsible AI \autocite{barredoarrietaExplainableArtificialIntelligence2020c}.

\section{A Proposed  Taxonomy of Deep Learning  Safety Methodologies}
\label{subsec:proposed-taxonomy}
\begin{figure}[htbp]
\centering
\includegraphics[width=\linewidth]{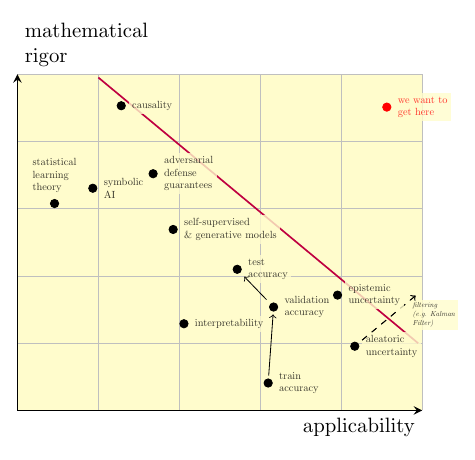}
% \includestandalone[width=\linewidth]{tikz-pictures/2d-taxonomy/standalone/2d-taxonomy}
\caption{\label{fig:2d-taxonomy-alignment}Our taxonomy of deep learning methodologies useful for certification, roughly aligned by their applicability (as of 2022) for addressing real-world problems, and their mathematical rigor, each estimated informally by the author. In this work we will discuss all listed aspects except for statistical learning theory.}
\end{figure}

In \cref{fig:2d-taxonomy-alignment} we propose a novel taxonomy of machine learning subfields that can be used to increase trust and certifiability of deep learning systems.
We informally evaluate each subfield by its \emph{applicability} to problems arising in the real world, and by its \emph{mathematical rigor}, i.e. how strong any provided guarantees are if required assumptions hold.

\chapter{Fundamental Assumptions and a Use Case}
\label{sec:fundamental-assumptions}
Learning systems and the problems we apply them to come in a large variety of forms.
Considering deep learning certification for arbitrary problems therefore is a Herculean task without significantly constraining the problem.
In this section, we therefore introduce a strongly restricted setting, motivated by the applications outlined in \cref{sec:industry-needs}, and introduce a set of confining assumptions which will help us further develop concrete certification requirements in the following chapters.
In particular, we place restrictions on the inherent structure of the problem by making assumptions about the existence of high-level semantic properties of the inputs, about certain statistical independencies, and about properties of the error distributions.
We note that, although these assumptions are restrictive, most can be partially relaxed to arrive at similar certification requirements for more general settings.
Where appropriate, this work provides footnotes to discuss where such relaxations are possible.

In order to motivate each assumption, we start by providing a concrete use case that highlights properties of real-world scenarios which we can use to derive desiderata used in the certification process.
Then, we specify a list of five concrete assumptions about the problem task and relate them to the use case.
Subsequently, in \cref{sec:deriving-a-framework} we discuss how we can use each assumption to derive specific tests and principles that we can apply to our system.

\section{Use case: Pose estimation on a runway}
\label{subsec:use-case-pose-estimation}
\begin{figure}[t]
\centering
\includegraphics[width=0.5\textwidth]{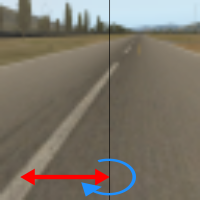}
\caption{\label{fig:runway-annotated}Example use case: Prediction of two pose variables, namely \emph{horizontal offset} (red) and \emph{rotational offset} or \emph{yaw} (blue), for an aircraft on a runway.}
\end{figure}

Motivated by \cref{sec:industry-needs}, we present a concrete use case as an illustrative example throughout the following sections.
In particular, we consider the problem of recovering the pose (position and orientation) of an aircraft on a runway using a single camera mounted on the aircraft.
\cref{fig:runway-annotated} illustrates a possible input for this problem.
A camera is mounted on the airplane and has a view of the runway.
Both sidelines and some runway features are visible, as well as parts of the background and other image artifacts.
For simplification, we restrict the problem to a two-dimensional pose estimation by considering only (i) horizontal offset and (ii) rotational offset (yaw), but note that we can easily extend the problem to the full 6-degree-of-freedom setting.
We further make the assumption that the operational range of the pose is clearly defined; specifically, we consider horizontal offsets of \(\pm 10\mathrm{m}\) and rotational offsets of \(\pm 15^{\circ}\).
Crucially, this implies that we do not require the model to generalize outside these ranges.
Instead, we expect the model to detect if the operational range has been violated and ``reject'' such inputs.

Although we dictate precise restrictions on the pose variables, we still require the model to operate in a wide variety of other conditions.
These include irregular lighting conditions, different runways, and undefined objects, but may also include other artifacts which can not be well-defined a priori.
It therefore makes sense to distinguish how the image is influenced by the pose and by other factors, or, in other words, by the variables we are trying to \emph{predict} and by those we are trying to \emph{ignore}.

To this end, we make the fundamental assumption that the input image is the result of a generative process that is a function of a set of high-level semantic variables.
These variables include the pose, the time of the day, the choice of the runway, and others, which we may not be able to define.
Intuitively, each high-level semantic variable ``causes'' certain intermediate semantic properties and ultimately some pixels in the image.
There may be complex dependencies between any two or more variables during the generative process.
For example, an increase in aircraft rotation will cause the sidelines of the runway to rotate in the image as well.
How much the sidelines rotate depends on the horizontal offset, but does not depend on the time of the day or the choice of the runway.
Conversely, the sun angle may cause reflections or shadows on the runway, which will look different depending on the type of tarmac; the sun angle however will not influence the pose of the aircraft or the angle of the sidelines.

Motivated by these dependencies, or the lack thereof, we argue that it seems reasonable to split the semantic variables into separate groups which do not have a strong dependency on each other during the data generating process.
More specifically, for two variables from different groups we require them to be independent of each other; in other words a change in one does not cause a change in the other.

\begin{figure}[t]
\centering
\begin{subfigure}[t]{0.45\textwidth}
\includegraphics[width=.9\linewidth]{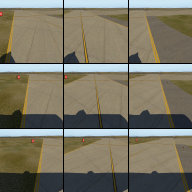}
\caption{\label{fig:runway-example1} Content change.}
\end{subfigure}
\begin{subfigure}[t]{0.45\textwidth}
\includegraphics[width=.9\linewidth]{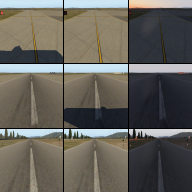}
\caption{\label{fig:runway-example2} Style change.}
\end{subfigure}
\caption{\label{fig:runway-example-both}Example of independent \emph{content} and \emph{style} changes. In (a), the camera/airplane changes rotation (left to right) and horizontal offset (top to bottom). Conversely, in (b) the time of day (left to right) and runway location (top to bottom) changes.}
\centering
\end{figure}

We now introduce the crucial assumption that the values that we want to predict, e.g. horizontal and rotational offset, may be directly related to high-level semantic variables, and that we may group those semantic variables into a single group by themselves.
This group is disjoint from a second group containing all ``other'' semantic variables, modeled or not.
In the runway example the first group therefore consists of only the horizontal and rotational offset, and the second group consists of the time of day, runway choice, as well as any other known or unknown high-level semantic variables.
These two groups will play a major role in the rest of this work, and we will refer to them as the \emph{content} variables and the \emph{style} variables, where the content variables are associated with the numerical values we are trying to predict.

To illustrate this idea, \cref{fig:runway-example-both} exhibits the effect of changing the variables in one group while keeping the variables in the other group constant.
In particular, \cref{fig:runway-example1} illustrates how the semantics of the input change when varying the pose, but keeping everything else constant.
On the other hand, \cref{fig:runway-example2} illustrates how the semantics of the input change when the pose stays constant, but the image is recorded on different runways and during different times of the day.
In the next section we will further formalize the dependencies of the two groups, as well as dependencies within the groups.

\begin{figure}[htbp]
\centering
\includegraphics[width=0.5\textwidth]{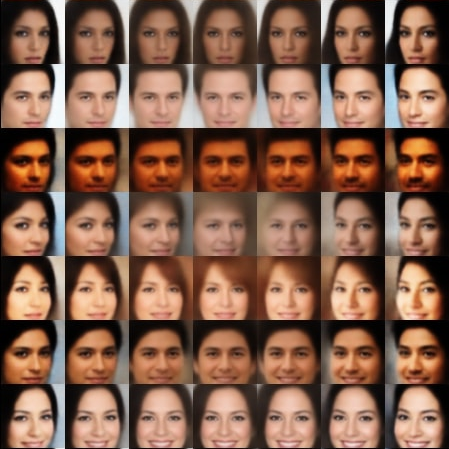}
\caption{\label{fig:azimuth-disentanglement}Content and style separation for faces: Azimuth angle is content and change in person is styles. Reprinted from \textcite{chenIsolatingSourcesDisentanglement2018}.}
\end{figure}

One premise of this work is that we can make the differentiation between content, which contains the semantic features necessary to make the prediction, and style, which we may ignore, in many different problems.
For example, in \cref{fig:bible-verses}, we can see a similar idea applied to natural language: Verses from the bible are written in the style of different authors, but convey the same idea and content.
\cref{fig:azimuth-disentanglement} demonstrates this differentiation for a dataset of faces where the viewpoint angle (content) is varied independently of the specifics of the face (style).
Note that in the original paper presenting the latter figure \autocite{chenIsolatingSourcesDisentanglement2018} the authors also consider other properties like ``baldness'', ``face width'', ``gender'', and ``mustache''.
Depending on the exact task, each of these features may be grouped either with the content or the style variables, as long as they can be separated from the variables in the other group.

\begin{figure}[htbp]
\centering
\begin{tikzpicture}
  \node {
    \begin{tabular}{ll}
      Version & Verse\\
      \hline
      KJV & The heart of the prudent getteth knowledge; and the ear of the wise seeketh knowledge.\\
      ASV & The heart of the prudent getteth knowledge; And the ear of the wise seeketh knowledge.\\
      BBE & The heart of the man of good sense gets knowledge; the ear of the wise is searching for knowledge.\\
      DARBY & The heart of an intelligent getteth knowledge, and the ear of the wise seeketh knowledge.\\
      DRA & A wise heart shall acquire knowledge: and the ear of the wise seeketh instruction.\\
      LEB & An intelligent mind will acquire knowledge, and the ear of the wise will seek knowledge.\\
      WEB & The heart of the discerning gets knowledge. The ear of the wise seeks knowledge.\\
      YLT & The heart of the intelligent getteth knowledge, And the ear of the wise seeketh knowledge.\\
    \end{tabular}
  };

\end{tikzpicture}
\caption{\label{fig:bible-verses}Content and style separation in language: A verse from the bible (Proverbs 18:15) written in the style of different translators, while maintaining the same meaning. Reprinted from \textcite[, Table 7]{vishnubhotlaEvaluationDisentangledRepresentation2021}.}
\end{figure}

\begin{figure}[htbp]
\centering
\includegraphics[width=0.3\textwidth]{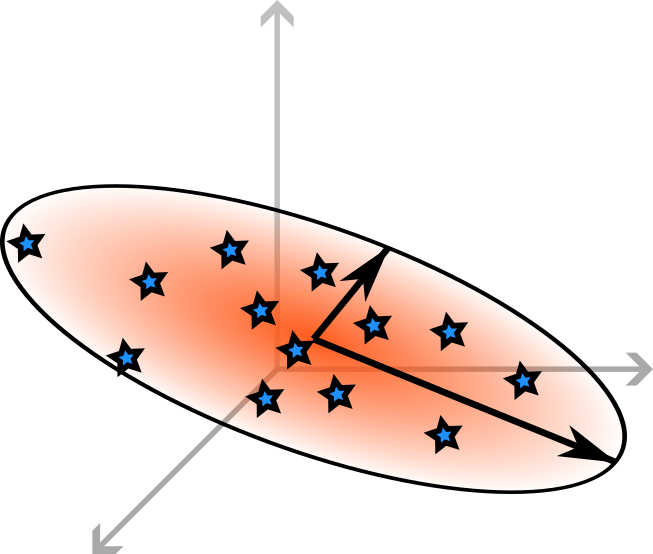}
\caption{\label{fig:pca}An example of principle component analysis constructing a two dimensional latent space from three dimensional data. The latent dimensions are uncorrelated, but generally interpretable, i.e. do not represent a semantic property.}
\end{figure}

\section{Five central assumptions}
\label{subsec:five-central-assumptions}
Motivated by the use case in \cref{subsec:use-case-pose-estimation} we present five central assumptions which constrain the problem class and help us derive concrete principles and testing procedures in \cref{sec:deriving-a-framework}.

To this end we first briefly introduce the concept of \emph{disentanglement}, which helps us formalize the relation between semantic variables which do not have a strong dependence during the data generation process.
\begin{quote}
\textbf{Definition: (Feature disentanglement).} We define feature disentanglement as statistical independence between two variables on a semantic level.
In particular, following \textcite{higginsBetaVAELearningBasic2017}, we consider two variables \(V_1\) and \(V_2\) as disentangled if they are
\begin{enumerate}
\item conditionally independent, i.e. \(p(v_1, v_2) = p(v_1)p(v_2)\), and
\item represent humanly interpretable, high-level semantic concepts in a numerical form.
\end{enumerate}
To denote disentanglement between \(V_1\) and \(V_2\) we write \(V_1 \dis V_2\).
\end{quote}
The idea of disentanglement therefore let's us more clearly define how variables from the content and style groups are related.
Intuitively, we can say that knowing the value of one semantic variable does not tell us anything about the others.
For instance, only knowing the time of the day does not help us improve our estimate of the likelihood of any particular aircraft rotation.

We can also understand the existence of disentangled features as a restricted form of dimensionality reduction.
Regular dimensionality reduction techniques are typically only statistical; for instance Principal Component Analysis will find uncorrelated features from statistical data; see \cref{fig:pca} for an illustration.
These features will generally lie in an abstract embedding space, which is hard or impossible to interpret; in other words they have no semantic meaning.
Disentangled variables on the other hand must represent the high-level ``causes'' of the image features and may need to be specified by manually, rather than finding them computationally.\footnote{We will discuss the automatic discovery of semantic features in \cref{subsubsec:self-supervised-recovery-of-disentangled-vars}.}

We will now formalize five assumptions about the problem setting motivated by the discussion so far.

\subsubsection{Assumption 1: Semantic representation and hidden generative model.}
\label{sec:org0235e59}
Given a specific problem, we assume that we can represent each input using a low-dimensional set of semantic variables, namely its \emph{content} and \emph{style} features, denoted respectively as \(v^c\) and \(v^s\), or in conjunction as \(v = (v^c, v^s)\).
Further we assume the existence of an unknown stochastic generative process
\begin{equation} x = g^{\ast}(v) \end{equation}
that samples the data \(x\) from an unknown distribution
\begin{equation}p^{\ast}(x \mid v)\end{equation}
of all feasible data points which we may associate with \(v\).\footnote{Instead of sampling from a distribution, we could instead model the output of \(g^{\ast}(v)\) as choosing from a set \(\mathcal{S}\).
    The issue is that a set implies a clear boundary, i.e. the existence of points \(\bar{x}\) on the boundary of \(\mathcal{S}\) for which \(\lim_{x \nearrow \bar{x}} x \in \mathcal{S}\) and \(\lim_{x \searrow \bar{x}} \notin \mathcal{S}\), where \(\nearrow\) and \(\searrow\) denote limits from opposite directions.
    For images we assume that generally no such set exists, and that instead the probability of any \(x\) approaches zero as we move away from the most likely samples.}

We motivate this as follows: When considering the relationship between \(v\) and \(x\) it seems clear that, given only the concise semantic description \(v\), we may associate an infinite number of input samples \(x\) to \(v\), each varying in small details.
For instance, in the runway example, the input images may vary in the specifics of the precise sun position, the grass or tarmac texture, or some noise in the pixel values, while maintaining the same semantic description \(v\).
Additionally, the set of semantic variables may be incomplete, giving rise to a variety of likely values for \(x\).

In subsequent chapters we will use the existence of \(g^\ast\) to motivate the feasibility of constructing an approximate generative model \(g \approx g^\ast\).

\subsubsection{Assumption 2: Full content disentanglement and numerical representation.}
\label{sec:orgca295e7}
We assume that we are in a regression setting where the goal of the model is to predict a numerical representation of the content variables, e.g. the pose in the runway example.
This implies that each content variable has a representation on a bounded subset of \(\mathbb{R}\).
(In contrast, the style variables may be inherently continuous or discrete.)
Crucially, we further assume that each content variable \(v^c_i\) is \emph{disentangled from all other content variables}, i.e. for all \(i \neq j\) we can write
\begin{equation} v^c_i \dis v^c_j
\end{equation}
where \(\dis\) denotes feature disentanglement as introduced above.\footnote{Although this work only discusses the simplified ``fully disentangled'' setting, the ideas are generalizable to a ``groupwise disentangled'' setting.
In that setting, instead of each \(i\) and \(j\) being disentangled, subsets of content variables may be entangled with each other, but still disentangled from all other content variables.
More formally, the disentanglement structure can then be defined by a set of mutually disjoint index sets \(\{\mathcal{I}_k\}_k\) such that, given \(i \in \mathcal{I}_k\) and \(j \in \mathcal{I}_{k'}\), \(v^c_i\) and  \(v^c_j\) are entangled iff \(i\) and \(j\) come from different index sets, i.e. \(k \neq k'\).}

Applied to the runway example, this means that knowing only the rotational offset \(v^c_2\) gives us no further information about the horizontal offset \(v^c_1\), as \(p(v^c_1, v^c_2) = p(v^c_1)\cdot p(v^c_2)\) and therefore
\begin{equation}p(v^c_1 \mid v^c_2) = \frac{p(v^c_1, v^c_2)}{p(v^c_2)} = p(v^c_1).\end{equation}

\subsubsection{Assumption 3: Disentanglement between content and style.}
\label{sec:orgef89c7c}
In a similar fashion to Assumption 2, we also assume that each content variable is disentangled from all style variables, i.e.
\begin{equation}p(v^c_i \mid v^s) = p(v^c_i) \end{equation}
for all \(i\), and therefore also
\begin{equation}p(v^c \mid v^s) = p(v^c). \end{equation}

In the runway example this means that knowing something about the time of day, runway choice, lighting conditions, etc. does not help us improve our prediction of the pose.

\subsubsection{Assumption 4: Known prior and complete coverage of content variables.}
\label{subsubsubsec:assumption-4}
We assume that precise operating requirements for the application are defined, i.e. that we know all possible realizations of the content variables, and that we have access to training samples that cover \emph{the whole support} of the content variables.

In other words, as long as the algorithm is within operating requirements, we assume that the content features will not have novel realizations.
If realizations of the content variables outside of the operational parameters do occur, we do not require the model to generalize, but instead require a ``rejection'' by the model.
(Note however that for the style variables novel realizations may occur, and the model must be able to either recover the content variables, or reject the input. It is however not necessary to recover the style variables precisely.)

In addition to knowing the possible realizations of the content variables, we also assume that we know their prior distribution \(p(v^c_i)\) for each \(i\), which we will assume to be uniform.\footnote{We may also assume other prior distributions, for example a Gaussian distribution, but note that if the prior has infinite support we must relax the assumption of having ``full coverage''.}

Applied to the runway example this assumption implies that we expect the content realizations \(v^c\) associated with the training set to be i.i.d. in \[\mathit{Uniform}[-10 \mathrm{m}, 10\mathrm{m}]  \times \mathit{Uniform}[-15^\circ, 15^\circ].\]

\subsubsection{Assumption 5: Unimodal mapping between input and semantics.}
\label{sec:org684a2a2}
While we don't make many assumptions on the mapping \(v \mapsto x\), we assume the mapping \(x \mapsto v^c\) to be unique in the sense that there is only one \(v^c\) that may be associated with any \(x\), up to some error margin or uncertainty.
To illustrate, a counter example in the runway use case would be an image that contains two adjacent runways, as it is not clear whether \(v\) should represent the state with regard to the one or the other runway.
Instead there must be a clear ``truth'' to each content variable, although it may be some uncertainty associated.
More formally we assume the distribution \(p(c^c \mid x)\) to have one clearly distinct mode. \footnote{This assumption is mostly a practical one, as many machine learning methods have difficulty predicting a multimodal distribution.
Nonetheless, given an appropriate model we can relax this assumption.}

\chapter{Towards a Certification Framework}
\label{sec:deriving-a-framework}
\begin{quote}
\emph{In this section, we will showcase several algorithms in the form of executable code snippets, implemented in the Julia programming language.
They are meant for providing an immediately applicable proof-of-concept and for providing an unambiguous definition of the proposed algorithms.
For readers unfamiliar with the Julia language, \cref{sec:julia-crash-course} gives a brief overview over non-trivial language features and conventions, but assumes familiarity with the Python programming language.
Nonetheless, we expect readers to be able to follow the implementations even without knowledge of the Julia or Python programming language.}
\end{quote}

\begin{figure}[t]
\centering
\begin{tikzpicture}[framed]

  \node[draw, fill=green!100!red!10, anchor=north] (lhs) {
    \begin{tikzpicture}
      \node[align=center] (title) {\textsc{Inherently Safe} \\ \textsc{Design} };
      \node[text width=5.5cm, below=0 of title] {
          \begin{enumerate}
            \item Recovering and Representing Semantic Features
            \item Disentangling Content and Style Features
            \item Fulfilling Priors for a Good Representation
            \item Model Transparency in the Failure Case
          \end{enumerate}
        };
    \end{tikzpicture}
  };

  \node[draw, right=8 of lhs.north, fill=green!100!red!10, anchor=north] (rhs) {
    \begin{tikzpicture}
      \node[align=center] (title) {\textsc{Run-time Error} \\ \textsc{Detection}};
      \node[text width=5.5cm, below=0 of title] (items) {
          \begin{enumerate}
            \item Calibrated Uncertainty Quantification
            \item Principled Out-of-Distribution Detection
            \item Avoiding Feature Collapse \phantom{looonngg teexxtt}
            \item On Adversarial Attacks and Defenses
          \end{enumerate}
        };
    \end{tikzpicture}
  };
  
\end{tikzpicture}
\caption{\label{fig:my-taxonomy}Major steps in the certification framework developed throughout \cref{sec:deriving-a-framework} of this work.}
\end{figure}
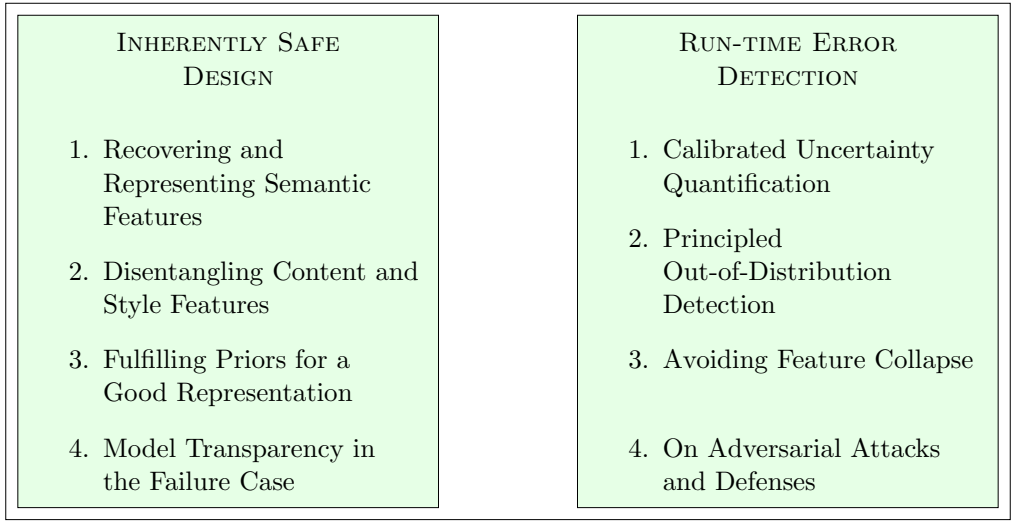

\noindent Having established the use case, goals, and fundamental assumptions of our problem, we will now turn to the next part of this work:
Establishing a framework of principles and empirical evidence that brings us closer to the certification of deep learning based systems in safety-critical domains.

To this end, we build upon the \emph{taxonomy of machine learning safety} introduced by \textcite[, Table 1]{mohseniTaxonomyMachineLearning2022}, and distinguish between (i) inherently safe design, (ii) enhancing performance and robustness, and (iii) run-time error detection (see \cref{fig:mohseni-taxonomy}).
In this work we focus in particular on the topics (i) and (iii), and explore a set of desiderata that a functionally safe and certifiable deep learning system should fulfill.

\section{Inherently Safe Design}
\label{subsec:inherently-safe-design}
Typically, deep learning systems are defined through an end-to-end training problem that constructs a black-box model mapping input samples or observations to output samples or predictions, and capturing the statistical dependencies \(p(y \mid x)\).
Deep learning models are typically over-parameterized, i.e. they have many more parameters than training samples, and are theoretically capable of approximating any arbitrary function \autocite{hornikMultilayerFeedforwardNetworks1989}.
If good machine learning practices are followed, and sufficient training data is available, often a model's performance on the training samples and unseen validation samples can be better than for more traditional methods, especially in high-dimensional domains like computer vision, natural language processing or when computing on data structured as graphs.

Unfortunately, due to this parametrization, it in general very hard to know whether the model learned fundamental principles relating the inputs to the outputs.
In fact, deep learning models can often make ``good'' predictions on the training or validation data using only \emph{spurious correlations}, i.e. using artifacts of the data collection process or data representation, but without capturing any of the underlying mechanisms connecting the inputs and outputs.
Due to the unstructured nature of deep learning models, this case is not trivial to detect, but deploying such a model can have catastrophic consequences if the same spurious correlations are not present during run-time.

Motivated by these issues, we argue that models which are designed to act in safety-critical settings need to exhibit some amount of \emph{computational structure} that allows us to reason about their inner workings.
In particular, having defined a list of physical or conceptual invariants about the data or the problem (e.g. as in \cref{subsec:five-central-assumptions} and later in \cref{sec:deriving-a-framework}), the explicit model structure must allow us to assess whether these invariants are correctly encoded by the model.

This imposes significant restrictions on the type of model we can use.
However, we argue that this conclusion is inevitable.
In safety-critical systems, for example involving human lives, the risks and causes of model failures must be well understood, both during certification and at run-time.
Consider for example that a model failure does occur.
In order to keep operating this system, the cause of the failure must be understood, such that steps can be taken to reliably prevent a similar failure from occurring again, before redeploying this system.

Consequently, in the following sections we examine the role of self- or semi-supervised learning for certifiable models.
Then, we first investigate the question of what a ``perfectly structured model'' may look like by considering ``causal'' models, and we discuss mathematical feasibility, as well as recent advances in the research community.
Then, we explore how structure can be imposed on a model design by returning to the idea of \emph{disentanglement}, introduced in \cref{subsec:five-central-assumptions}.
Finally, we draw a connection between causal models and the disentangled setting, and subsequently construct testing procedures which can provide empirical evidence that the assumptions in \cref{subsec:five-central-assumptions} are satisfied.

\subsection{Semi-supervised representation learning for certifiable and structured models}
\label{subsec:ssl-for-certifiable-models}
\emph{Semi-supervised learning} considers uncovering structural information using only a dataset of input samples \(x\), as well as some additional labeling like structural annotations, but crucially does not use labels \(y\) during training \autocite{zhuSemiSupervisedLearningLiterature2005,chapelleSemisupervisedLearning2006,zhuIntroductionSemiSupervisedLearning2009}.
Common applications include density estimation, clustering, semantic distance prediction, or even classification and regression.
Additionally, the closely related field of \emph{representation learning} considers recovering representations that are ``useful'' in some measure, for example for multi-task prediction with unknown tasks, as a backbone to other models, or even for causal inference \autocite{bromleySignatureVerificationUsing1993,chopraLearningSimilarityMetric2005,bengioRepresentationLearningReview2013,kingmaSemisupervisedLearningDeep2014,scholkopfCausalRepresentationLearning2021a}.
For example, recently large computer vision networks have been used to learn feature representations of a large variety of input images using only annotated pairs of images, and have shown to produce representations that can be used for various ``downstream tasks'' which are not defined at training time \autocite{grillBootstrapYourOwn2020,zbontarBarlowTwinsSelfSupervised2021,bardesVICRegVarianceInvarianceCovarianceRegularization2022}.
Additionally, some research has suggested that the use of self-supervised methods, i.e. using no annotations at all, can improve model robustness and uncertainty \autocite{hendrycksUsingSelfSupervisedLearning2019}.

In this work we propose the use of semi-supervised (sometimes called weakly-supervised) models a crucial component for inherently safe design of a prediction system.
In particular, we note:
\begin{enumerate}[label=(\roman*)]
\item If the model manages to recover a representation \(z^c\) of the semantic content variables \(v^c\) in a \emph{metric space},
i.e. in a space where the distance between two \(z^c\) and \(z'^c\) is directly related to the distance between \(v^c\) and \(v'^c\), and
\item if we can additionally show that the computed embeddings \(z^c\) are \emph{disentangled} from the style variables,
i.e. that they do not contain any information except for the content,
\end{enumerate}
then it is \emph{very unlikely for the model to rely on spurious correlations} for making predictions!
This is a strong result, however severely limits the number of deep learning approaches we can use.
Nonetheless, we note that recently self- or semi-supervised approaches have managed to match performance of supervised approaches in large-scale computer vision tasks \autocite[see e.g.][for a brief discussion]{bardesVICRegVarianceInvarianceCovarianceRegularization2022}, and have been able to outperform supervised methods in most natural language tasks \autocite[see e.g.][]{brownLanguageModelsAre2020}.
Additionally considering the continuous rise of data and compute availability, we argue that the semi-supervised setting does restrictions in practice when compared to its benefits.

\subsection{Towards causal and structured models}
\label{sec:org9a70a9f}
A ``perfectly structured model'' would capture the causal relations and variables that dictate how an input is formed by finding high-level and intermediate causal variables that govern the data generating process, and deriving a connection between those and the observations.
The idea of a causal model is therefore to capture the interaction of causal variables through a \emph{directed acyclic graph} (DAG) such that the variables influence each other in a single direction, and that variables are only influenced by their parents \autocite{pearlCausality2009}.
In this setting the semantic variables ``cause'' the features in the input samples; for instance the time of the day causes certain pixels to be brighter or less bright.

Let us now denote the high-level semantic variables as \(V\) and assume they ``cause'' a set of unmodeled, or hidden, intermediate semantic variables \(H\), which in turn result in the realization of the data \(X\).
To denote this causal relationship we write \(V \rightarrow H \rightarrow X\).

\begin{figure}[t]
\centering
\includegraphics[width=\linewidth]{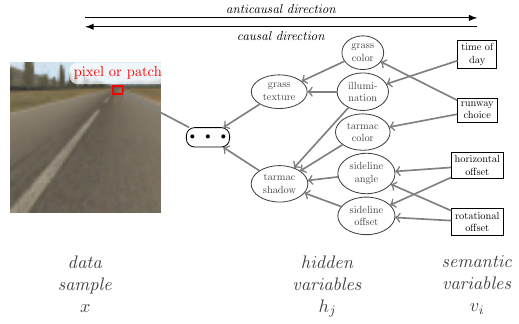}
% \includestandalone[width=\linewidth]{./tikz-pictures/causal-graph-example/causal-graph-example}
\caption{\label{fig:causal-anticausal-graph}Example of a causal structure in the image setting, indicating the \emph{causal} and \emph{anticausal} direction. Typical deep learning systems can not infer the hidden variables in symbolic form. Even if they are captured, it is not always possible to infer the direction of the causal relations, i.e. the arrows in the plot.}
\end{figure}

To formalize this setting we assume a set of random hidden semantic variables \(h_i\) such that
\begin{equation}h_i = f_i(\mathit{Pa}(h_i), u_i)
\end{equation}
for some functions \(f_i\) and noise realizations \(u_i\), where \(\mathit{Pa}(h_i)\) denotes the ``parent'' nodes of \(h_i\) as determined by the DAG.
In other words, the variable \(h_i\) is only dependent on its parent variables \(\mathit{Pa}(h_i)\) though the function \(f_i\), together with some amount of modeled or umodeled noise \(u_i\).
Such a model is called a \emph{Structured Causal Model} (SCM) \autocite{pearlCausality2009,scholkopfCausalRepresentationLearning2021a}.
In an SCM, we can therefore write
\begin{equation}p(x, h, v) = p(x \mid h) \prod_{i} p(h_i \mid \mathit{Pa}(h_i)) \prod_j p(v_j)
\label{eq:SCM}
\end{equation}
where \(\mathit{Pa}(h_j) \subset \left\{ v_1, v_2, \dots \right\} \cup \left\{ h_1, h_2, \dots \right\}\).
See \cref{fig:causal-anticausal-graph} for an illustration inspired by our proposed use case.

\begin{figure}[t]
\centering
\includegraphics[width=0.75\linewidth]{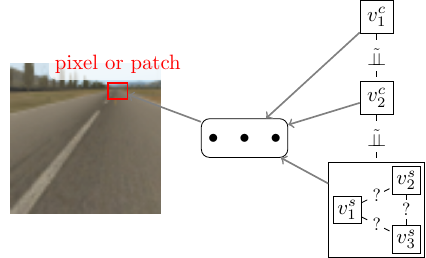}
% \includestandalone[width=0.75\linewidth]{./tikz-pictures/causal-graph-simple/causal-graph-simple}
\caption{\label{fig:causal-graph-simple}A simple disentangled structure in the image setting. The content variables are disentangled mutually and w.r.t. the style variables. The style variables have an unknown entanglement structure.}
\end{figure}

A crucial question is now whether a model can recover an SCM defined by the DAG together with the set of functions \(f_i\) in a self-supervised matter, i.e. using only access to inputs \(x\); in other words, whether it would be sufficient to take a large collection \({\bf x} = \left\{ x_1, x_2, \dots \right\}\) of diverse input samples such that the model finds the semantic variables \(v\), and hidden variables \(h\) and their relations to \(x\) automatically.

This task is called \emph{causal discovery} and in the next sections we briefly summarize results for the ``full'' causal discovery, as well as for a related problem where only the high-level semantic variables \(v\) are to be recovered.
Then, we recall a practical model framework that aims to respect some of the causal properties, but does require some prior knowledge about the semantic variables.
Afterwards we return to the assumptions stated in \cref{subsec:five-central-assumptions} and show how this framework can be used to empirically validate those assumptions.
Finally we recall a set of ten desiderata for a ``good'' representation, originally introduced by \textcite{bengioRepresentationLearningReview2013}, and compare them to the proposed framework.

\subsubsection{Causal discovery}
\label{sec:orgc9469b3}
Suppose now that we have observations of \(v\) and \(x\), and that the causal structure is \(V \rightarrow H \rightarrow X\).
Can a model find the hidden semantic variables \(h_i\) together with the causal relations \(f_i\) to compute \(p(x \mid v) = p(x \mid h) p(h \mid v) p(v)\) purely through self-supervised training, where a \emph{semantic variable} can refer to being \emph{robust}, \emph{transferable}, \emph{interpretable}, \emph{explainable}, or \emph{fair}?
In recent years, multiple works have come forwards towards these goals in low-dimensional domains \autocite{chalupkaMultiLevelCauseEffectSystems2016,rubensteinCausalConsistencyStructural2017,kilbertusAvoidingDiscriminationCausal2017,kusnerCounterfactualFairness2017,zhangFairnessDecisionMakingCausal2018}.
However, it is still challenging to state precisely under what conditions such a ``semantic variable discovery'' would be feasible.
Some work has been done in this direction by putting strong restrictions on the functions \(f_i\), with a common choice being linear-Gaussian models \autocite{pearlCausality2009,ScholkopfJPSZMJ2012,petersCausalDiscoveryContinuous2014,lopez-pazLearningTheoryCauseEffect2015,parascandoloLearningIndependentCausal2018a,cundyBCDNetsScalable2021}.
However, we consider this too restrictive for our application.
Additionally we note that, even if the hidden variables were observed, it turns out that already the simpler problem of simply recovering the \emph{causal directions} between the hidden variables is impossible in the general setting \autocite{pearlCausality2009,ScholkopfJPSZMJ2012,scholkopfCausalRepresentationLearning2021}.

We therefore conclude that a factorization of the data in the \emph{causal direction}, i.e.
\begin{equation} p(x, h, v) = p(x \mid h) p(h \mid v) p(v)
\end{equation}
does still seem unattainable in a high-dimensional setting, e.g. for computer vision problems.
However, theory suggests that this factorization may be possible in the \emph{anticausal} direction, i.e.
\begin{equation} p(x, h, v) = p(v \mid h) p(h \mid x) p(x),
\end{equation}
that we can compute \(p(v \mid x)\) \autocite{ScholkopfJPSZMJ2012,kilbertusGeneralizationAnticausalLearning2018}.
This is great news, as this is exactly the direction we want to infer, since the input to our prediction model \(x\) is the output of the causal generative process \(g^\ast(v)\).

Indeed, in recent years some work has been done to infer anticausal relationships and recover some semantic features specifically in images \autocite{lopez-pazDiscoveringCausalSignals2017,sauerCounterfactualGenerativeNetworks2021}, though this direction has yielded limited results as of yet.
We note however, that in 2021 a new Workshop for Causality in Vision has been introduced at the European Conference of Computer Vision, and we are looking forward to further progress in this area in the coming years.

We conclude that, unless further progress is made, the recovery of hidden semantic variables and the causal or anticausal relationships of these variables does not seem possible for non-trivial cases.

\subsubsection{Self-supervised Recovery of Disentangled Variables}
\label{subsubsec:self-supervised-recovery-of-disentangled-vars}
Next, we revisit the question whether only the semantic variables \(v\) can be recovered purely from observations \(x\); i.e. instead of recovering hidden intermediate variables, we are only interested in a set of high-level variables that sit at the ``root'' of the causal graph.
In order to formalize this notion, we first recall the definition of \emph{disentanglement} provided in \cref{subsec:five-central-assumptions}.
Note that in the language of SCMs (\cref{eq:SCM}) this implies that semantic factors do not have a parent node, i.e. are at the root of the causal graph, and are mutually conditionally independent.

Several works have investigated the possibility of recovering disentangled semantic variables in the purely self-supervised setting and have put forth a number of approaches  \autocite{higginsBetaVAELearningBasic2017,kimDisentanglingFactorising2018a,shuWeaklySupervisedDisentanglement2019,hosoyaGroupbasedLearningDisentangled2019,ridgewayLearningDeepDisentangled2018} .
Surprisingly though, \textcite{locatelloChallengingCommonAssumptions2019} published an important paper proving that, in the general setting, unsupervised discovery of disentangled variables is fundamentally impossible.
Instead, in a follow up paper \textcite{locatelloWeaklySupervisedDisentanglementCompromises2020} presented an approach that learns a mapping \(x \mapsto v\), but requires additional knowledge about the semantic features \(v\).
In particular, for each semantic feature \(v_i\) that should be disentangled, the method requires pairs of input samples for which a known set of semantic features change, but that otherwise stay constant.\footnote{In fact, the authors even show some success by only providing the number of indices, i.e. \(\mathit{length}\left(\left\{ i, j, \dots \right\}\right)\), for each pair of inputs. But for our setup it is sufficient to consider knowing which semantic feature changes.}
Subsequently, \textcite{dittadiTransferDisentangledRepresentations2021} showed that this approach can be used successfully in a real-world setting using images from a real camera.

\subsection{Introducing a structured model architecture}
\label{subsec:a-simple-mathematical-model}
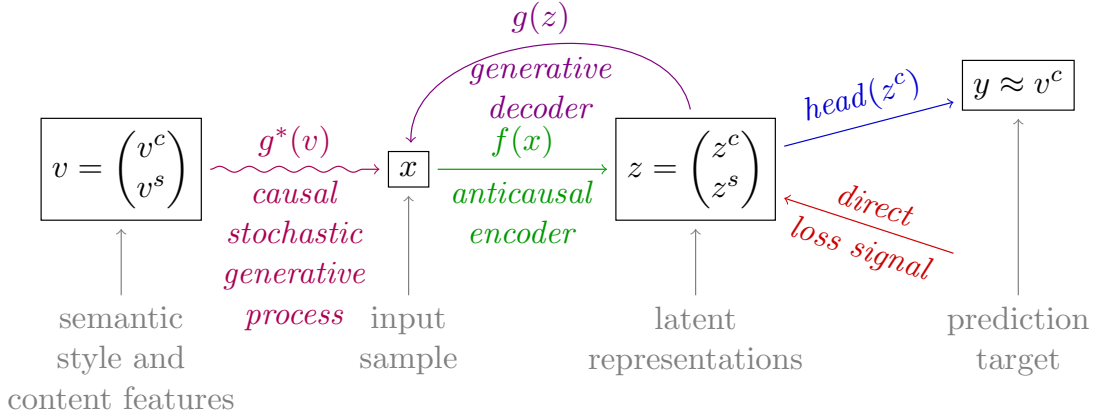
\begin{figure}[t]
\centering
\begin{tikzpicture}[mynode/.style={
    draw, outer sep=3pt,% minimum height=1.0cm, minimum width=1.0cm
  }]
  \node[mynode] (vs) {$v = \begin{pmatrix} v^c \\ v^s \end{pmatrix}$};
  \node[mynode, right=2 of vs] (x) {$x$};
  \draw[->, decorate, decoration={snake, amplitude=1pt, post length=2pt}, color=purple!90!blue] (vs) to node[below, align=center]{\emph{causal} \\ \emph{stochastic} \\ \emph{generative} \\ \emph{process}} node[above]{$g^\ast(v)$} (x);
  \node[mynode, right=2 of x] (zs) {$z = \begin{pmatrix} z^c \\ z^s \end{pmatrix}$};
  \draw[->, color=green!60!black] (x) to[bend right=0] node[above]{$f(x)$} node[below, align=center]{\emph{anticausal} \\ \emph{encoder}} (zs);
  % \node[mynode, right=2 of zs, yshift=-1.0cm] (x-tilde) {$\tilde{x}$};
  \coordinate[right=2 of zs, yshift=-1.0cm] (x-tilde);
  \draw[<-, color=red!80!black] (zs) to node[above, sloped, align=center]{\emph{direct}} node[below, sloped, align=center]{\emph{loss signal}}  (x-tilde);
  \node[mynode, right=2 of zs, yshift=+1.0cm] (ys) {$y \approx v^c$};
  \draw[->, color=blue!80!black] (zs) to node[above, sloped]{$\mathit{head}(z^c)$} (ys);

  \coordinate (annot-line) at (0, -1.5cm);
  \draw[<-, gray] (vs) to (vs |- annot-line) node[anchor=north, align=center, text width=3cm]{semantic  style and  content  features};
  \draw[<-, gray] (x) to (x |- annot-line) node[anchor=north, align=center]{input \\ sample};
  \draw[<-, gray] (zs) to (zs |- annot-line) node[anchor=north, align=center]{latent \\ representations};
  % \draw[<-, gray] (x-tilde) to (x-tilde |- annot-line) node[anchor=north, align=center]{input \\ reconstruction};
  \draw[<-, gray] (ys) to (ys |- annot-line) node[anchor=north, align=center]{prediction \\ target};

  \draw[->, red!50!blue] (zs) to[bend right=85] node[above] {$g(z)$} node[below, align=center, scale=1.0] {\emph{generative} \\ \emph{decoder}} (x);

\end{tikzpicture}
\caption{\label{fig:assumption-flow}Relation of semantic features \(v\), input sample \(x\), latent features \(z\), and predictions \(y\) and \(\tilde{x}\) to the ``true'' generator \(g^{\ast}\), the encoder, decoder and head.}
\end{figure}

Before proceedings, we will now propose an abstract model structure that can satisfy the inherently safe design requirements discussed in the previous sections.
A concrete instantiation of this structure will be discussed in \cref{sec:proposed-model-and-dataset}.

The goal is that a model with the proposed structure is able to recover a disentangled representation of the content features \(z^c\), which correctly encodes the distance between two \(z^c\).
A module computing the representation is constructed through semi-supervised training, following the discussion in \cref{subsec:ssl-for-certifiable-models}, for example by using a variational autoencoder (VAE) \autocite{kingmaAutoEncodingVariationalBayes2014} or a generative adversarial network (GAN) \autocite{goodfellowGenerativeAdversarialNets2014}, together with a well-suited loss function.

As introduced in \cref{sec:fundamental-assumptions}, we assume that an input \(x\) is generated through a stochastic and unknown generative process \(g^\ast\) which relies on a hidden set of semantic content and style features \(v = (v^c, v^s)\), i.e.
\begin{equation}
x = g^\ast(v).
\end{equation}
Then, an encoder in the anticausal direction, denoted \(f(x)\), computes the content and style representations \(z = (z^c, z^s)\), and a decoder in the causal direction, denoted \(g(z)\), tries to recover a reconstruction \(x^{\rm rec}\) of the input \(x\) from the latent representations, i.e.
\begin{equation}
\begin{aligned}
z &= f(x) \\
x^{\rm rec} &= g(z).
\end{aligned}
\end{equation}
Together, the encoder and decoder form the generative model, i.e. the VAE or GAN, which is trained without considering the outputs \(y\).

Then, in a second step, the ``model head'' is constructed, which uses only the content representations to make the final regression prediction.
Notably, the model head should be a simple function, preferably linear or almost linear, and may be trained in a supervised way, i.e. using labels \(y^{\rm label}\).
However, in \cref{sec:proposed-model-and-dataset} we will see that it is possible to construct the model head without any labels, using information about the operating parameters and a simple linear model using a transformed version of the content representations.
\cref{fig:assumption-flow} illustrates the proposed model structure.

\subsection{Disentanglement}
\label{subsubsec:disentanglement}
Assumptions 2 and 3 in \cref{subsec:five-central-assumptions} state that the content variables are disentangled from each other, and that the content is disentangled from the style.
In the following, we therefore review several methods to measure disentanglement.
We then propose a simple new method to verify our assumptions by fitting a simple linear model from \(z\) to \(y\) and using t-testing to assert independence between variables.

\subsubsection{Measuring disentanglement}
\label{subsubsec:testing-for-disentanglement}
Multiple criteria to quantify the ``amount of disentanglement'' have been proposed  \autocite{bengioRepresentationLearningReview2013,higginsBetaVAELearningBasic2017,kimDisentanglingFactorising2018a,eastwoodFrameworkQuantitativeEvaluation2018,ridgewayLearningDeepDisentangled2018} .
We adapt the terminology of \citeauthor*{ridgewayLearningDeepDisentangled2018}, who specify the three notions of \emph{modularity}, \emph{compactness}, and \emph{explicitness}.
The underlying assumption is that a data point is described by a ground-truth set of latent factors \(v_k\), which we try to reconstruct with a set of latent encodings \(z_j\).
Then, the three terms can roughly be described as
\begin{description}
\item[{Modularity}] Each encoding dimension corresponds to a single ground truth factor, i.e. \(\forall j\ \exists k\ s.t.\ z_j \rightarrow v_k\).
\item[{Compactness}] Each ground truth factor is encoded by one or a few latent encodings, i.e. there is a small index set \(\mathcal{J}, j \in \mathcal{J}\) s.t. \(v_k \rightarrow z_{\mathcal{J}}\).
\item[{Explicitness}] There exists a linear relationship between \(z_j\) and \(v_k\), i.e. \(\theta^\intercal z_j + b = v_k\).
\end{description}

We focus on the particular case that there exists a 1-to-1 linear relationship between the ground truth factors and the latent encodings.
For instance, in a pose estimation problem this could correspond to a 6-degree-of-freedom description of the state which corresponds to six disentangled latent encodings.

In the following, we will first construct a test showing a 1-to-1 correspondence of the latent encodings to the ground truth factors by using hypothesis testing, specifically the Student's t-test.
To this end, we pick a particular ground truth factor and a significance level of, for example, \(5\%\).
Then, we train a linear model mapping the latent encodings to the ground truth factor and formulate the Null-hypothesis such that it states that the linear factors are normally distributed around zero.
For each estimated coefficient \(\hat \beta_j\) we can then compute the t-test statistic
\begin{equation}T_j = \frac{\hat \beta_j}{\sqrt{\hat\sigma^2(X^\intercal X)^{-1}_{jj}}}\end{equation}
and can compute the corresponding p-values as
\begin{equation}p_j = \mathbb{P}(T_j \sim t_{n-p}).\end{equation}

Using the p-values, we can now assert that each ground truth factor \(v_k\) is only explained by a single latent encoding (plus the bias term), and that the explanatory latent encodings are all distinct, i.e. no encoding is used for multiple ground truth factors.

\begin{listing}[htbp]
\caption{\label{alg:test-1-to-1}\texttt{function test\_1\_to\_1\_mapping}. We fit a linear model to each ground truth content factor and assert that only a single content representation has predictive power.}
\begin{Code}
\begin{Verbatim}
\color{EFD}\EFk{function} \EFf{test\_1\_to\_1\_mapping}(model, data; significance\_level=\EFhn{0.05})
    zs = embeddings = model.encoder.(data.Xs) .\EFt{|>} z\_\EFt{->}z\_[\textcolor[HTML]{008b8b}{:mu}]
    used\_latent\_dims = []
    \EFk{for} v\_k \EFt{::} \EFt{Vector} \EFk{in} data.vs[\textcolor[HTML]{008b8b}{:content}]
        lm = fit(LinearModel, \textcolor[HTML]{483d8b}{\textbf{@formula}}(v\_k \EFt{\char126{}} zs \EFt{+} \EFhn{1}))
        \textcolor[HTML]{483d8b}{\textbf{@assert}} sum(pvalues(lm) .\EFt{<} significance\_level) \EFt{==} \EFhn{2} \EFcd{\# }\EFc{beta and bias}
        push!(used\_latents,
              firstindex(pvalues(lm) .\EFt{<} significance\_level))
    \EFk{end}
    \textcolor[HTML]{483d8b}{\textbf{@assert}} allunique(used\_latent\_dims)
\EFk{end}
\end{Verbatim}
\end{Code}
\end{listing}

The test in \cref{alg:test-1-to-1} shows that each latent content encoding only contains information about one of the ground truth content factors, which satisfies the \emph{modularity} and \emph{compactness} conditions.
Next we want to verify that the latent content encodings additionally don't contain any information about the style.
We can construct a similar test, this time predicting ground truth factors of variation that are \emph{not part of the content}, i.e. that are style features.
For instance, we could use the time of day or background features.
If we have labels for these, we can again construct a linear model for each style feature and compute the t-test statistic.
This time we assert that each coefficient for the latent content encodings \emph{accepts the Null-hypothesis}.
The corresponding code example is provided in \cref{alg:test-style-content-separation}.

\begin{listing}[htbp]
\caption{\label{alg:test-style-content-separation}\texttt{function test\_content\_style\_separation}. We fit a linear model to each ground truth style variable and assert that no latent content encoding has significant predictive power.}
\begin{Code}
\begin{Verbatim}
\color{EFD}\EFcd{\# }\EFc{k content variables}
\EFcd{\# }\EFc{l style variables}
\EFk{function} \EFf{test\_content\_style\_separation}(encoder, xs, vs\_c, vs\_s;
                                       significance\_level=\EFhn{0.05})
    zs\_c, zs\_s = encoder.(xs)
    \EFk{for} i \EFk{in} \EFhn{1}:l
        vs\_s\_i = getindex.(vs\_s, i)
        lm = fit(LinearModel, \textcolor[HTML]{483d8b}{\textbf{@formula}}(vs\_s\_i \EFt{\char126{}} zs\_c \EFt{+} zs\_s \EFt{+} \EFhn{1}))
        \textcolor[HTML]{483d8b}{\textbf{@assert}} all(pvalues(lm)[\EFhn{1}:k] .\EFt{<} significance\_level)
    \EFk{end}
\EFk{end}
\end{Verbatim}
\end{Code}
\end{listing}
\subsection{Priors for a ``good'' representations}
\label{sec:org464d91e}
In their seminal paper, \textcite[, Section 3.1]{bengioRepresentationLearningReview2013} define a list of ten priors that learned representations should fulfill in order to be ``useful''.
We will consider these priors as a list of \emph{desiderata}, which gives us the opportunity to validate whether the disentanglement framework introduced in the previous section is conceptionally sound w.r.t. these priors.

In the following, we briefly recall each condition and discuss to what extent they are applicable to our problem setting, and how the proposed prediction framework addresses each prior.
Where applicable, we introduce principles and empirical evidence to validate how well each desideratum is satisfied.

\paragraph{Desideratum 1: Smoothness of \(f(x)\).}
\label{sec:org4b29bbb}
\emph{For two inputs \(x\) and \(x'\) that are similar, i.e. \(x \approx x'\), the corresponding function outputs should be similar as well, i.e. \(f(x) \approx f(x')\).}

While the disentangled setup does not directly enforce smoothness in the mapping from \(x\) to \(z\), it does enforce smoothness in the mapping \(z = f(g^\ast(v))\).
If we therefore assume that \(g^\ast\) has sufficient smoothness properties and \(z^c\) is related to \(v^c\) in a smooth and monotonous way, we can deduce that the encoder \(f\) must also be smooth for the content variables.

Considering now the runway use case, it makes sense that the image may not be a smooth function of the style variables, as there may be large discontinuities when changing for instance the location of the runway.
For the content variables however we assume that a small change in \(v^c\) will result in a small change in \(x\) and therefore a small change in \(z\) and \(y\), which is consistent with the claim above.

To gain empirical evidence of the smoothness property we can leverage the generative model introduced in \cref{subsec:a-simple-mathematical-model}.
Using the generative model, we can manually traverse the semantic latent space \(z\) and inspect the generated images \(x_{\rm rec} = g(z)\).
We expect to see smooth transitions, especially for the content variables, as long as they are within the specified operating parameters.

We note that, while this does tell us something about the relationship of the latent space and the decoder, this does not directly investigate smoothness properties of the encoder \(f\), as it is not involved in the process.
Therefore we propose chaining the encoder and decoder together, and comparing \(z\) with \(f(g(z))\).
This has the added benefit that we can use this to automate the empirical validation process.

\paragraph{Desideratum 2: Multiple explanatory factors.}
\label{sec:org81c5c71}
\emph{``The data generating distribution is generated by different underlying factors, and for the most part what one learns about one factor generalizes in many configurations of the other factors. The objective [is] to recover or at least disentangle these underlying factors of variation''.}

Naturally, this desideratum is closely related to our assumptions (\cref{subsec:five-central-assumptions}) and discussion about disentangled representations (\cref{subsubsec:disentanglement}).
We therefore argue that our setting fulfills this desideratum by construction and refer to the previous section for discussion.

\paragraph{Desideratum 3: Hierachical organization of factors.}
\label{sec:org1f6ed38}
\emph{Explanatory factors should be arranged in a /hierarchy} ranging from abstract to concrete concepts.
More concrete concepts can be used to refine the description of a sample, whereas abstract concepts describe high-level properties./

We argue that this hierarchy is equivalent to specifying the first few layers of the causal graph.
Unfortunately, as denoted previously, currently no methods are known which are able to discover these layers automatically.
On the other hand, for manually specified causal relations it is hard to prove that they capture all necessary relations of the true generative process.
We therefore argue that argue that in the heavily restricted setting that we are working in it is sufficient to obtain the high-level semantic variables without modeling the causal relationships explicitly.
We do note, however, that this desideratum may be necessary for certification of models in less restricted settings and are looking forward to seeing new approaches to causal discovery emerge in the coming years.

\paragraph{Desideratum 4: Semi-supervised learning.}
\label{sec:orgff66657}
\emph{Factors which explain \(p(x)\) and can be computed in a self-supervised fashion should also help compute \(p(y \mid x)\), therefore making a sort of model transfer from a self-supervised to a supervised setting possible.}

This is consistent with our problem formulation in which we learn an approximation \(z\) of the factors \(v\) which ``cause'' \(p(x)\).
Since our output \(y\) is chosen to be equal to a subset of \(v\) we can therefore naturally write \(p(y \mid x) = p(y \mid v) p(v \mid x)\), where \(p(y \mid v)\) is learned in a supervised manner and \(p(v \mid x)\) is learned in a weakly-supervised manner as previously described.

\paragraph{Desideratum 5: Shared factors across tasks.}
\label{sec:orgd5d4816}
\emph{Factors which are recovered by the model can be used not only to address the original task \(p(y \mid x)\), but also for other tasks.}

Since we specifically construct the latent space such that only the content variables are directly related to the tasks, we do not expect our proposed setup to fulfill this desideratum better than any other model building a representation.
Of course, one can use the representation of the style variables to predict other values, for instance, the time season or date.
However, the disentangled structure is lost for the style variables, which invalidates most of the benefits that our approach has.

\paragraph{Desideratum 6: Manifolds.}
\label{sec:orgfdc4ee9}
\emph{The concentration of probability mass of the input samples is located on a much smaller manifold than the data itself.}

This is naturally given in our approach, since we compress a high-dimensional input space into a comparatively small latent representation.

\paragraph{Desideratum 7: Natural clustering.}
\label{sec:orgf2e54af}
\emph{Data points are clustered in the representation space, where clusters can be determined by categorical or discrete variables.
A linear interpolation between such clusters does not always produce sensible results.}

Since we assume our content variables to lie on a single consecutive interval each, we associate them with only a single cluster.
The style variables however may need to encode discrete values, and clustering is therefore expected.

\paragraph{Desideratum 8: Temporal and spatial coherence.}
\label{sec:orgfda947d}
\emph{Factors observed consecutive in time, or spatially related, should be associated with a small movement on a high-density manifold.}

We expect this desideratum to be very practical in our problem setting.
Since the predicted variables represent the state of the system, we expect that an upper bound on the magnitude of change in the content variables can be directly derived from a system model for many systems.
For instance, in the case of an airplane an estimate of the airplanes velocity can be used, and individual bounds for the different content variables can be computed analytically.
For the style variables on the other hand we expect very small changes over time, although discontinuous changes may occur if features come in- or out-of-scope.

\paragraph{Desideratum 9: Sparsity of representations.}
\label{sec:orgf32cba5}
\emph{For an input \(x\), only some of the features \(v\) are ``active'', and many are equal to zero.}

Although this may be true for the style variables, if their dimensionality is large enough, the content variables are dense by definition.
Still, one could imaging a different problem setting where not all content variables are always defined.
In that case, we would have to define how a ``missing'' content variable should be encoded, since in our setting ``zero'' may just represent a regular value.

\paragraph{Desideratum 10: Simple factor dependencies.}
\label{sec:org98d9df7}
\emph{If the representations are sufficiently high-level, they should have a simple dependency to the true labels.}

We will see that this is precisely the case for our content representations, requiring only a simple probability density transformation, followed by a single linear transformation to make the regression predictions.

\newpage
\section{Run-time error detection}
\label{subsec:runtime-error-detection}
In this section, we first explore how uncertainty in the predictions can be dealt with through the use of explicit \emph{uncertainty quantification}.
We provide a set of desiderata that model outputs should fulfill in order to be ``correct'' and show how they can be validated.
Then, we discuss the topic of \emph{out-of-distribution detection}, i.e. the task to reject inputs for which the model can not generalize.
Afterwards, we discuss the concept of \emph{feature collapse}, i.e. the problem that a model may represent a novel input with the same representation as a known one.
Finally, we discuss whether the discussed models are susceptible to \emph{adversarial attacks} and what should be done to mitigate them.

\subsection{Causes of ambiguity}
\label{sec:org63e54a8}
\begin{figure}[t]
\centering
\includegraphics[width=\textwidth]{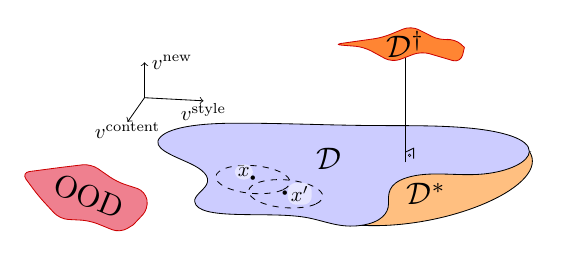}
% \includestandalone[width=\textwidth]{./tikz-pictures/distributions2/distributions}
\caption{\label{fig:distributions}An illustration of different distributions.}
\end{figure}
Before we begin with uncertainty estimation and out-of-distribution detection, we present a brief illustration why we need these methods in the first place.
To that end, \cref{fig:distributions} gives an example of data distributions we are dealing with.
\(\mathcal{D}_{\rm train}\) denotes the main training distribution, i.e. all data points that are collected, and is later split into a train, validation, calibration and test set.
We expect to perform well on this set.
\(\mathcal{D}^{\ast}\) denotes datapoints that contain combinations of features which are all individually included in \(\mathcal{D}\), but occur in a new combination.
In a strict sense these points could be considered ``out-of-distribution'', but we can hope our model to generalize to these points well and give good predictions.
Another distribution pictured is \(\mathcal{D}^{\dag}\), which generally consists of similar features to those in \(\mathcal{D}\) and \(\mathcal{D}\), but has additional new (style) features that are not included in the training dataset.
Finally, some ``true'' out-of-distribution points are pictured, which may have nothing in common with the original distribution.

In addition to the distributions, the illustration also shows the possible ambiguity in the relation \(v \leftrightarrow x\).
Illustrated by the dashed ellipses the figure shows all semantic feature variables \(v\) that can satisfy \(v = f^\ast(g^\ast(v))\) with non-zero likelihood for a ``perfect'' model \(f^\ast: x \rightarrow v\) and the true generating function \(g^ast: v \rightarrow x\).
A consequence of this is that for any point \(x = g^{\ast}(v)\) in the intersection of the two ellipses there is no single variable \(v\) that can be predicted.

\subsection{Uncertainty quantification}
\label{subsec:uncertainty-quantification}
In many engineering applications, decisions need to be made based on an estimate of the current state, together with an associated uncertainty.
Especially in safety-critical domains, the cost of failure needs to be weighed against the likelihood of making an erroneous prediction.
For example, in our runway use case the plane controller needs to decide in time whether to complete the approach or abort.
If the chance of transitioning into an unrecoverable state is too high (or even marginally above zero), the model must reflect this in the state estimation.
In other words, a model must assign a sufficient amount of ``probability mass'' to all events that may occur, even if they are unlikely.

Whether for filtering or when directly using the prediction, it is important that the quality of the uncertainty estimation is well understood.
More specifically, even bad predictions can be useful if the probability distribution of the errors is well understood.
Conversely, if the model is over-confident about a prediction that is critical to the system's safety, it can lead to catastrophic results.

In the following, we therefore explore different aspects of uncertainty and how the quality of a prediction with associated uncertainty can be evaluated.
To this end we recall, among others, the notions of calibration and dispersion.
Then, we discuss different qualitative and quantitative ways to evaluate calibration in both the \emph{marginal} and \emph{conditional} sense.

\subsubsection{Separating risk from uncertainty}
\label{sec:orga7d39b7}
\textcite{knightRiskUncertaintyProfit1921} separates \emph{risk} from \emph{uncertainty} in the sense that uncertainty is something we can quantify with some confidence.
For example, we can predict the sum of ten consecutive dice throws with a fair dice and simultaneously quantify the quality of our prediction, i.e. how often we are going to be correct, off by one, etc.
This is uncertainty after Knight's definition.
Risk on the other hand occurs if we can not even judge our level of uncertainty, for example when predicting the sum of ten dice throws with an unknown weighted dice.
In this case, our model (the fair dice) does not correspond well to the physical process, and therefore can not give good uncertainty estimates.
Still, we might at least be aware that our model of a fair dice is not suitable for prediction, and reject the input all together.

In this section we discuss \emph{quantifiable uncertainty}, or \emph{uncertainty quantification} (UQ), and in \cref{subsec:OOD-detection} pick up on Knight's notion of risk (confusingly sometimes called \emph{Knightian uncertainty}) and its connection to out-of-distribution (OOD) detection.

\subsubsection{Sharpness and calibration}
\label{sec:orgec86aa5}
\begin{figure}[t]
\centering
\includegraphics[width=0.4\textwidth]{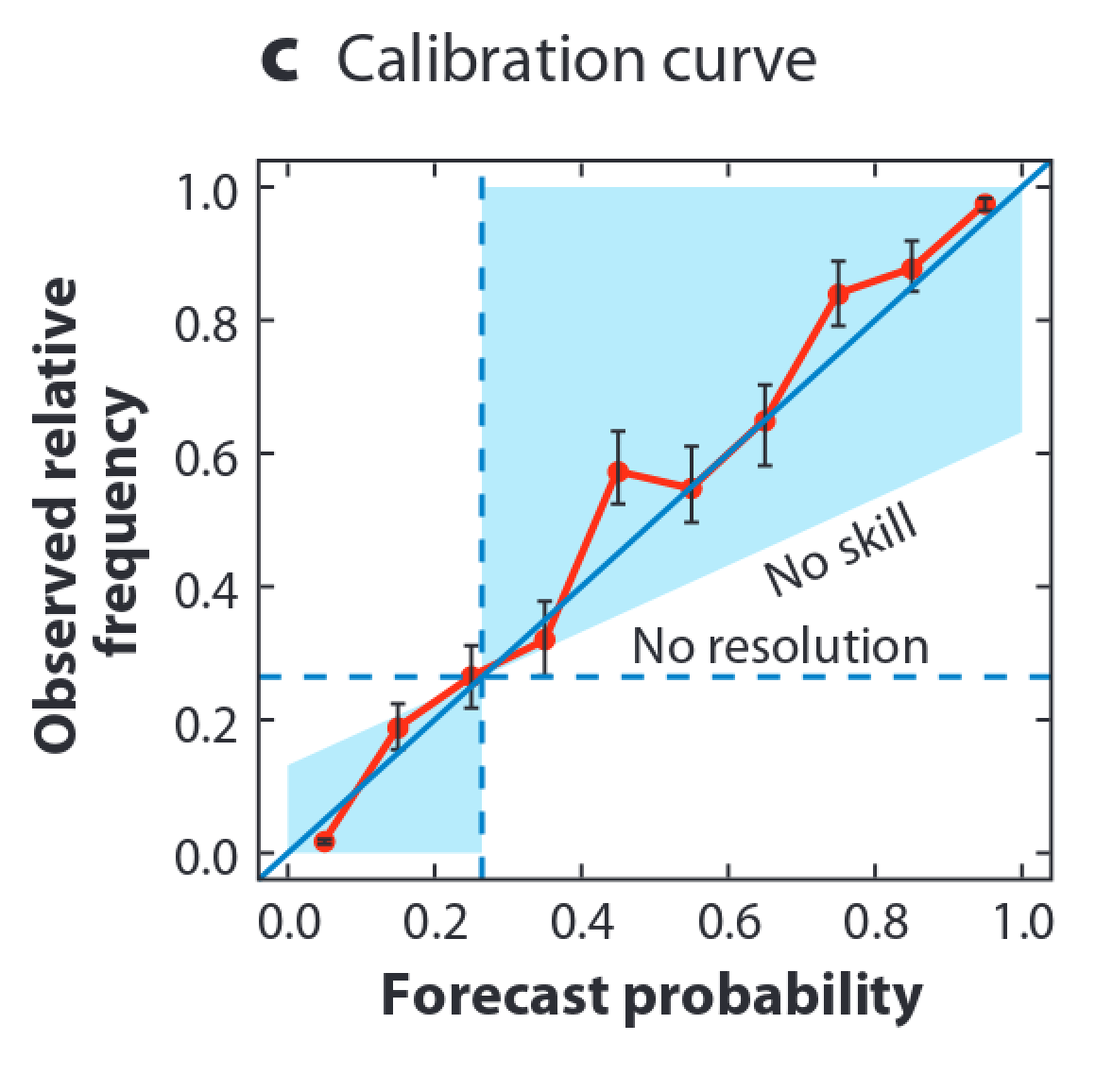}
\caption{\label{fig:binary-calibration-plot}Calibration curve of a binary wind speed prediction task, reprinted from \textcite[, Figure 3c.]{gneitingProbabilisticForecasting2014}.}
\end{figure}

Fundamentally, we want uncertainty predictions to be \emph{sharp} and \emph{well-calibrated}.
Following \textcite{gneitingProbabilisticForecasting2014}, we recall the definitions or \emph{calibration} and \emph{sharpness} as follows\footnote{See also \textcite{gneitingProbabilisticForecastsCalibration2007} for an earlier, but less concise definition by the same (first) author.}:
\begin{description}
\item[{Calibration}] ``[the] statistical compatibility of probabilistic forecasts and observations; essentially, realizations should be indistinguishable from random draws from predictive distributions'', and
\item[{Sharpness}] ``the concentration of the predictive distributions in absolute terms; a property exclusive to the forecasts''.
\end{description}
Additionally, we recall a measure related to sharpness, namely
\begin{description}
\item[{Dispersion}] ``the concentration of the predictive distributions relative to the observations; a joint property of forecasts and observations.''
\end{description}

\textcite{gneitingProbabilisticForecastsCalibration2007} introduce three different \emph{modes} of calibration in the context of probability forecasting.
They denote by \((G_{t})_{t=1,2,\dots}\) the ``ground truth'' cumulative density function (CDF) of the forecast at time \(t\), and by \((F_{t})_{t=1,2,\dots}\) the CDF of the estimated forecast, and define
\begin{description}
\item[{marginal calibration}] as
\begin{equation}
  \lim_{T \to \infty} \frac{1}{T} \sum_{t=1}^T G_t(x) = \lim_{T \to \infty} \frac{1}{T} \sum_{t=1}^T F_t(x) \qquad \text{for all } x \in \mathbb{R},
\end{equation}
\item[{probabilistic calibration}] as
\begin{equation}
  \frac{1}{T} \sum_{t=1}^T [G_t \circ F_t^{-1}] (p) \qquad \text{ for all } p \in (0, 1),
\end{equation}
and
\item[{exceedance calibration}] as
\begin{equation}
  \frac{1}{T} \sum_{t=1}^T [G_t^{-1} \circ F_t] (x) \qquad \text{for all } x \in \mathbb{R}.
\end{equation}
\end{description}
A concrete example together with an illustration is provided in \cref{sec:calibration-illustration}.

The authors show by example that in principle all combinations of being calibrated are possible, i.e. a forecast \(F_t(x)\) can be marginally and exceedance calibrated, but not probabilistically calibrated and so forth.
Still, for many types of forecasting, probabilistic and marginal calibration are equivalent \autocite[, part 2.4]{gneitingProbabilisticForecastsCalibration2007}.
Exceedance calibration on the other hand is usually a much stronger condition, as it requires conditioning on \(x\), and, adopting a more modern terminology, we will refer to \emph{exceedance calibration} as \emph{conditional calibration}.

\subsubsection{Evaluating marginal calibration}
\label{sec:org97d8416}
\paragraph{The binary setting.}
\label{sec:org12b3f18}
For a binary event \(y \in \{\top, \bot\}\), let a model \(f(x)\) predict the probability \(p=f(x)\) of the event coming true.
Then, we can understand marginal calibration of \(f\) as follows:
For any choice of \(p \in (0, 1)\), consider all inputs \(x\) for which \(f(x) = p\).
Under marginal calibration, for those \(x\) we expect the observed frequency of \(y = \top\) to equal \(p\).
Mathematically we can write
\begin{equation} \underset{\{(x,y) \mid f(x)=p\}}{\mathbb{P}}
\label{eq:calibration-eq}
\left[y = \top\right] \approx p \qquad \text{for all } p \in (0,1),
\end{equation}
where we compute the expectation over all \((x,y)\) tuples for which the model predicts probability \(p\), and \(\mathbbm{1}[\cdot]\) denotes the Iverson bracket.

 \cref{fig:binary-calibration-plot} presents a visualization of this property, where a model predicts whether wind speeds at a sensor station will exceed \(10 \mathrm{m}/\mathrm{s}\) in the following two hours \autocite{gneitingProbabilisticForecasting2014}.
The observed relative frequency of the event coming true is plotted against binned probability predictions by the model.
Confidence intervals for the observed relative frequency have been obtained through bootstrapping.

We will now expand this idea to the safety-critical case.
Let the binary value of the event denote whether the event is \emph{safe} (\(y=\top\)) or \emph{unsafe} (\(y=\bot\)).
Given the safety-critical setting, we assume that \emph{false positive} predictions incur a large safety risk, whereas \emph{false negative} predictions are tolerated.
Consequentially, when predicting an event to be \emph{safe} with probability \(p\), we claim that the observed frequency of the event being safe should be at high, or higher, than \(p\).
Mathematically, we can therefore modify \cref{eq:calibration-eq} to
\begin{equation} \underset{\{(x,y) \mid f(x)=p\}}{\mathbb{P}} \left[y = \top\right] \geq p \qquad \text{for all } p \in (0, 1)
\label{eq:calibration-ineq}
\end{equation}
for the safety-critical case and binary events.
In \cref{fig:binary-calibration-plot} this would imply that, for all \(p\), the red curve must consistently lie on or above the blue line.

\paragraph{The regression setting.}
\label{sec:org00711d9}
Let us now go beyond the binary case and into a regression setting.
We first introduce a suitable definition of calibration for this case, similar to \cref{eq:calibration-eq}, and then consider again the safety-critical case, as with \cref{eq:calibration-ineq}.
Assume that, for a given input \(x\), ``nature'' draws the true outcome \(y\) from an unknown distribution.
Our model \(f(x)\) then tries to match this outcome distribution with a predicted distribution \(D\), defined through a cumulative density function (CDF).

\begin{figure}[t]
\centering
\begin{subfigure}[t]{0.30\textwidth}
\includegraphics[width=.9\linewidth]{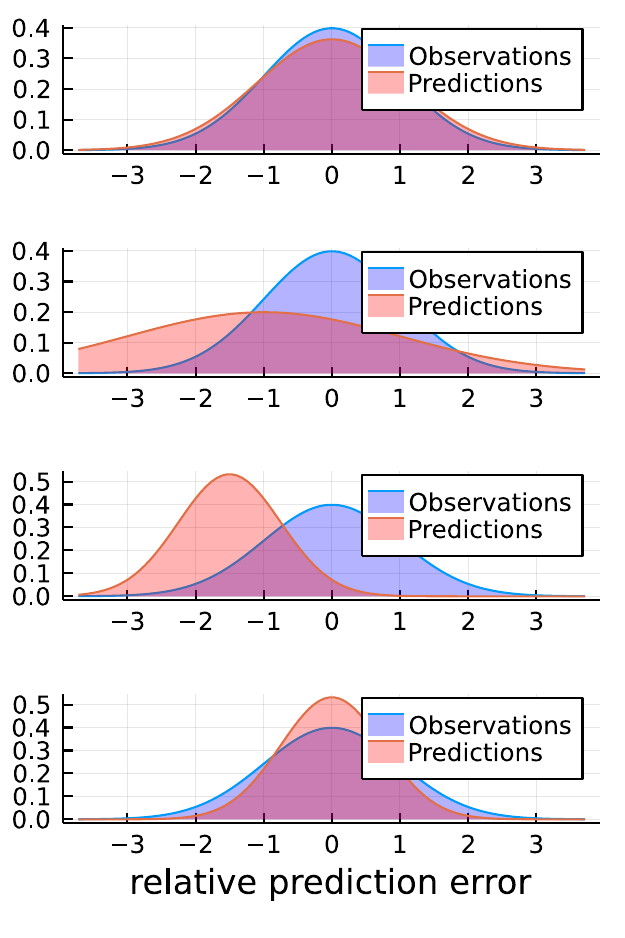}
\caption{\label{fig:normal-comparisons}Observation and prediction PDFs.}
\end{subfigure}
\begin{subfigure}[t]{0.30\textwidth}
\includegraphics[width=.9\linewidth]{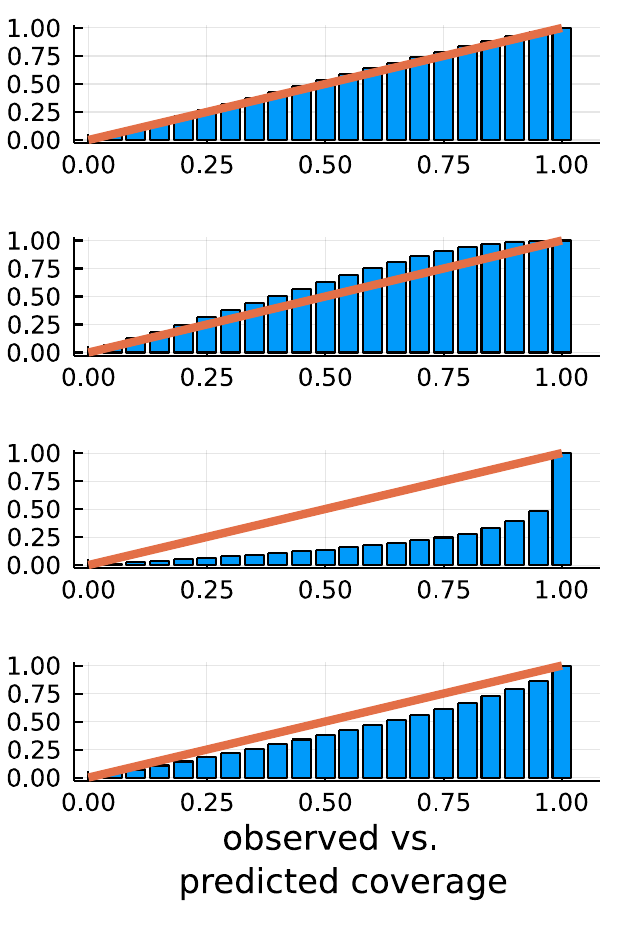}
\caption{\label{fig:normal-calibration-curves}Calibration curves.}
\end{subfigure}
\begin{subfigure}[t]{0.30\textwidth}
\includegraphics[width=.9\linewidth]{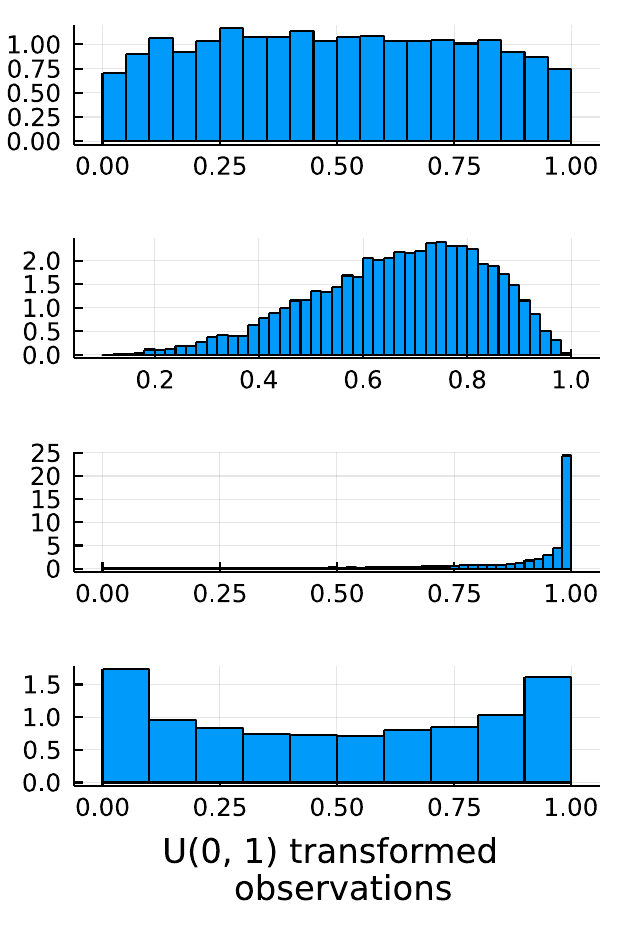}
\caption{\label{fig:normal-PIT-histograms}PIT histograms.}
\end{subfigure}
\caption{\label{fig:calibration-normal}Diagnostic plots for different prediction and observation distributions.}
\centering
\end{figure}

In order to relate this setting to the binary setting, we can define a binary event as follows:
For a given probability \(p\)  we ask our model to return a set or interval \(\mathcal{T}_p(D)\) that that contains the true realization \(y\) with probability \(p\).
For instance, we can define
\begin{equation}
\label{eq:prediction-interval}
\mathcal{T}_p(D) = \Bigl[\mathit{cdf}^{-1}(D, 0.5-\frac{p}{2}),\quad  \mathit{cdf}^{-1}(D, 0.5+\frac{p}{2})\Bigr] \end{equation}
where \(\mathit{cdf}^{-1}(D, p)\) denotes the inverse CDF of \(D\) at \(p\).\footnote{Note that there are many choices for constructing \(\mathcal{T}_p(D)\); for example, two other possible intervals are \(\left[y_{\rm min}, \mathit{cdf}^{-1}(D, p)\right]\) or \(\left[\mathit{cdf}^{-1}(D, 1-p), y_{\rm max}\right]\), and one could even use the set \(\left[y_{\rm min}, \mathit{cdf}^{-1}(D, 1-\frac{p}{2})\right] \cup \left[ \mathit{cdf}^{-1}(D, 1-\frac{p}{2}), y_{\rm max} \right]\).
For symmetric and unimodal distributions, the choice in \cref{eq:constructing-T} has two advantages: (i) for all \(p\) the prediction is \emph{sharpest}, i.e. the size of the interval is the smallest possible, and (ii) the prediction interval includes realizations with the highest predicted probability first.
For nonsymmetrical or multimodal distributions there are other, better suited choices for \(\mathcal{T}_p(D)\), but we expect the choice in \cref{eq:prediction-interval} to be a good default for many distributions.}
Then, with \(D = f(x)\) we can redefine \cref{eq:calibration-eq} as
\begin{equation}
\label{eq:calibration-eq2}
\underset{(x, y)}{\mathbb{P}} \left[y \in \mathcal{T}_p(f(x))\right] \approx p \qquad \text{for all } p \in (0, 1) \end{equation}
which generalizes our definition of marginal calibration to the regression setting and allows us to construct a similar diagnostic plot as for the binary case above.
\cref{fig:normal-calibration-curves} showcases several calibration curves for different observation and prediction distributions.
In particular, we note that the first calibration curve shows an almost perfectly calibrated prediction.

For a numerical example, consider a predicted distribution \(D=\mathcal{N}(0, 1)\), and let the target probability \(p\) be \(0.682\).
Then, we can compute the prediction interval
\begin{equation}\begin{aligned}
\label{eq:constructing-T}
 \mathcal{T}_{p}(D) =&\Bigl[\mathit{cdf}^{-1}(D, 0.5-\frac{p}{2}),& & \mathit{cdf}^{-1}(D, 0.5+\frac{p}{2})&\Bigr] \\
=&\Bigl[\mathit{cdf}^{-1}(D, 0.159),& & \mathit{cdf}^{-1}(D, 0.841)&\Bigr] \\
\approx &\left[-1, 1\right]. & & & &
\end{aligned}\end{equation}
In other words, if we have a realization \(y = 1\), we can add it to the plot in \cref{fig:normal-calibration-curves} as follows:
Find the smallest \(p\) for which \(y \in \mathcal{T}_p(D)\), i.e. \(p=0.682\).
Then, add a count to each bin associated with \(p \geq 0.682\).
If in the end each bin with associated probability \(p\) has (normalized) height \(p\) we consider the model marginally calibrated.

Let us finally modify \cref{eq:calibration-eq2} for the safety-critical setting.
Consider a regression variable that represents the state of the system, and consider states that are \emph{safe}, i.e. no special action is necessary, or \emph{unsafe}, i.e. an appropriate action needs to be taken.
We claim that in a safety-critical system, a model may output a set of likely states including both safe and unsafe states, even if the true state is safe, but must not erroneously output a set of only safe states when the true state is unsafe.

We therefore claim that, if a state has predicted probability \(p\), it must be observed with frequency \(p\) or higher.
Mathematically, we can therefore rewrite \cref{eq:calibration-eq2} as an \emph{inequality}, i.e.
\begin{equation}
\label{eq:calibration-ineq2}
\underset{(x, y)}{\mathbb{P}} \left[y \in \mathcal{T}_p(f(x))\right] \geq p \qquad \text{for all } p \in (0, 1).\end{equation}
Graphically, this means that the calibration curve must lie \emph{above} the identity line, as can be seen in the first two rows of \cref{fig:normal-calibration-curves}.
When comparing with the associated observation and predictions distributions on the left side (\cref{fig:normal-comparisons}), we can see that \cref{eq:calibration-ineq2} can be satisfied, even if the mode of the predicted distribution is somewhat wrong, as long as the standard deviation is sufficiently large.
In \cref{alg:test-calibration-curve} we present a simple algorithm to empirically validate marginal calibration for the safety-critical case, and suggest a \(10\%\) margin of tolerance if the inequality is violated.
Note that the computed observed frequencies in this algorithm can be plotted against the array of probabilities to construct \cref{fig:normal-calibration-curves}.

\begin{listing}[htbp]
\caption{\label{alg:test-calibration-curve}\texttt{function test\_calibration\_curve}. Empirical validation of marginal calibration following \cref{eq:calibration-ineq2}.}
\begin{Code}
\begin{Verbatim}
\color{EFD}\EFk{function} \EFf{test\_calibration\_curve}(predictions  \EFt{::} \EFt{Vector}\{\EFt{<:Distribution}\},
                                observations \EFt{::} \EFt{Vector}\{\EFt{<:Real}\};
                                eps=\EFhn{0.10})
    ps = \EFhn{0.05}:\EFhn{0.05}:\EFhn{1.0}
    counts = zeros(size(ps))
    \EFk{for} (pred, obs) \EFk{in} zip(predictions, observations)
        \EFk{for} (p\_idx, p) \EFk{in} enumerate(ps)
            counts[p\_idx] \EFt{+=} obs in invcdf\_interval(pred, p)
        \EFk{end}
    \EFk{end}
    observed\_frequencies = counts \EFt{./} length(observations)
    \textcolor[HTML]{483d8b}{\textbf{@assert}} all(observed\_frequencies .\EFt{>=} (ps \EFt{.*} (\EFhn{1}\EFt{-}eps)))
\EFk{end}

\EFk{function} \EFf{invcdf\_interval}(D\EFt{::Distribution}, p\EFt{::Real})
    Interval(invcdf(D, \EFhn{0.5}\EFt{-}p\EFt{/}\EFhn{2}), invcdf(D, \EFhn{0.5}\EFt{+}p\EFt{/}\EFhn{2}))
\EFk{end}
\end{Verbatim}
\end{Code}
\end{listing}

\paragraph{Assessing calibration and dispersion through the Probability Integral Transform.}
\label{sec:orgc870fec}
So far we have assessed calibration through the lens of \emph{coverage}, i.e. by computing a set of probable outcomes for a given probability \(p\), and computing its frequency of containing (covering) the true realization.
The Probability Integral Transform (PIT) provides another way of assessing calibration directly from the predicted cumulative density function \autocite[, Definition 3]{gneitingProbabilisticForecasting2014}.
In particular, the predicted distribution \(D\) is marginally\footnote{Note that in the definition is originally for probabilistic calibration, which is not distinguished here, but essentially equals marginal calibration if our prediction strategy does not change over time, see \textcite[, Section 2.2]{gneitingProbabilisticForecastsCalibration2007}.} calibrated iff the cumulative density of the observations \(y_i\) under \(D\) is equal to a uniform distribution, i.e.
\begin{equation}\mathit{cdf}(D, y_i) \sim U(0, 1).\end{equation}
\label{eq:uniform-distribution}
\cref{fig:normal-PIT-histograms} shows the cumulative density for several predictions and observations.
The problem with this approach for our application is twofold:
\begin{enumerate}
\item it is hard to modify this definition for the safety-critical setting, and
\item it is in general not easy to check a distribution for ``approximate'' uniformity.\footnote{The problem is that it is in general hard to come up with acceptable thresholds for simple measures like the mean, variance or skewness from first principles, especially in complex systems.
In addition, statistical tests are unapplicable if the distribution is not exactly uniform, and the sample size is large.
See \textcites{berkowitzTestingDensityForecasts2001}{hamillInterpretationRankHistograms2001}[, Section 3.1]{gneitingProbabilisticForecastsCalibration2007} for further discussion.}
\end{enumerate}
We can nonetheless use \cref{eq:uniform-distribution} to assess the \emph{dispersion} of the distribution and relate it to the safety-critical case.
Recall that dispersion relates the concentration of the predictions to the concentration of the observations.
Following a similar argument to the sections above, we claim that in the safety-critical case a prediction must always be either \emph{correctly dispersed} or \emph{overdispersed}.
In other words, a prediction must never be more concentrated than the observations.

We can now define a simple empirical test to verify acceptable dispersion for the safety-critical setting.
We first relate the dispersion to the variance of the transformed observations as computed in \cref{eq:uniform-distribution}.
Note that a perfect uniform distribution has a variance of \(1/12\).
Then, a predicted distribution is \emph{correctly dispersed} given the observations if the variance of the transformed observations is equal to \(1/12\), and the distribution is \emph{overdispersed} if the variance is lower.
In \cref{alg:test-dispersion} we showcase a simple implementation of this test, which includes a \(10\%\) margin of tolerance.
Additionally, in \cref{fig:normal-PIT-histograms} the result of over- and underdispersed predictions can be seen:
The first row shows an approximately correctly dispersed prediction, whereas the second and fourth row show over- and underdispersed predictions.

\begin{listing}[htbp]
\caption{\label{alg:test-dispersion}\texttt{function test\_dispersion}. Test the dispersion of the predictions w.r.t. the observations by comparing the variance of the transformed observations with the variance of a uniform distribution.}
\begin{Code}
\begin{Verbatim}
\color{EFD}\EFk{function} \EFf{test\_dispersion}(predictions  \EFt{::} \EFt{Vector}\{\EFt{<:Distribution}\},
                         observations \EFt{::} \EFt{Vector}\{\EFt{<:Real}\};
                         eps=\EFhn{0.10})
    \textcolor[HTML]{483d8b}{\textbf{@assert}} var([ cdf(pred, obs)
                 \EFk{for} (pred, obs) \EFk{in} zip(predictions, observations)]
                ) \EFt{<=} \EFhn{1}\EFt{/}\EFhn{12} \EFt{*} (\EFhn{1}\EFt{+}eps)
\EFk{end}
\end{Verbatim}
\end{Code}
\end{listing}
\subsubsection{From marginal to conditional calibration.}
\label{sec:orgd7a40e5}

\begin{figure}[t]
\centering
\begin{subfigure}[t]{0.30\textwidth}
\includegraphics[width=.9\linewidth]{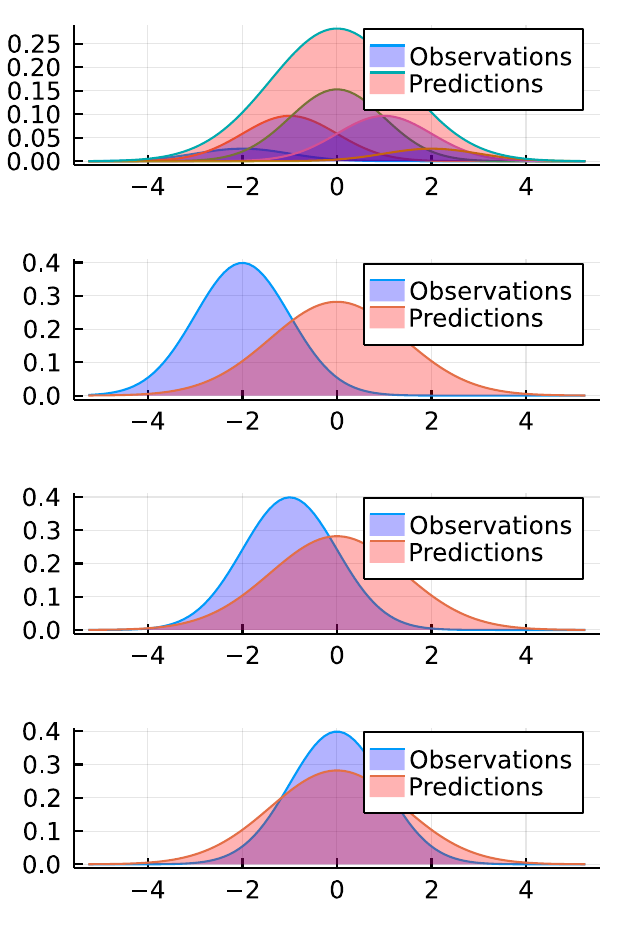}
\caption{\label{fig:marginal-calibration-failure-1}Observation and prediction PDFs for \(x \in \left\{ 1..5 \right\}\), \(x=1\), \(x=2\), and \(x=3\).}
\end{subfigure}
\begin{subfigure}[t]{0.30\textwidth}
\includegraphics[width=.9\linewidth]{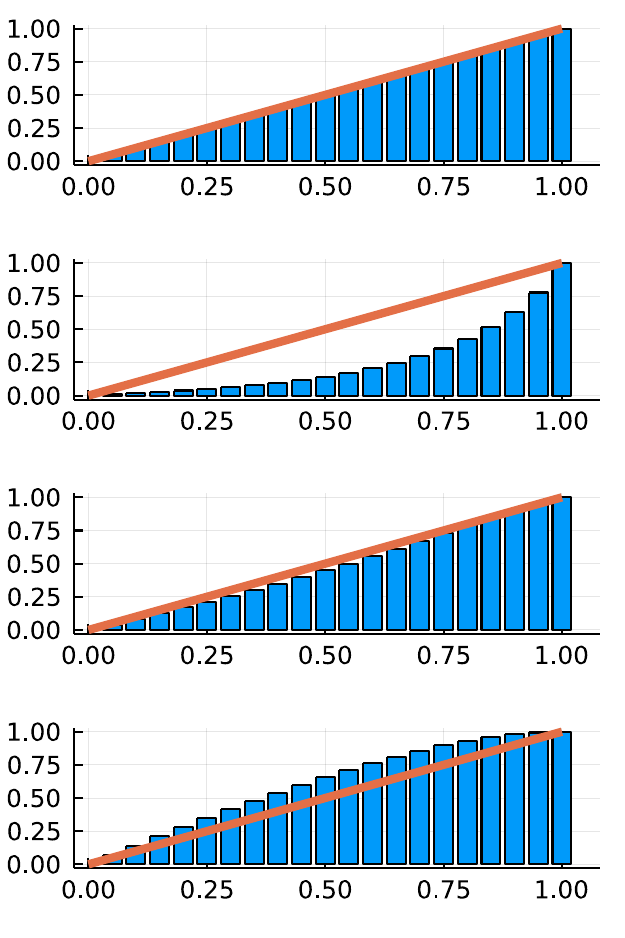}
\caption{\label{fig:marginal-calibration-failure-2}Calibration curves.}
\end{subfigure}
\begin{subfigure}[t]{0.30\textwidth}
\includegraphics[width=.9\linewidth]{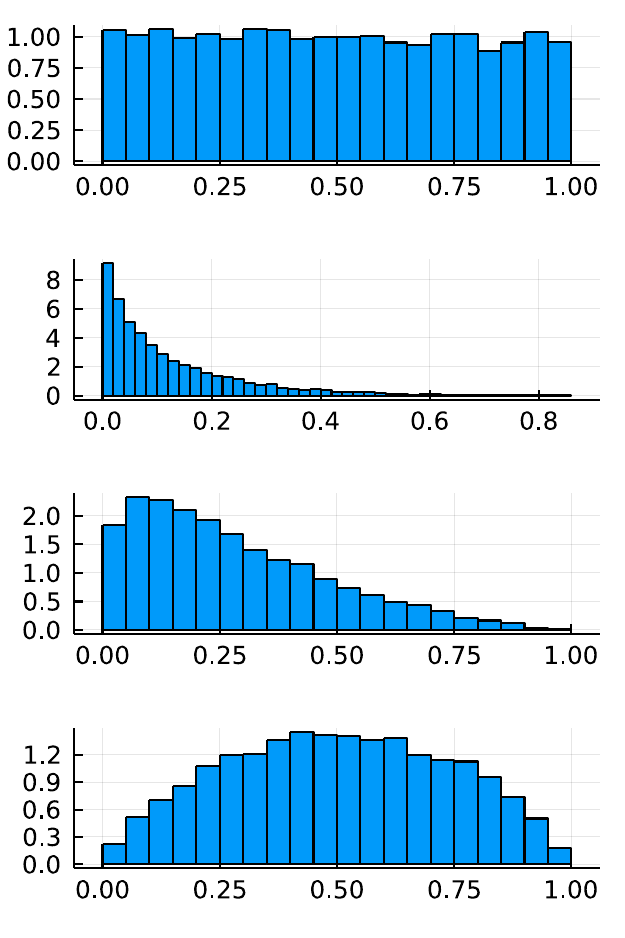}
\caption{\label{fig:marginal-calibration-failure-3}PIT histograms.}
\end{subfigure}
\caption{\label{fig:marginal-calibration-failure}The prediction is perfectly marginally calibrated (first row), but is uncalibrated if conditioned on \(x=1\) (second row) or \(x=5\) (not shown).}
\centering
\end{figure}

In the previous section we have investigated calibration in the \emph{marginal} sense.
In particular, we have considered metrics conditioned on \(p\), but marginalized over all \(x\), i.e.
\begin{equation}
\underset{\left\{ (x, y) \mid f(x)=p  \right\}}{\mathbb{P}} \left[ \mathit{event}(x) = \top \right]
\end{equation}
or
\begin{equation}
\mathbb{P} \left[ \mathit{event}(x) = \top \mid p(x) \right].
\end{equation}
In general we also require calibration \emph{conditioned on \(x\)}, i.e.
\begin{equation}
\mathbb{P} \left[ \mathit{event}(x) = \top \mid x, p(x) \right].
\end{equation}
In this section, we first motivate this through an example and then derive a practical algorithm that verifies conditional calibration of the model by conditioning on specific semantic groupings.

To understand the problem when computing only marginal calibration, consider a set of inputs \(x \in \left\{ 1..5 \right\}\) where \(x=3\) occurs most often, \(x=2\) and \(4\) occur less often, and \(x=1\) and \(5\) occur rarely.
Let observations be sampled from distributions \(\mathcal{N}(\mu(x), 1)\) with \(\mu(x) = x-3\).
Consider now a model that ignores the input \(x\) entirely and always predicts the distribution \(\mathcal{N}(\mu, \sigma^2) = \mathcal{N}(0, 2)\).
It turns out that this model has perfect marginal calibration even though it does not consider the influence of \(x\) at all!
Instead, the model predicts the distribution of observations correctly ``on average'', i.e. over all \(x\).
\cref{fig:marginal-calibration-failure} illustrates this problem.
In \cref{fig:marginal-calibration-failure-1}, the top row displays the predicted distribution against the observed distributions for each \(x\), which are weighted by the prior likelihood of \(x\).
Note that if we combined the observation distributions \(\mathcal{N}(\mu(x), 1)\) into a Gaussian Mixture Model with weights \(p(x)\) it would almost exactly match \(\mathcal{N}(0, 2)\).
Consequentially, in \cref{fig:marginal-calibration-failure-2} and \cref{fig:marginal-calibration-failure-3} we can see that predicted distribution has almost perfect marginal calibration when computed (marginalized) over all \(x\).

The problem becomes apparent in the second row where we consider only observations for \(x=1\).
Since the conditional distribution of the observations is different from the marginal distribution in the first row, the predicted distribution has poor coverage, which becomes apparent in the calibration curve and PIT histogram.
A similar problem occurs for \(x=5\), which is not depicted, and it occurs to a lesser extent for \(x=2\) (third row) and \(x=4\).
Still, the last row shows that a model can be conditionally calibrated for some \(x\), even if \(x\) is not considered in the model.

We can now wonder if conditional calibration should be required for all \(x\).
We argue that this is not the case, but that conditional calibration should be verified for high-level semantic variables.
Indeed, if \(x\) is an unobservable variable, or represents an oracle for a stochastic outcome, it is impossible to achieve conditional calibration for those \(x\), and marginal calibration is the best we can hope for.
On the other hand, if a model is always overdispersed during the daytime, and underdispersed during dusk and dawn, this may impose a significant safety risk.

We therefore propose the following verification procedure:
For each semantic feature in \(v = (v^c, v^s)\) we create subgroups of adjacent feature realizations, such that each subgroup has at least \(N\) samples.
Then, for each subgroup we evaluate \cref{alg:test-calibration-curve} and \cref{alg:test-dispersion} and record the total number of test that ran and that failed, i.e. \(n_{\rm tests}\) and \(n_{\rm fail}\).
\cref{alg:conditional-calibration} presents an implementation of this procedure.
Note that if enough samples are available, one may also condition on multiple semantic features at the same time.

\begin{listing}[htbp]
\caption{\label{alg:conditional-calibration}\texttt{function test\_conditional\_calibration}. An expansion of the marginal calibration and dispersion tests by conditioning on semantic feature subgroups.}
\begin{Code}
\begin{Verbatim}
\color{EFD}\EFk{function} \EFf{test\_conditional\_calibration}(model, xs, ys, vs; N=\EFhn{10}\_000)
    preds = model.(xs)
    n\_tests, n\_fail\_calibration, n\_fail\_dispersion = \EFhn{0}, \EFhn{0}, \EFhn{0}
    \EFk{for} i \EFk{in} eachindex(vs[\EFhn{1}])
        vs\_i = getindex.(vs, i)  \EFcd{\# }\EFc{i-th feature for each sample}
        sorted\_indices = sortperm(vs\_i)
        subgroup\_indices = partition(sorted\_indices, N)
        \EFk{for} idx \EFk{in} subgroup\_indices
            n\_tests \EFt{+=} \EFhn{1}
            \EFk{try} test\_calibration\_curve(preds[idx], ys[idx])
            \EFk{catch}; n\_fail\_calibration \EFt{+=} \EFhn{1} \EFk{end}
            \EFk{try} test\_dispersion(preds[idx], ys[idx])
            \EFk{catch}; n\_fail\_dispersion \EFt{+=} \EFhn{1} \EFk{end}
        \EFk{end}
    \EFk{end}
    \EFk{return} (n\_tests, n\_fail\_calibration, n\_fail\_dispersion)
\EFk{end}
\end{Verbatim}
\end{Code}
\end{listing}

We now have two problems left to solve: (i) Choosing a minimum number of samples per subgroup \(N\), and (ii) deciding how many tests may fail and still accept the system as certified.
As we will see now, these choices are directly related.

Naïvely, we could simply choose a small subgroup size \(N\) and allow no tests to fail.
Unfortunately, this fails both in practice and in theory, even when a prediction is perfectly calibrated; marginally and conditionally.
This is because random fluctuation will result in violation of \cref{eq:calibration-ineq2} almost certainly if the sample size is small.
For this reason we introduced an acceptable error margin \(\epsilon\) in \cref{alg:test-calibration-curve} and \cref{alg:test-dispersion}.
Together with an appropriate sample size \(N\) we can therefore expect the random fluctuation to be ``averaged out'' and small violations to be tolerated.
To determine \(N\) we can run a computational simulation that assumes perfectly matching prediction and calibration distributions and measure the empirical frequency of test failure, or ``false rejection'', for a given \(N\) and \(\epsilon\).
Further details can be found in \cref{sec:samples-for-conditional-calibration}, and we summarize that for an accepted error margin \(\epsilon\) of \(10\%\) we need a sample size \(N\) of approximately \(10'000\) in order to have a false rejection rate of below \(1\%\).

Finally, we can combine all of this to come up with a single binary decision for certification of a model's uncertainty quantification.
First, we verify marginal calibration and dispersion.
Then, we compute conditional calibration and dispersion for a set of subgroups as described above, see [alg].
We count the number of executed tests and failed tests for each, and consider the false rejection rate for the perfectly calibrated setting given \(N\), e.g. \(1\%\) for \(N=10'000\) and \(\epsilon=10\%\).
Finally, we compare \(n_{\rm fails}\) to the Binomial distribution, which indicates the probability of \(n_{\rm fail}\) failures given \(n_{\rm tests}\) total tests with a failure probability of e.g. \(1\%\) each.
This probability should be very low, and we suggest a threshold of \(0.1\%\).
An implementation of this final step of the certification for the model's uncertainty quantification is presented in \cref{alg:certify-model-uncertainty-quantification}, which concludes this section.

\begin{listing}[htbp]
\caption{\label{alg:certify-model-uncertainty-quantification}\texttt{function certify\_model\_uncertainty\_quantification}. The final step of model certification for uncertainty quantification, leveraging the previously established tests.}
\begin{Code}
\begin{Verbatim}
\color{EFD}\EFk{function} \EFf{certify\_model\_uncertainty\_quantification}(model, xs, ys, vs)
    \EFcd{\# }\EFc{Marginal}
    test\_calibration\_curve(model.(xs), ys)
    test\_dispersion(model.(xs), ys)
    \EFcd{\# }\EFc{Conditional}
    n\_tests, n\_fail\_calibration, n\_fail\_dispersion =
        test\_conditional\_calibration(model, xs, ys, vs)

    \textcolor[HTML]{483d8b}{\textbf{@assert}} test\_probability\_n\_fails(n\_tests, n\_fail\_calibration) \EFt{\&\&}
            test\_probability\_n\_fails(n\_tests, n\_fail\_dispersion)
\EFk{end}

\EFk{function} \EFf{test\_probability\_n\_fails}(n\_tests\EFt{::Int}, n\_fail\EFt{::Int}, p\_fail=\EFhn{0.01};
                                 thresh=.\EFhn{001})
    \textcolor[HTML]{483d8b}{\textbf{@assert}} (n\_fail \EFt{<=} n\_tests\EFt{*}p\_fail) \EFt{||}
            (pdf(Binomial(n, p\_fail), n\_fail) \EFt{>=} thresh)
\EFk{end}
\end{Verbatim}
\end{Code}
\end{listing}

\subsection{Out-of-distribution detection}
\label{subsec:OOD-detection}
\begin{figure}[t]
\centering
\includegraphics[width=0.4\linewidth]{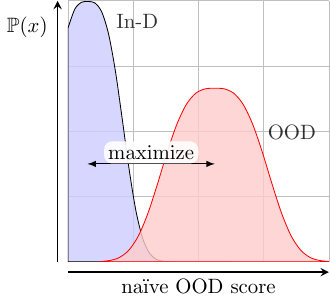}
% \includestandalone[width=0.4\linewidth]{./tikz-pictures/nominal-vs-anomaly/nominal-vs-anomaly}
\caption{\label{fig:nominal-vs-anomaly}In naïve OOD detection methods often significant overlap exists between \(OOD(x_{\rm In-D})\) and \(OOD(x_{\rm OOD})\).}
\end{figure}

This section discusses the usage of out-of-distribution (OOD) detection and its relation to model generalization, uncertainty estimation and feature collapse.
We argue that the term ``out-of-distribution'' is a misnomer for many applications and provides an alternative definition based on the model performance given a data sample.
Then, we establish a set of five desiderata which partially build on the assumptions and ideas introduced in \cref{sec:fundamental-assumptions} and \cref{subsec:inherently-safe-design}.
Each desideratum is discussed through the lens of both \emph{(mathematical) principles} and \emph{empirical evidence}, as in \cref{fig:how-to-certify}.
Principles are used to restrict possible classes of models, and empirical tests are defined where possible.

\par OOD detection is a topic closely related to uncertainty quantification, but addresses the Knightian notion of \emph{risk} instead of \emph{uncertainty} (see \cref{subsec:uncertainty-quantification}).
Although models can be empirically verified to produce calibrated uncertainty estimates on the training or validation distribution\footnote{Using a conformal prediction framework, correct coverage on the training distribution can even be proven, see \textcite{angelopoulosGentleIntroductionConformal2022a}. Still, the proof does not hold for samples from another distribution.\label{org3f12c23}}, generally uncertainty estimation must be understood as a \emph{model output} and is therefore subject to the model's generalization capabilities.
\textsuperscript{\ref{org3f12c23}}
In effect this means that for a novel data sample a model may predict a wrong, yet overly confident output.
In safety-critical scenarios this can lead to catastrophic consequences.
It is therefore necessary to not only try to predict the properties of output error in \emph{quantitative} terms, but also \emph{qualitatively} whether this quantity can be correctly estimated, i.e. whether the uncertainty quantification correctly generalizes to the current data sample.

To this end, \emph{we argue that OOD detection is often a misnomer}, since it is generally not clear \emph{which} distribution a sample should be checked against.
One could define a sample as OOD if it does not come from the exact same distribution as the training samples.
This definition quickly collapses in the high dimensional setting, often occurring in real-world computer vision problems.
For instance, recall again the runway example, and assume a run-time input image with an eagle in the top left corner -- an object which has not been included in the training distribution -- that is otherwise similar to the training samples.
Should this image be considered OOD?
It could clearly have never been sampled from the training distribution (since there were no eagles), yet we expect our model to make a good prediction.
One could say ``yes, but this image is close enough, and also this change should not really affect the prediction anyway''.
Unfortunately, it is often extremely difficult to clearly define what is ``close enough'' -- especially in the original input space (e.g. the pixel space) -- without involving large amounts of human bias or very extensive datasets.
Instead, we need to rely on the existence of semantic features which can be found in the latent representations of a model.

Instead of comparing exactly to the training distribution, one could also define ``the distribution of all feasible runway pictures when approaching a runway''.
Perhaps some other properties can be added, such as ``while the sun is at least 20 minutes from dusk or dawn'', ``while the sky is clear'', ``with an altitude between 300 and 3000 feet'', etc.
We argue that such a definition is impractical, mainly for two reasons:
\begin{enumerate}
\item For any description, an additional property could be added to make it more precise, and it is not clear when a sufficient description has been achieved.
\item Even if a precise description is specified, this gives little guarantees a lot about the model performance.
Since ``perfect coverage'' of this input distribution is generally impossible, any finite training distribution has to be considered incomplete, and the model may therefore make unforeseen prediction errors on any new inputs.
\end{enumerate}
In light of these conceptual difficulties we propose a more ``useful'' definition for out-of-distribution detection when applied to real-world, safety-critical applications:
\phantomsection
\label{ood-redefinition}
\begin{quote}
\textbf{Definition: (Out-of-distribution).}
Assuming a distribution of (trained) models \(\mathcal{M}\), we define a distribution of samples \(x \in \mathcal{D}\) for which
\begin{enumerate}[label=(\alph*)]
\item a true label exists,
\item the uncertainty predictions \(m(x)\) are well-calibrated or achieve good coverage with very high likelihood for \(m \sim \mathcal{M}\), and
\item the model predictions are close for all \(m \sim \mathcal{M}\).
\end{enumerate}
Then, a sample is considered ``in-distribution'' if it lies in \(\mathcal{D}\) and ``out-of-distribution'' otherwise.
\end{quote}
This definition helps us define OOD detection through the lens of whether we can rely on a model's predictions and its uncertainty estimates.
In practical terms, \(\mathcal{M}\) is usually the result of (re-)training a model multiple times with different random seeds or architectures, bootstrapping the training samples, or picking slightly different model-hyperparameters (network width, optimizer parameters, etc.).
The point of this is that whether a sample is considered OOD should be a function of the training setup, including training data, model architecture, optimizer, etc., and a sample should not be considered in-distribution if only a single model with an exact set of parameters can achieve the above properties.

\subsubsection{Six practical desiderata.}
\label{sec:org4605746}
Following the definition above, we propose six practical desiderata that aim to bridge the gap between definition and implementation.
Each desideratum is discussed with regard to \emph{principles} and \emph{empirical evidence} in the subsequent sections.
\begin{quote}
\textbf{Desiderata:} Assuming that \(OOD(x) = \bot\) we wish to imply that
\begin{enumerate}
\item the input contains all necessary semantic features and a true value \(y^{\ast}\) exists;
\item the model can successfully recover a low-dimensional, possibly disentangled, description of the semantic features;
\item the low-dimensional features can be combined correctly to produce the final prediction, even if the combination is novel;
\item the prediction is stable when sampling from the model distribution;
\item the predicted uncertainty measurement is well-calibrated and/or has correct coverage.
\end{enumerate}
Further, we require that
\begin{enumerate}
\setcounter{enumi}{5}
\item the OOD detection function must be constructed without the use of explicitly labeled OOD data.
\end{enumerate}
\end{quote}

\paragraph{Desideratum 1: The input contains all the necessary semantic features and a true label exists.}
\label{subsubsec:input-contains-necessary-features}
As listed in the definition, a sample is naturally OOD if it lacks a true label, i.e. if it inherently lacks the information required to make the prediction, irrespective of the model used.
For instance, predicting the current runway approach angle does not make sense given the picture of a cat.
In this desideratum we further refine this idea by assuming that the existence of a true label is caused by the existence of a set of \emph{necessary semantic features}.
Which semantic features must be present is dependent on the input and label and may vary from sample to sample.
Still, when necessary semantic features are occluded, cropped out, or otherwise obfuscated, this can lead to the inability of any model to make an accurate prediction.
Note that this desideratum is model-agnostic and instead is a property only of the input sample.

\textbf{Principles.}
Even though this desideratum is a property of the sample only, we generally can only verify it through the use of the model, usually lacking another way.
We conjecture that a principled way of determining the existence of necessary semantic features can happen through an inherently structured, and possibly disentangled, latent space model, as is described in \cref{subsec:inherently-safe-design}.
In this setting, the latent encoding of any semantic feature is located in a predictable place (e.g. in specific latent variables), and the \emph{lack} of any given semantic information has a predictable latent representation (e.g. by falling back to a prior).
Using this, OOD decisions can be made on the basis of the existence of \emph{all necessary} semantic features.

\textbf{Empirical evidence.}
Empirically, we aim to establish the relationship
\begin{equation}OOD(x) = \bot \Rightarrow x \text{ contains all necessary semantic information and } y^{\ast} \text{ exists}\end{equation}
\label{eq:ood-implies-necessary-info}
only using a trained model (or model ensemble) and the standard train and validation sets.

If a dataset exists which contains labels for the presence or absence or individual semantic features, this can be used directly.
If no such dataset exists, a new labeled dataset can be created by using regular training or validation samples, manually obfuscating individual semantic features and labeling the created sample accordingly.
Then, an intermediate ``existence detector'' must be created based on the model latent space, and its ability to predict the existence of each semantic feature must be measured by comparing it to the (additionally created) labels from the dataset.

We can additionally try to replace the implication (\(\Rightarrow\)) \cref{eq:ood-implies-necessary-info} with an equivalence (\(\Leftrightarrow\)) by asserting that
\begin{equation}OOD(x) = \top \Rightarrow x \text{ lacks some necessary semantic information }.\end{equation}
Note that unlike \cref{eq:ood-implies-necessary-info}, this equivalence is not strictly required.
Still, a simple test can be constructed by evaluating the model on the regular train and validation sets and selecting all samples for which \(OOD(x) =\top\).
For all of those samples we must then manually assert the lack of semantic information.

\paragraph{Desideratum 2: Successfully recovering a low-dimensional representation of the semantic features.}
\label{sec:orgc433da3}
While the previous desideratum was concerned with properties of the data, this one is about the capability of the model to successfully transform a sample from the input space to the latent space.
During this transformation we point out two possible classes of failures:
\begin{description}
\item[{Feature misrepresentation, or feature collapse.}] A semantic feature exists but is encoded as missing; a semantic feature is missing but is encoded as non-missing; a feature exists, but is mapped to the wrong representation; unseen new features collapse onto the same representation and loose significant information (feature collapse).
\item[{The structural bias is not suitable for the input.}] The structural biases encoded in the model architecture and training do not apply for the current input. For instance, a disentangled representation is assumed but does not hold, or the latent space is not expressive enough for a sufficient representation.
\end{description}
We argue that the former failure class can be addressed as follows:
The encoding (or lack thereof) of an existing or missing feature can be measured similarly to the method described in Desideratum 1.
For this case, both the correct \emph{presence} and the correct \emph{absence} of given features must be established.
A correct encoding of a present feature can then be verified by verifying a good prediction quality over the validation dataset.
Finally, \cref{subsec:feature-collapse} discusses the (challenging) topic of feature collapse in more detail.

For the latter failure class we argue that the most important step is to understand the problem at hand deeply and design the structural bias in the model accordingly.
For instance, the latent space must be carefully chosen to be large enough, e.g. by repeatedly increasing the size, retraining and monitoring an appropriate error metric.
Stronger biases like the validity of disentanglement must be derived from the problem directly and should therefore be made sure to hold ``in principle''.
For further empirical evidence we refer back to \cref{subsec:inherently-safe-design}.

\paragraph{Desideratum 3: Correctly combining the low-dimensional representation.}
\label{sec:orgbd5848c}
\begin{figure}[t]
\centering
\includegraphics[width=0.4\textwidth]{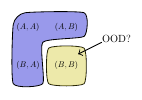}
% \includestandalone[width=0.4\textwidth]{tikz-pictures/distribution-cross-product/distribution-cross-product}
\caption{\label{fig:distribution-cross-product}An illustration of considering a novel feature combination. During training, feature combinations \((A, A), (A, B)\) and \((B, A)\) have been observed, but never \((B, B)\). Should \((B, B)\) be considered out-of-distribution, even if the model performs well?}
\end{figure}

As explained in the introduction to \cref{subsec:OOD-detection} and illustrated in \cref{fig:distribution-cross-product}, it needs to be assured that the inference based on the latent features generalizes to new feature combinations.

\subparagraph{Principles.}
\label{sec:orgc03163a}
A simple way is by making sure that for certain latent features \emph{all combinations} are included in the training set.
For instance, if the separation between \(v_{\rm style}\) and \(v_{\rm content}\) can be made, often it can be assured that all combinations of \(v_{\rm content}\) are in the training set.
In that case, an inference model only using \(v_{\rm content}\) for the prediction fulfills this desideratum.

Otherwise, we propose using a much-lower complexity model to do the inference from the latent space to the final prediction.
If the quality of the latent space is already very expressive, many problems can be solved with simple models for which generalization guarantees from statistical learning theory can be applied (see for instance VC-dimension based bounds \autocite{vapnikOverviewStatisticalLearning1999,bousquetIntroductionStatisticalLearning2004} or PAC-based bounds \autocite{mcallesterPACBayesianTheorems1998,guedjPrimerPACBayesianLearning2019a}).

\subparagraph{Empirical evidence.}
\label{sec:orge886998}
We can empirically test the prediction quality for missing feature combinations by manually filtering out certain feature combinations during training time and then evaluating on them later.
First, a baseline is established by training and evaluating the model using a standard train-test split.
Then, the whole dataset is considered again, and two random features \(v_i\) and \(v_j\) of a random sample \(v\) are selected.
A new training set is created by selecting all samples \(v'\) which differ from \(v\) in at least one of the two features.
In the continuous setting, we consider two features the same if they differ by less than some threshold \(\Delta\).
The samples with equal \(v\) in both features make up the test set.
Using these sets, the model is trained and evaluated, and the performance is compared to the baseline.
If this desideratum is satisfied, we expect the performance of the constructed dataset to be very close to the baseline, up to some margin; i.e.
$$\frac{\rm constructed}{\rm baseline}  - 1 < {\rm margin}$$
which can be established for multiple random feature selections.
See \cref{alg:test-new-feature-combinations} for an implementation.

\begin{listing}[htbp]
\caption{\label{alg:test-new-feature-combinations}\texttt{function test\_new\_feature\_combinations}. Test new feature combinations by explicitly withholding random combinations from the training set and using them to evaluate.}
\begin{Code}
\begin{Verbatim}
\color{EFD}\EFk{function} \EFf{test\_new\_feature\_combinations}(xs, ys, vs;
                                       Delta=\EFhn{0.2}, N\_repeat=\EFhn{20}, margin=\EFhn{0.1},
                                       N\_train=round(Int, length(xs)\EFt{*}\EFhn{0.80}))
    baseline\_perf = \EFk{begin}
        idx = sample(eachindex(xs), N\_train)
        model = train\_model(xs[idx], ys[idx])
        \EFk{return} eval\_model(model, xs[Not(idx)], ys[Not(idx)])
    \EFk{end}
    holdout\_perfs =
        [\EFk{begin}
            idx = samples\_without\_holdout\_feature(vs, Delta)
            model = train\_model(xs[idx], ys[idx])
            \EFk{return} eval\_model(model, xs[Not(idx)], ys[Not(idx)])
         \EFk{end} \EFk{for} \_ \EFk{in} \EFhn{1}:N\_repeat]

    \EFcd{\# }\EFc{performance: larger = better}
    \textcolor[HTML]{483d8b}{\textbf{@assert}} all(holdout\_perfs \EFt{./} baseline\_perf .\EFt{>} \EFhn{1} .\EFt{-} margin)
\EFk{end}
\EFk{function} \EFf{samples\_without\_holdout\_feature}(vs, Delta)
    v\_select = sample(vs)
    i, j = sample(eachindex(v\_select))
    \EFf{filter\_fn}(v) =\EFt{ !}((v[i] in v\_select[i] ± Delta) \EFt{\&}
                     (v[j] in v\_select[j] ± Delta))
    idx = filter\_fn.(vs)
\EFk{end}

\end{Verbatim}
\end{Code}
\end{listing}

\paragraph{Desideratum 4: Stable prediction under the distribution of model-hypotheses.}
\label{sec:org365d785}
When making a prediction, we need our prediction to not be overly sensitive to the concrete choice of a model; otherwise this is an indication that our model is not well suited for the input.
Indeed, for samples that are well in-distribution, we require that a prediction must be resistant to the specifics of the training procedure, random seeds, some model biases, architecture choices, and dataset bootstrapping.
To illustrate this need, imagine the opposite:
Given a sample and a set of models that achieve a similar training loss, and let each model output a prediction that is incompatible with the others.
We can conclude that either no true label exists for this sample, or that our choice of models are not able to generalize to this sample.
Taking this idea further, we can also try to guarantee this induction in the opposite direction:
Given that no label exists, or that our model choices are ill-suited, we can rely on a set of models disagreeing in their prediction.
The former idea is almost trivially true, but usually we are interested in the latter.
In the following we try to derive some principles which need to hold for the latter to be true.
Informally, let \(\mathcal{M}(\mathcal{D})\) be the distribution of models that arises from varying hyperparameters during training.
The more varied the hyperparameters are, the more evidence for the following conclusions is present.
For instance, \(\mathcal{M}(\mathcal{D})\) may be constructed simply by retraining the model with a new random seed (weak), or by varying architectural parameters like the activation functions on hidden layers (strong).
Then, again slightly informally, we expect
\begin{equation}OOD(x) = \bot \Leftrightarrow \underset{m \sim \mathcal{M}(\mathcal{D})}{\mathit{Var}} z_k = \mathit{encoder}_k(x) \text{ is \emph{small}}\end{equation}
\label{eq:ood-false-var-small-long}
for each latent index \(k\), and similarly
\begin{equation}OOD(x) = \bot \Leftrightarrow \underset{m \sim \mathcal{M}(\mathcal{D})}{\mathit{Var}} y_l = \mathit{model}_l(x) \text{ is \emph{small}}\end{equation}
for each output index \(l\).  Conversely, we expect
\begin{equation}OOD(x) = \top \Leftrightarrow \underset{m \sim \mathcal{M}(\mathcal{D})}{\mathit{Var}} z_k = \mathit{encoder}_k(x) \text{ is \emph{large} }\end{equation}
\label{eq:ood-true-var-large-long}
for any latent index \(k\)  and similarly for \(y_{l}\).
(Note that the operator \(\mathit{Var}\) may be replaced with an appropriate measure e.g. when \(z_k\) or \(y_l\) are distributions or sets.)

In order to reason about the equivalence (\(\Leftrightarrow\)) we will examine each induction direction (\(\Rightarrow\) and \(\Leftarrow\)) individually, for \(OOD(x) = \top\) and \(OOD(x) = \bot\) each, and we use the shorthand \(OOD(x) = \top \Rightarrow \mathit{Var} \text{ large }\) for \cref{eq:ood-true-var-large-long} etc.
We start by conjecturing that \(\mathit{Var} \text{ large } \Rightarrow OOD(x) = \top\) is trivially true as is explained above.
Similarly, since we know that each model \(m\) achieves an approximately equal, sufficiently small training loss, we can conclude that if the problem is well-behaved and a sample is in-distribution, each model will output approximately the same prediction. Therefore, \(OOD(x) = \bot \Rightarrow \mathit{Var} \text{ small }\) must hold.

The difficulty arises when trying to prove
\begin{equation}
\label{eq:ood-true-var-large}
OOD(x) = \top \Rightarrow \mathit{Var} \text{ large }
\end{equation}
or
\begin{equation}
\label{eq:ood-false-var-small}
    \mathit{Var} \text{ small } \Rightarrow OOD(x) = \bot.
\end{equation}
To illustrate this difficulty, let us first consider a low-dimensional regression example for which these implications do not hold.
A low-dimensional regression dataset is fitted using support vector machines (SVMs) that are trained using a stochastic optimizer.
Despite being a random process, generally speaking we expect each model to converge to approximately the same minimum, robustly with respect to dataset bootstrapping or different random seeds.
Critically though, due to their linearity the models will generally agree even when evaluated very much out-of-distribution, violating \cref{eq:ood-true-var-large} and \cref{eq:ood-false-var-small}.
This example highlights the fact that \emph{structural similarity} may result ``false generalization'', i.e. agreement between different models even if the prediction is wrong.

Another example occurs when replacing the SVMs with a Bayesian approach that falls back to a prior if the input sample is too far away from the training data.
For instance, if an ensemble of Gaussian processes with similar priors were to be used, the ensemble would show high agreement (low variance) in an out-of-distribution case due to the predictions being dominated by the prior.

In higher dimensions and when using neural networks, similar problems can occur.
For a simple illustration, consider a neural network with a final \emph{tanh}, \emph{sigmoid} or \emph{softmax} activation function.
In the OOD case, each ensemble member may generate arbitrarily large activations \(\xi\), which are all mapped \(\mathit{activation}(\xi) \overset{\xi \text{ large}}{\to} 1\); i.e. there is no variance between ensemble outputs despite the OOD case.
Some other models like variational autoencoders (VAEs) \autocite{kingmaAutoEncodingVariationalBayes2014} are inherently based on (latent-space) priors, and predictions might fall back to the same priors for all models in the out-of-distribution case.
In particular, VAEs are commonly optimized by maximizing the evidence lower bound (ELBO)
\begin{equation}
\label{eq:elbo2}
ELBO = \mathbb{E}_{z \sim \phi(x)} p(x \mid z) - D_{\rm KL}(p(z \mid x) \mid\mid p(z))
\end{equation}
with the latent space prior \(p(z)\).
In the case that \(p(x \mid z)\) is small everywhere (i.e. the generator can not reconstruct a matching image), the term is dominated by the KL-divergence, which encourages the encoder \(\phi(z)\) to make predictions close to the prior \(p(z)\), removing any variance between ensemble members in the OOD case.
As a final point, recently also implicit biases which result from the choice of loss function and optimizer have been investigated.
For instance, for two-layer networks, using a logistic loss has been shown to have strong implicit bias \autocite{chizatImplicitBiasGradient2020,yunUnifyingViewImplicit2021}.

In all these cases \cref{eq:ood-true-var-large} and \cref{eq:ood-false-var-small} may be violated, or at least it is not clear that they should hold for arbitrary OOD samples.
Still, approaches have been suggested for which empirically \cref{eq:ood-true-var-large} seems to hold; for instance Bayesian neural networks \autocite{bishopBayesianNeuralNetworks1997} as well as several approximations thereof have been proposed (for instance Deep Ensembles \autocite{lakshminarayananSimpleScalablePredictive2017}, Dropout \autocite{galDropoutBayesianApproximation2016}) and have empirically produced good results.
Nonetheless, we argue that despite their empirical success, for the goal of certification it is necessary to go beyond \emph{empirical evidence} and require \emph{principled reasoning} for \cref{eq:ood-true-var-large} and \cref{eq:ood-false-var-small} to hold.

\subparagraph{Principles.}
\label{sec:orgdb9f8af}
To alleviate the problem in the prior section, we aim to find principled approaches to ``guarantee'' \(OOD(x) = \top \Rightarrow \mathit{Var} \text{ large }\), i.e. \cref{eq:ood-true-var-large}.
Many principled ways to derive model architectures that satisfy both \cref{eq:ood-true-var-large} and \cref{eq:ood-false-var-small} may exist, many of which may not have been discovered yet or are not known to the author.
Therefore, here we only aim to give one exemplary approach.

We conjecture that \cref{eq:ood-true-var-large} holds with high probability if an ensemble of models is formed that has strong and varying inductive bias in each ensemble member.
Some forms of (automatically selected) hyperparameter ensembles have been proposed with the goal of improving generalization through diversity.
\textcite{wenzelHyperparameterEnsemblesRobustness2020} propose \emph{hyper-deep ensembles}, however the approach only varies ``hyperparameters related to regularization and optimization''.
\textcite{zaidiNeuralEnsembleSearch2021} propose ensembles with varied architectures that impose different ``anchor points'' for the weights of each ensemble member.

For our purposes, we propose going further and using even stronger inductive biases in each ensemble member.
For instance, in the VAE setting (\cref{eq:elbo2}) one could use different priors \(p(z)\) for each ensemble member.
In the in-distribution setting, one would expect the posterior \(p(z \mid x)\) to agree for all ensemble members, but to fall back to the priors \(p_i(z)\) in the out-of-distribution setting.

Another example would be varying the choice of activation functions among ensemble members.
For example, by employing the \emph{relu} activation function in one ensemble member, and the \emph{tanh} activation function in another, their behavior for large activations would vary drastically.

Finally, an extreme variant of this idea would be to vary the fundamental building blocks of a neural network, namely convolutional layers, transformer and attention layers, dense layers, etc.
If ensemble members that strictly rely on, say, convolutional layers agree in their prediction with other ensemble members that rely on transformer layers, we conjecture that this gives strong evidence for \cref{eq:ood-true-var-large} and conversely \cref{eq:ood-false-var-small}.

\subparagraph{Empirical evidence.}
\label{sec:org14c6b64}
In order to generate empirical evidence, first a set of models \(M\) is constructed by sampling from the model distribution \(\mathcal{M}\) which arises from any of the ensembling strategies.
For instance, \(M = \{m_1, m_2, m_3\}\) might be constructed by retraining a model architecture three times with different random seeds, by sampling three networks from a Bayesian neural network, etc.
Then, \emph{before} running the following tests, an acceptable false negative and false positive rate must be defined.
Using those, the ``OOD threshold'' \(\tau\) can be determined using a suitable method that must not rely on labeled out-of-distribution samples.
Finally, we can try to empirically verify \cref{eq:ood-false-var-small-long} and \cref{eq:ood-true-var-large-long} by again individually checking \(\Rightarrow\) and \(\Leftarrow\) for \(OOD(x) = \top\) and \(OOD(x) = \bot\).
For this, we replace \(m \sim \mathcal{M}\) with the sampled models \(m \in M\) and choose a suitable set of points \(x \in D\) for each case.
\begin{description}
\item[{Check \(\underset{m \in M}{Var} m(x) > \tau \Rightarrow OOD(x) = \top\) for ``most'' \(x \in D\).}] We first test that we do not reject ``too many'' points that are actually in-distribution.
We consider the training and validation sets and assume that they satisfy \(OOD(x) = \bot\) for the majority, but not all points (many datasets nonetheless can include ``dirty'' or corrupted samples of some form).
We set \(D = D_{\rm train} \cup D_{\rm val}\), compute the variance for each \(x \in D\) and select all samples for which the variance exceeds \(\tau\).
These samples are then inspected ``by hand'' and for each sample the cause of the high variance should be understood.
At the same time the number of ``false rejections'' can be empirically counted and subsequently compared with the acceptable false positive rate.
\item[{Check \(OOD(x) = \bot \Rightarrow \underset{m \in M}{Var} m(x) < \tau\) for ``almost all'' \(x \in D\).}] For this implication we propose constructing a smaller dataset \(D \subset D_{\rm train} \cup D_{\rm val}\) that contains only ``easy'' samples, for instance images taken in good conditions and in well-covered situations.
The variance is then computed for each sample, and the number of rejections should lie well below the specified false positive rate.
\item[{Check \(OOD(x) = \top \Rightarrow \underset{m \in M}{Var} m(x) > \tau\) for ``almost all'' \(x \in D\).}] Even though the OOD method must be constructed without the use of labeled OOD data, we can still use any data we have for evaluation.
Still, one must take great care to not involve this metric into the tuning process of the model to avoid overfitting to the choice of OOD data.
During test time we can construct \(D\) as the union of any labeled OOD datasets that are available, for instance the one constructed in \cref{subsubsec:input-contains-necessary-features}.
The variance is then computed and the number of rejections can be counted and compared against the false negative rate, which is typically very low.
\item[{Check \(\underset{m \in M}{Var} m(x) < \tau \Rightarrow OOD(x) = \bot\)}] As mentioned before, this is the most difficult condition to show empirically, and therefore requires the largest amount of ``principled-ness'' when constructing the OOD method.
Empirically we can not do much better than evaluate the variance on the training and validation datasets and verify that the implication holds for those samples, as has been done in the first check.
\end{description}

\paragraph{Desideratum 5: The resulting prediction is well-calibrated and/or has correct coverage.}
\label{sec:org8f825f2}
The previous desiderata aim to answer whether certain conditions about the input sample and the model transformations are satisfied.
In this desideratum we aim to establish properties of the \emph{model output}, rather than the input or latent encodings, which can be seen, in some sense, as the actual goal of OOD detection (and certification as a whole).

\subparagraph{Principles.}
\label{sec:orge1e9799}
Although it is difficult to establish this desideratum by itself, we argue that it can be concluded a result of the previous desiderata holding true.
More precisely, we must first assume that correct calibration and coverage is satisfied on the training (and test) set, as in \cref{subsec:uncertainty-quantification}.
Then, through the previous desiderata it is shown that the current input has semantic properties sufficiently similar to the training set, and that the model can encode them correctly.
It is thus natural to assume the model's generalization capabilities to the current input, and therefore conclude that the specified calibration and coverage properties hold.

\subparagraph{Empirical evidence.}
\label{sec:org669354d}
If no additional datasets are available we refer back to the analysis from \cref{subsec:uncertainty-quantification}.

\paragraph{Desideratum 6: OOD method construction without labeled OOD data.}
\label{sec:org5923cdf}
Furthermore, we note that the OOD detection function must be constructed without the use of explicitly labeled OOD data in order to avoid ``overfitting''.
This is commonly done by tuning thresholding parameters such that empirically a satisfactory true-negative (i.e. true in-distribution) rate is achieved; see for instance \textcite[Sec. 3]{liangEnhancingReliabilityOutofdistribution2020a}.

\subsection{Feature collapse}
\label{subsec:feature-collapse}
Novel realizations of a semantic features may ``collapse'', i.e. may get falsely mapped to the same representations as previously seen features.
This happens because the model has not learned to map these inputs differently and may therefore not have any incentive to do so.
In particular, new semantically meaningful features for the \emph{content} variables must be correctly mapped to new representations, and must not \emph{collapse} to representations created during training.
This scenario is called \emph{feature collapse}, see e.g. \textcite{vanamersfoortFeatureCollapseDeep2022}.
Slightly informally, we therefore formulate the following desideratum:

\textbf{Desideratum (bi-Lipschitz constraint):} Given two inputs, the (semantic) distance in the input space must be related to the distance in the embedding space.
In other words, if the difference between two embeddings is large, their corresponding inputs must be semantically significantly different.
Conversely, if two inputs are semantically distinct, they may not be represented by the same embedding.

Considering a model \(f(x)\) with lower and upper Lipschitz constants \(\underline{C}\) and \(\overline{C}\) (and \(\underline{C} < \overline{C}\)) we can therefore write
\begin{equation}
\underline{C} \|x - x'\|_X \leq \|f(x) - f(x') \|_Z \leq \overline{C} \|x - x'\|_X
\end{equation}
where \(X\) and \(Z\) are distance norms in the input and embedding space.

It is important to note that the norm in the input space is generally not easy to define, as it must capture \emph{semantic difference} rather than e.g. a pixel-wise euclidean difference.
On the other hand, the distance in the embedding space can often be simply chosen as the Euclidean distance.
Nonetheless, several works have argued that such a bi-Lipschitz constraint can be used to motivate regularization and lead to avoiding feature collapse.

\textbf{Principles.}
Most prominently, \autocite{vanamersfoortFeatureCollapseDeep2022} reviews three principles how a bi-Lipschitz constraint can be achieved, namely through
\begin{description}
\item[{a two-sided gradient penalty}] which enforces the gradient of the embedding w.r.t. the input to stay high (roughly \(\min \left(\|\nabla_x (y(x)-y_{\rm label})^2\|^2_2 - 1\right)^2\), essentially making it ``difficult'' to land on the training label distribution \autocites[see also][]{gulrajaniImprovedTrainingWasserstein2017}{mukhotiDeepDeterministicUncertainty2022}.
In practice this method has stability problems.
\item[{using spectral normalization}] of each layer, which keeps the spectral radius \(\rho\) controlled. Recall that, for a (bounded) linear transformation \(A\), the spectral radius defines ``how much an input may be scaled'', i.e. \(\|Av\|_2 \leq \rho(A)\|v\|_2\), and is equal to the largest singular value.
The desideratum can therefore ``naturally'' be enforced by applying spectral normalization to every layer, which is relatively straightforward for dense and residual layers, as well as batch normalization and certain activation function (see \cref{subsec:bilipschitz-for-layers}).
For further discussion, see \textcite{miyatoSpectralNormalizationGenerative2018,goukRegularisationNeuralNetworks2021,liuSimplePrincipledUncertainty2020,vanamersfoortFeatureCollapseDeep2022}.
\item[{use of a reversible model}] which explicitly enforces a bijective relationship between the inputs and the embeddings \autocite{jacobsen2018irevnet,behrmannInvertibleResidualNetworks2019}.
Although sometimes applicable, for the here discussed use case we consider this as a too restrictive and computationally demanding condition.\footnote{Note though that the generative modeling discussed in \cref{subsec:inherently-safe-design} may also be seen as a sort of reversible model.}
\end{description}

\textbf{Empirical evidence.}
For empirical evidence we suggest two approaches:
\begin{enumerate}
\item Computing the global lipschitz constant in both directions (see \cref{subsec:bilipschitz-for-layers} for more information), and
\item empirically test feature collapse for chosen features.
\end{enumerate}
For the empirical test we suggest selecting a content dimension \(v_i, i \in 1..k\) and removing the largest and smallest \(10\%\), respectively.
The other data is used as training data, the model is retrained, and evaluated on the training and test inputs.
An interval is constructed that empirically contains \(98\%\) of the embeddings computed from the training data, and it is asserted that at most \(2\%\) of the embeddings computed from the evaluation data lie in the interval.
An implementation is provided in \cref{alg:test-no-feature-collapse}.

\begin{listing}[htbp]
\caption{\label{alg:test-no-feature-collapse}\texttt{function test\_no\_feature\_collapse}. We withhold a section of the training data from the training and verify that it does not ``collapse'' during evaluation.}
\begin{Code}
\begin{Verbatim}
\color{EFD}\EFk{function} \EFf{test\_no\_feature\_collpase}(xs, ys, vs, v\_idx)
    vs\_i = getidx.(vs, i)
    idx = sortidx(vs[i]); N = length(vs)
    k = length(N)÷10
    idx\_train = idx[k:N\EFt{-}k]
    idx\_eval = setdiff(idx, idx\_train)

    model = train\_model(xs[idx\_train], y[idx\_train])
    \EFf{model\_i}(x) = model(x)[i]

    ys\_i\_train = model\_i.(xs[idx\_train])
    ys\_range = Interval(quantile(ys\_i\_train, [\EFhn{0.01}, \EFhn{0.99}])...)

    y\_i\_eval = model\_i.(xs[idx\_eval])
    \textcolor[HTML]{483d8b}{\textbf{@assert}} mean(y\_i\_eval .in [y\_range]) .\EFt{<} \EFhn{0.02}
\EFk{end}
\end{Verbatim}
\end{Code}
\end{listing}

\subsection{Adversarial attacks and defenses}
\label{subsec:adversarial-attacks-and-defenses}
In recent years, it has repeatedly been shown that neural network based models can be ``fooled'' by adversarial attacks that carefully perturb the input by a carefully chosen, yet small perturbation.
In this setting, highly erroneous model outputs can be induced, and sometimes even arbitrary wrong predictions can be enforced.
Several works indicate that such attacks may pose a realistic threat to systems acting in the real world.
Successful adversarial attacks on vision systems have been constructed by modifying real-world objects, e.g. by applying ``adversarial patches'', or by modifying the camera stream directly.
They have been shown to be robust to changes in viewpoints, distance, and resolution, and were successfully applied to still images and video streams.
Assuming an adversary, and in the setting of safety-critical applications, we must therefore conclude that such attacks can have catastrophic consequences.

In light of such problems it is therefore natural to ask if suitable defense mechanisms exist and whether they should be required in a certification protocol.
To answer this question, we first recall a selection of literature on adversarial attacks, including attacks on dynamic video scenes.
We discuss the applicability of certain classes of attacks and draw conclusions to necessary defense strategies in the presence of an adversary.
In addition, we motivate why adversarial attacks need only be considered in the presence of an adversary, and do not need to be assumed to occur naturally.

Then, the viability and necessity of adversarial defenses, both formal and empirical, is discussed.
In particular, by referring to recent competition results on adversarial defenses, we showcase the current gap between computationally feasibility and real applications.
Finally, we argue that even if computationally feasible, it is in general not clear what benefits formal, or empirical, adversarial defense guarantees would have for a certification protocol.

\subsubsection{Adversarial attacks.}
\label{sec:org420bb26}
Adversarial attacks on neural networks were first introduced by \textcite{szegedyIntriguingPropertiesNeural2014} and further discussed by \textcite{goodfellowExplainingHarnessingAdversarial2015}.
In the common setting, an input sample \(x\) is perturbed by a perturbation \(\delta\) such that the model prediction \(\mathit{model}(x+\delta)\) is far from the true prediction, where \(\delta\) is carefully chosen from an \(l_p\) ball\$, usually with \(p \in \{1, 2, \infty\}\).

Several adversarial attack methods exist, and can be broadly classified as follows:
\begin{description}
\item[{White box attacks}] are simple yet powerful methods which usually leverage access to gradient information, or other information about architecture and parameter to construct an adversarial example.
Important examples include the Fast Gradient Sign Method \autocite{szegedyIntriguingPropertiesNeural2014}, Projected Gradient Descent \autocite{madry2018towards}, or the Carlini \& Wagner attack \autocite{carliniEvaluatingRobustnessNeural2017}.
\item[{Black box attacks with inference access}] do not require access to the ``model internals'', but instead have access to model inference. They construct adversarial examples by repeatedly evaluating the model and considering the predicted output.
For instance, boundary attacks \autocite[e.g.][]{brendel2018decisionbased} use only model inference to construct adversarial examples by finding and traversing the model's decision boundary.
\item[{Black box attacks without model access}] typically construct adversarial examples though transfer attacks \autocites[e.g.][]{papernotPracticalBlackBoxAttacks2017}{liu2017delving}, which attack a separate, typically self-trained, model and apply the found adversarial example to the original target model.
\end{description}
Although originally adversarial examples considered only very small perturbations which are directly applied to the data (instead of being captured by a camera sensor), adversarial attacks have since been successfully applied to real-world settings by directly modifying objects, for instance \textcite{brownAdversarialPatch2018,thys2019fooling}.
\textcite{eykholtRobustPhysicalWorldAttacks2018} have shown that such attacks can be robust to changes in viewpoint, distance and resolution, and others have achieved similar results \autocite{athalyeSynthesizingRobustAdversarial2018}.
Finally, both white- and black-box attacks have also been successfully applied to video data, again either by modifying objects directly, or by applying a constant perturbation to the video stream \autocite{liAdversarialPerturbationsRealTime2019,jiangBlackboxAdversarialAttacks2019a,ijcai2019-134}.

Despite these results, others have pointed out that adversarial attacks might not be of concern in changing environments and while moving, for example in the context of autonomous driving \autocite{luNONeedWorry2017a,zengAdversarialAttacksImage2019}.
Yet, in light of the recent progress in adversarial attacks, and uncertainty about possible future methods, we conclude that the possibility of adversarial attacks must not be completely ignored.
For an extended discussion we refer the reader to \textcite[, Chapter IV]{akhtarThreatAdversarialAttacks2018}, as well as the upcoming work by \textcite{joshiAnalysingImpactAdversarial2022}.

\subsubsection{No adversarial attacks without an adversary.}
\label{sec:orgdc4bfe0}
On first sight, the existence of adversarial attacks could be understood as an inherent failure of the model, and a certified model could be required to be completely resistant to such attacks.
Yet multiple works have proposed that the existence of adversarial examples are inevitable, and must sometimes be understood as ``features, not bugs'' \autocite{ilyasAdversarialExamplesAre2019}, and that adversarial robustness may be at odds with accuracy and generalization \autocite{raghunathanAdversarialTrainingCan2019,tsipras2018robustness}.

Others have reasoned that the existence of adversarial examples is due to the violation of the i.i.d. assumption \autocite[see e.g.][, Chapter VII.B]{scholkopfCausalRepresentationLearning2021a}.
In particular, models are generally trained to minimize the \emph{risk}
\begin{equation}\min\ \underset{(x, y) \sim \mathcal{D}}{\mathbb{E}} \mathcal{L}(f(x), y),\end{equation}
where the samples \((x, y)\) are assumed to be independently and identically distributed (i.i.d.).
Since adversarial examples are carefully constructed, they violate the i.i.d. assumption, and thus the assumptions fundamental to the statistical learning setting.
The argument is therefore that if we assume real world images are i.i.d., adversarial examples will not occur naturally, or only with minimal likelihood.
In other words, without the presence of an adversary, there is no need to prevent or defend against adversarial attacks.
Consequentially, for the rest of the discussion we assume an adversary, and argue that no action needs to be taken otherwise.

\subsubsection{Attack prevention.}
\label{sec:orgdb73c0f}
Before discussing adversarial defenses based on the model, we first analyze the ``attack surface'' and how the construction of a successful adversarial attack can be prevented in the first place.
For the discussion it is assumed that an adversary exists who has access to the device running the model, and can make arbitrary inferences with it as many times as they like.

The first prevention strategy is prevention of access to model and dataset.
As discussed in the above section, white box attacks are generally the most powerful, but they require access to the full model parameters, or gradient information.
It is therefore natural to restrict the user from ``model introspection'', i.e. from accessing model parameters or architectural details, effectively removing the applicability of white box models.

Other attacks only rely on the combination of model inference and dataset access.
This may be partially mitigated by restricting partial or full access of the dataset used for training and development.
Finally, also attacks only relying on model inference itself exist.
We argue that this type of attack is generally very hard to mitigate without severely restricting access to model inference.

Another necessary prevention strategy becomes apparent when considering how an attack can be executed in the first place.
In the physical world, each attack must either modify the hardware (e.g. the camera sensor) directly, or modify the objects in the world.
We argue that in the case of hardware access, any kind of certification becomes near impossible, as not only the sensor might be modified, but also computer components might be exchanged, code may be modified, or direct physical attacks irrespective of the computing system might be made. \cref{table:adversarial-prevention} summarizes these points.

\begin{table}[t]
\caption{\label{table:adversarial-prevention}Adversarial prevention strategies by attack type and attack vector.}
\centering
\begin{tabular}{l|ll}
 & modifying the world & modifying the sensor\\
\hline
white box attack & \emph{restrict model introspection} & \emph{restrict model introspection}\\
inference access & \textbf{feasible} & \emph{restrict sensor access}\\
no model access & \emph{restrict dataset} & \emph{restrict sensor access}\\
\end{tabular}
\end{table}

\subsubsection{Adversarial defenses.}
\label{sec:org94b2df0}
So far we have seen that adversarial attacks are a feasible risk for real-world systems if no action is taken to protect the model.
Several defense strategies have been proposed, which can roughly be divided in \emph{heuristic} and \emph{formal} approaches.
For instance, \textcite{madry2018towards} proposes a simple heuristic defense by training on adversarially perturbed samples, instead of the original samples.
Others have introduced methods to derive formal guarantees for adversarial robustness.
This is usually done though a form of abstract interpretation, i.e. by defining a convex set of input points and computing every possible output for that set (\autocites[e.g.][]{wongProvableDefensesAdversarial2018}{tjeng2018evaluating}{singhAbstractDomainCertifying2019}).
A comprehensive exposition can be found in \autocite{salmanConvexRelaxationBarrier2019}.
The provided guarantees are typically as follows:
\begin{quote}
\textbf{Adversarial defense guarantee:} Given a sample \(x\), a perturbation set \(\mathcal{P}\), and a postcondition \(\mathit{cond}\), an algorithm tries to construct a proof that no perturbation \(\delta\) in \(\mathcal{P}\) exists such that \(\mathit{model}(x+\delta)\) violates the postcondition.
The postcondition is typically that the classification output is correct, or that a regression output is within certain bounds, and \(\mathcal{P}\) is typically chosen as an \(l_p\) ball with \(p \in \{1, 2, \infty\}\).
\end{quote}
In the following sections we highlight both conceptual and computational problems with this approach.

\paragraph{Conceptual problems.}
\label{sec:org7ad0863}
The kind of adversarial defense guarantee specified above leads to two key conceptual problems: First, the \(l_p\) balls become exponentially sparse in high dimensions, and second, the choice of the \(l_p\) ball is somewhat arbitrary, and motivated by methods, rather than the problem itself.

To illustrate the sparsity of the ``defended regions'' we can compute the volume covered by an \(l_p\) ball for a toy problem.
Let the domain of \(x\) be restricted to \(D=[0, 1]^d\), and let the radius \(\epsilon=0.1\) and \(p=\infty\).
Then, the volume of the defended \(l_p\) ball \(B_{\epsilon=0.1}^{p=\infty}(x)\) is \(0.2^d\), and the ratio of the domain covered by the ball is \(\frac{0.2^d}{1.0^d}=\left( \frac{1}{5}\right)^d\), which quickly approaches \(0\) as \(d\) grows large.
Note additionally that \(vol(B_{\epsilon}^{p=\infty}) > vol(B_{\epsilon}^p)\) for all \(p\in \mathbb{N}_{>0}\).
This problem is exaggerated by the fact that a finite set of samples becomes increasingly sparse in high dimensions in the first place.\footnote{One counterargument the low coverage is that the manifold on which the data lies also becomes exponentially small in high dimensions. Therefore, the ratio of the covered manifold may not shrink as fast. Unfortunately, no study analyzing this hypothesis is known to the author.}

The other problem is that in the real world, an \(l_p\) ball is usually neither a good representation of naturally occurring perturbations, nor of possible adversarial perturbations.
In fact, since arbitrary unrealistic perturbations are ``allowed'', it is often hard or impossible to construct adversarial defense guarantees for radii of non-negligible size.
Some works try to mitigate this problem and demonstrate adversarial defense guarantees for geometric properties like rotation \autocite{balunovicCertifyingGeometricRobustness2019} or semantic properties \autocite{mirmanRobustnessCertificationGenerative2021} but still show severe limitations in their applicability.

\paragraph{Computational problems.}
\label{sec:org6ca9c18}
Using convex over-relaxations to verify adversarial robustness can be done exactly (for piecewise linear networks), but requires solving a Mixed-Integer Linear Program, which is NP-hard and scales poorly with network size \autocite{tjeng2018evaluating}.
Instead, sound approximation methods are used that construct an over-approximation of the reachable output set, but may fail to verify properties that actually hold.
Much work has been done in this domain and significant speedup has been achieved \autocites[e.g.][]{wongProvableDefensesAdversarial2018}{wangEfficientFormalSafety2018}{zhang2018efficient}{singhAbstractDomainCertifying2019}{salmanConvexRelaxationBarrier2019}{xu2021fast}{wang2021beta}.
Still, due to the inherent limitations of the approach, computational concerns remain.

In order to understand the capabilities of current approaches, we refer to results from the latest ``International Verification of Neural Networks Competition (VNN-COMP) `21'' \autocite{bakSecondInternationalVerification2021}.
In this competition, twelve different teams (algorithms) competed on nine different benchmarks, which included vision and decision-making benchmarks, for instance:
\begin{description}
\item[{ACAS Xu benchmark}] verifying a 6-layer MLP with a total of 300 neurons, evaluated an airspace decision-making problem with a five-dimensional input and five-dimensional output space;
\item[{Cifar10-ResNet benchmark}] verifying a small ResNet \autocite{heDeepResidualLearning2015} with up to 4 residual blocks\footnote{For reference, the smallest network proposed in the original ResNet paper had 18 residual blocks, and the biggest 152 \autocite{heDeepResidualLearning2015}.}, evaluated on cifar10 images \autocite{Krizhevsky09} which are \(32 \times 32\) colored images of a diverse set of objects; and
\item[{Eran benchmark}] verifying an 8-layer MLP with width 200, evaluated on \(28 \times 28\) greyscale MNIST images \autocite{lecun1998gradient}.
\end{description}
By using a time-limit of up to five minutes per sample, for each benchmark a method existed that verified 100\% of samples.
Notably though, it was not the same method for each benchmark, and for no benchmark a verification coverage of 100\% was achieved by more than one method.

Although these results are certainly encouraging, they also showcase that as of today, most vision tasks are not within the realm of certifiability.
For instance, the Cifar10-ResNet benchmark considers a network which is between one and three orders of magnitude smaller than typically employed ResNet models.
In addition, the inputs were comparatively low resolution and were only verified with an allowed perturbation in \(B_{\epsilon=1/255}^{p=\infty}(0)\).

\subsubsection{Conclusions for certification}
\label{sec:orgc804bad}
It has been shown that current methods can manage to fool networks in real-world settings, both for dynamic image and video data, and future methods or unpublished work may still arise.
Still, it has to be recognized that standard adversarial attacks have to be carefully constructed, and do not typically arise naturally.
Therefore, mitigation of adversarial attacks must only be considered if the existence of an adversary is assumed.

If this is the case, it is likely that stronger adversarial attacks can be constructed by leveraging access to model information like the architecture and parameters, and by leveraging access to the dataset used for model construction and development.
We therefore suggest that a sufficient level of model obfuscation must be applied before deploying the model, such that an adversary can not access detailed information about model parameters.
Restricting access to the training and development data may be required in order to further reduce the attack surface, and physical access to the sensor and computing devices by an adversary must be prevented.

Finally, regarding adversarial defenses, we argue that formal defenses are conceptually incomplete and often not computationally feasible, and therefore should not be included in a certification requirement at this time.
Empirical defenses on the other hand may be helpful, but it is not clear what requirements would be reasonable in a general setting, and should therefore be left as a general performance improvement tool, but not part of certification requirements.

\chapter{A Proposed Model}
\label{sec:proposed-model-and-dataset}
\noindent In this final chapter, we propose a concrete model structure that aims to satisfy all assumptions and desiderata discussed up to this point, and can be trained using only weak supervision; i.e. it can make numeric predictions for the content variables by using only access to input pairs \((x, x')\) and knowledge of the operating parameters defined in \cref{sec:fundamental-assumptions} during training.

The proposed model structure consists of three separate modules, namely
\begin{enumerate}[label=(\roman*)]
\item an ensemble of \(E\) encoders
\begin{equation} f_i: x \mapsto D_z, \end{equation}
\item a single decoder (or generator)
\begin{equation} g: z \mapsto D_x,\end{equation}
and
\item a ``model head'' with
\begin{equation}\mathit{head}: D_{z^c} \mapsto D_y, \end{equation}
\end{enumerate}
where \(D_{\mathit{var}}\) denotes a distribution over the variable \(\mathit{var}\).
Additionally, we propose a novel method for OOD detection, i.e.
\begin{equation}
\mathit{OOD}: x \mapsto \top \text{ or } \bot,\end{equation}
which relies on the ensemble of encoders and the model head.

We first describe how the encoders and decoder are constructed using a variational autoencoder, and how a well chosen loss function results in the model being able to recover disentangled variables in a weakly-supervised manner, i.e. without access to numerical labels \(y^{\rm label}\).
Then, we show how we can construct a \emph{parameter free} model head through an unparameterized variable transformation and a simple linear model which can be constructed again without using labls \(y^{\rm label}\).
Additionally, we show how the model head can use uncertainty in \(z\) as quantified by the encoders to directly compute uncertainty in the predictions without any additional parameters.
Finally, we explain how the ensembling structure of the encoder, together with the model head, can be used to construct an out-of-distribution discriminator, again without the use of any additional parameters. The discriminator requires only a single threshold variable which can be chosen based physical properties of the problem.
We do not further discuss feature collapse and adversarial robustness, referring instead to what has been written in \cref{subsec:feature-collapse} and \cref{subsec:adversarial-attacks-and-defenses}.

\section{Recovering semantic variables in a metric space using weakly-supervised learning}
\label{sec:org22374c8}
As we have seen in \cref{subsubsec:self-supervised-recovery-of-disentangled-vars} it is impossible for a model to recover the underlying ``causes'' of the data in a fully self-supervised manner.
We therefore follow the method proposed by \textcite{locatelloWeaklySupervisedDisentanglementCompromises2020} and use a variational autoencoder \autocite[VAE,][]{kingmaAutoEncodingVariationalBayes2014} to recover a manually defined set of high-level semantic variables, which the model represents in a metric space, i.e. with a meaningful definition of distance.
In particular, the method uses a special modification of the regular ELBO loss that requires pairs of inputs \((x, x')\) together with an index set \(\mathcal{S}\) as an annotation, such that \(x\) and \(x'\) are equal in semantic variables \(v_i\) with \(i \in \mathcal{S}\), but vary otherwise.
In other words, we will see that it is sufficient to define the semantic information of the content variables only through example input pairs, but without having access to labels \(y^{\rm label}\).

Additionally, as introduced in \cref{subsec:ssl-for-certifiable-models}, we argue that the weakly-supervised training setting is a strong component for \emph{inherently safe design} (\cref{subsec:inherently-safe-design}), especially if the model additionally satisfies the desiderata for \emph{run-time error detection} (\cref{subsec:runtime-error-detection}).

We will now briefly review the loss function.

\subsection{A pairwise ELBO}
\label{sec:a-pairwise-elbo}
\begin{figure}[htbp]
\centering
\includestandalone[width=0.6\textwidth]{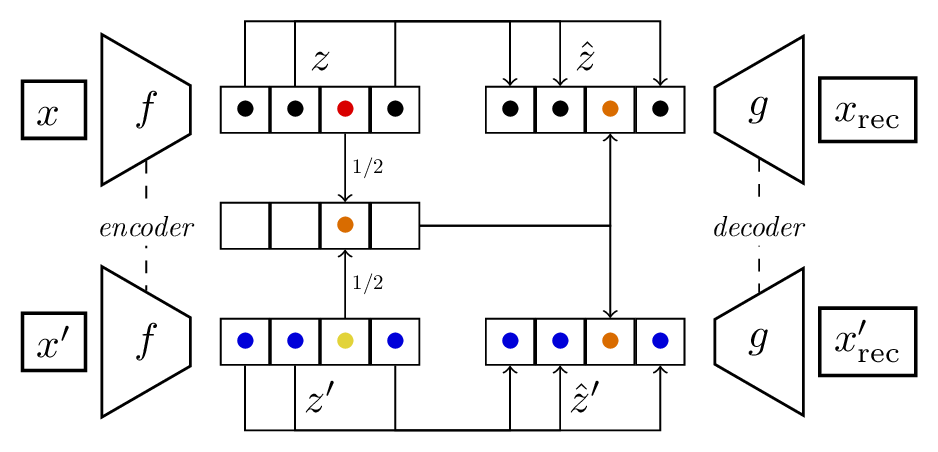}
\caption{\label{fig:disentanglement-loss}An illustration of the disentanglement loss. In this case \(v^c_{i=3}\) is constant, i.e. \(\mathcal{S}=\{ 3 \}\), and \(z^c_3\) is therefore averaged.}
\end{figure}

\noindent Suppose now we have an input pair \((x, x') = (g^\ast(v), g^\ast(v'))\) with associated index set \(\mathcal{S}\) such that \(v_i\) and \(v_i'\) are equal iff \(i \in \mathcal{S}\).
Then, assuming \(z \approx v\), we can write
\begin{equation}
\begin{aligned}
p(z_i \mid x) &= p(z_i \mid x') \qquad \text{for } i \in \mathcal{S} \\
p(z_i \mid x) &\neq p(z_i \mid x') \qquad \text{otherwise}
\end{aligned}
\end{equation}
or in other words \(f(x)_i = D_{z_i} \overset{!}{=} D_{z'_i} = f(x')_i\) for \(i \in \mathcal{S}\).
We now define two new variables \(\hat{z}\) and \(\hat{z}'\) with distributions \(\hat{f}(x) = D_{\hat{z}}\) and \(\hat{f}(x') = D_{\hat{z}'}\) defined as
\begin{equation}
\hat{f}(x)_{z_i} = D_{\hat{z}_i} = \begin{cases}
    \mathit{avg}(D_{z_i}, D_{z'_i}) & \text{for } i \in \mathcal{S} \\
    \phantom{\mathit{avg}(}D_{z_i} & \text{otherwise}
\end{cases}
\end{equation}
and similarly for \(\hat{f}(x')\), with a chosen \emph{averaging function} \(\mathit{avg}(D_1, D_2)\), which we define for normal distributions \(\mathcal{N}(\mu, \sigma)\) as
\begin{equation} \mathit{avg}\left(\mathcal{N}(\mu, \sigma), \mathcal{N}(\mu', \sigma')\right) = \mathcal{N}\left(\frac{\mu+\mu'}{2}, \left( \frac{\sigma+\sigma'}{2} \right)^2\right). \end{equation}
\cref{fig:disentanglement-loss} provides a related illustration.

Following \textcite[, Eq. 6]{locatelloWeaklySupervisedDisentanglementCompromises2020} we can now insert \(\hat{z}\) and \(\hat{z}'\) into the ELBO loss typically used for VAEs, i.e.
\begin{equation}
\label{eq:double-elbo}
\begin{aligned}
\max\ \mathbb{E}_{(x, x')}\ &\mathbb{E}_{\hat{z}\phantom{'} \sim \hat{f}(x)\phantom{'}} \log \mathit{pdf}(g(\hat{z})\phantom{'}, x)\phantom{'} - D_{\rm KL}\left(\hat f( x )\phantom{'} \| p(z)\phantom{'}\right) \\
+\ &\mathbb{E}_{\hat{z}' \sim \hat{f}(x')} \log \mathit{pdf}(g(\hat{z}'), x') - D_{\rm KL}\left(\hat f(x') \| p(z')\right) \\
\end{aligned}
\end{equation}
where \(p(z_i) = p(z_i') = \mathcal{N}(0, 1)\) is a prior chosen prior for each latent variable.

Intuitively, we can understand this approach as follows:
By averaging each dimension \(i \in \mathcal{S}\) we prevent the model from using those dimensions to explain the difference between \(x\) and \(x'\).
Instead, these dimensions must encode the similarities of the two inputs and, if \(|\mathcal{S}| = 1\), forces the model to do this in a single dimension.
Additionally, since the VAE samples from a distribution of predicted \(z_i\) during training, we get a strong regularization in the representation of the disentangled variables.
In particular, note that \(g(z)\) must resemble \(g(z \pm \epsilon_i)\) for all \(z \sim p(z)\) and small \(\epsilon_i\), but must recover the differences for larger \(\epsilon_i\).
Therefore, if all \(z_i \sim p(z_i)\) are sampled with sufficient probability during training, this enforces a sort of continuity constraint on the representations \(z_i\) and leads to them having a monotone relationship to \(v_i\)
Further, the distance between \(z_i\) and \(z_i'\) must be related to the distance between \(v_i\) and \(v_i'\), which enables us to use these representations to compute our final predictions \(y\) later.
Note however that the prior for \(z_i\) is chosen as \(\mathcal{N}(0, 1)\), whereas the prior for \(v_i\) is \(\mathit{Uniform}[a, b]\) (see \cref{sec:fundamental-assumptions}), which implies that the relationship between \(z_i\) and \(v_i\) is not linear.
We return to this observation in \cref{subsec:model-head}.

In order to use this loss function for our use case we can now proceed as follows:
\begin{enumerate}[label=\textbf{Step \arabic*:}, align=left]
\item Collect a number of samples \(x\) with semantic annotations \(v^c_i\).
\item Create a dataset of pairs \((x, x')\) such that they have one equal content variable \(v^c_i\) but vary otherwise, and set \(\mathcal{S} = \left\{ i \right\}\).
\item Set \(k\), the number of dimensions for \(z^c\), equal to the number of content variables.
\item Choose \(l\), the number of dimensions for \(z^s\), larger than the obvious number of style variables.
\item Train the VAE using the proposed loss function.
\item Verify the disentanglement properties of the model following \cref{subsubsec:disentanglement}.
\end{enumerate}

An implementation of the algorithm is provided in \cref{alg:disentanglement-loss} located in \cref{subsec:impl-details-loss}.

\section{Computing the final predictions using a linear model}
\label{subsec:model-head}
In the previous section we established that we can recover the content variables \(v^c\) in a metric space using \(z^c\), such that (i) the distribution of \(z^c_i\) will be close to the prior \(p(z_i) = \mathcal{N}(0, 1)\), (ii) \(z^c_i\) is correctly ordered, and (iii) distances between two \(z^c_i\) are meaningful.
Additionally we remember from \cref{subsec:five-central-assumptions} (Assumption 4) that the content variables \(v^c_i\) are distributed according to a \emph{known} uniform distribution \(\mathit{Uniform}[a_i, b_i]\).

To progress, we recall that we can transform any random normally distributed variable \(Z \sim \mathcal{N}(0, 1)\) to a uniform variable \(U \sim \mathit{Uniform}[0, 1]\) by defining \(U = cdf(\mathcal{N}(0, 1), Z)\).
Using this principle\footnote{Note that we make a hidden assumption here, namely that \(z\) has the correct ordering direction, i.e. \(z < z' \Rightarrow v < v'\).
This is however not necessarily true, as the representation of \(z\) might be ``flipped'', although distances are preserved.
To solve this we can use a single additional annotation \((x, x')\) with \(v^c_i \ll v'^c_i\) for each content variable that allows us to correct the direction.
If the representation does need to be flipped we can simply reassign \(u_i \leftarrow (1-u_i)\).}
, we simply transform each content representation \(z^c_i\) to a new variable
\begin{equation}u_i = cdf(\mathcal{N}(0, 1), z^c_i).\end{equation}
\label{eq:transform-z}
Now, assuming that the density of \(v^c\) is correctly distributed in the representations \(z^c\), the transformed variables \(u\) simply need to be rescaled according to the operating parameters to produce the final prediction \(y\), i.e.
\begin{equation}
\begin{aligned}
y_i &= u_i \cdot (b_i - a_i) + a_i \\
    &= cdf\left(\mathcal{N}(0, 1), z^c_i\right) \cdot (b_i - a_i) + a_i \\
    &= \mathit{cdf}\left(\mathcal{N}(0, 1), f^\mu(x)_{z^c_i}\right) \cdot (b_i - a_i) + a_i.
\end{aligned}
\end{equation}
where \(f^\mu(x)_{z^c_i}\) denotes the mean of \(D_{z^c_i}\).
Thus we have managed to make numerical predictions for the outputs \(y_i\), e.g. the horizontal or rotational offset of the runway example, without using any labels \(y^{\rm label}\).

Next we show how to also extract uncertainty estimates directly from the representations \(z^c_i\).
The intuition is that we can directly use the uncertainty estimates produced by the encoder \(f\) and simply rescale them.
In particular, the final prediction is again a Gaussian distribution with \(D_{y_i} = \mathcal{N}(y_i, \sigma^2_{y_i})\), where \(y_i\) as defined above and \(\sigma_{y_i} = c \cdot f^\sigma(x)_{z^c_i}\) for some \(c\).

Considering again the transformation in \cref{eq:transform-z} we note that distances on the space around \(f^\mu(x)_{z^c_i}\) are locally transformed by the functional determinant of the transformation, i.e. in this setting simply the derivative of the cumulative density function.
By also multiplying the scaling factor \((b_i - a_i)\) of then second transformation we can therefore compute
\begin{equation}
\sigma_{y_i} = f^\sigma(x)_{z^c_i}\cdot \mathit{cdf}'\left(\mathcal{N}(0, 1), f^\mu(x)_{z^c_i}\right) \cdot (b_i - a_i)
\end{equation}
and define
\begin{equation}
\mathit{head}: D_{z^c_i} \mapsto \mathcal{N}(y_i, \sigma_{y_i})
\end{equation}
with \(y_i\) and \(\sigma_{y_i}\) as above.

\section{Using diverse ensembles for prediction and OOD detection}
\label{sec:org14b58b1}
Finally, we show how to use the uncertainty estimates \(D_{y_i}\) of the predictions to detect OOD examples in \cref{subsec:OOD-detection} using an ensemble of artificially diversified encoders.
Furthermore, we show how a reasonable threshold \(\tau_{\rm OOD}\) can be determined without tuning on the parameters.

\subsection{Training and predicting with an ensemble}
\label{subsubsec:training-and-predicting}
To construct the ensemble we define a set of \(E\) encoders \(f_j(x)\) with \(j \in 1..E\) and train them simultaneously, together with a single generator \(g(z)\).
A typical value for \(E\) is five.
During training, for all \((x, x')\) we can compute the loss defined in \cref{eq:double-elbo} for each \(f_j\) paired with \(g\) simultaneously and sum them up before computing the gradient.
We note that, depending on the specifics of the model \(g\), the learning rate of \(g\) might have to be adjusted accordingly.

In order to use the ensemble to make a prediction, we propose a very simple method: Instead of combining the outputs of all \(f_j(x)\) in a complicated manner \autocite{lakshminarayananSimpleScalablePredictive2017}, we instead propose for each dimension \(j\) to select the single output that has the median distribution as measured by the distribution mean.
Mathematically we write
\begin{gather}
f_{1..E}(x)_i = f_j(x)_i \\ \text{ with } f^\mu_j(x)_i = \mathit{median}\left( f^\mu_j(x)_i, \dots, f^\mu_E(x)_i \right).
\end{gather}
The intuition is that since we know that the content variables are disentangled, i.e. their distributions are independent, we can predict each content variable with a different ensemble member, which we would not be able to do if the content variables had some dependence.

\subsection{Artificially diversifying the ensembles}
\label{subsec:artificially-diversified-ensembles}
Following the discussion in \cref{subsec:OOD-detection} we now propose a simple way to generate an ensemble that aims to show diversity for OOD inputs.
Although ensembles simply constructed by training the same model with different initial parameters has empirically be shown to produce diverse results in the OOD case \autocite{lakshminarayananSimpleScalablePredictive2017}, we argue that an inherent difference in architecture represents a more principled way to construct the ensemble.

To this end we propose for each ensemble member to replace the activation functions in a single layer with a different activation function, and to select a different layer for each ensemble member.
For instance, if each encoder in the ensemble has eight layers which use \emph{relu} activations, then we select layers \(3\) to \(7\) (i.e. not the last layer) and replace the activations for instance with a \emph{tanh} activation function.
Then, the ensemble gets trained as usual.
In the next section we will see how we can use this diversity to detect OOD inputs.

\subsection{Combining the ensemble to detect OOD inputs}
\label{sec:org6cd236f}
In this final section we introduce a simple algorithm to detect OOD inputs that only relies on a single threshold \(\tau_{\rm OOD}\) that we can select before evaluating any data.

We start by using the encoder ensemble \(f_{1..E}(x)\) to predict a set of content representation distributions \(D_{z^c}\).
For each index \(i\) in \(z^c\) we use the model head to construct a set of output distributions for \(y_i\), all of which are defined as a normal distribution.
Then, we combine these distributions into a Gaussian mixture model which we denote \(\mathit{gmm}_{y_i}\).
Having selected the median distribution as the ensemble prediction \(D_{y_i}\), we compute how much \emph{probability mass} of \(\mathit{gmm}_{y_i}\) lies outside of two standard deviations of \(D_{y_i}\).
If this value is larger than \(\tau_{\rm OOD}\) for any output \(y_i\) then we consider the input as OOD.
Mathematically, for each index \(i\) we can therefore write
\begin{equation}
\mathit{OOD}_i(x) = \top \Leftrightarrow \mathit{cdf}\bigl(\mathit{gmm}_{y_i}, \underline{y}_i\bigr) + \left(1 - \mathit{cdf}\bigl(\mathit{gmm}_{y_i}, \bar{y}_i\bigr) \right) > \tau_{\rm OOD}\end{equation}
where \(\underline{y}_i\) and \(\bar{y}_i\) are defined as \(y_i - 2\sigma_{y_i}\) and \(y_i + 2\sigma_{y_i}\), respectively, and
\begin{equation}
OOD(x) = \bigvee_i OOD_i(x),
\end{equation}
where \(i\) represents each index of the content variables.

The intuition behind this is that in the safety-critical setting it is crucial for the predicted probability distribution to sufficiently cover all likely outputs.
Since we do not know which ensemble member is the closest to the ``correct'' prediction we require that even in the worst case there is only a small likelihood that the realization does not fall within the prediction of the actual output.
Note that even if all ensemble members predict the exact same output, about \(5\%\) of the mass will lie outside of two standard deviations from the predicted mean.
Therefore we propose \(\tau_{\rm OOD} = 15\%\), i.e. three times higher than the perfect case, and argue that it denotes a suitable general trade-off without requiring any evaluation of the data.

\chapter{Conclusion and Discussion}
\label{sec:conclusion}
In this work we have investigated the current gap between (a) machine learning methodologies for verifying robustness of deep learning systems, and (b) the needs that arise in certification of safety-critical systems in the real-world.
To this end, we have argued that a push towards more structured or symbolic models is necessary for deep learning systems to enter safety-critical domains.
We have studied \emph{causal models} and found them unapplicable to the considered real-world use case.
Instead, we proposed using a model structure that relies on recovering \emph{disentangled variables}, which can be understood as a weaker form of causal structure.
Additionally, we argued that self- or weakly-supervised models are inherently more suitable for certification, because they exhibit more structure and are more likely to encode the fundamental mechanisms governing the data than fully-supervised models.

Next, we investigated steps towards a possible certification framework.
In particular, we examined (i) uncertainty quantification, (ii) out-of-distribution detection, (iii) feature collapse, and (iv) adversarial attacks and defenses with regard to their use in a certification framework.

We concluded that uncertainty quantification for deep learning can draw upon a rich history of research in the field of statistics, where suitable and mathematically rigorous principles have been proposed long before the rise of deep learning.
Nonetheless, despite the rich availability of literature in statistics and deep learning methodology, there is not much literature specifying precisely what properties a certified deep learning system should exhibit with regard to uncertainty quantification.
Therefore, in this work we proposed a precise set of tests that aims to fill this gap.

Regarding out-of-distribution detection, we have concluded that the term itself is a misnomer, since the definition of what is ``in'' and ``out'' of a distribution is unclear, and the ``distribution'' itself is usually not well defined.
We have therefore proposed a new practical definition for the term ``out-of-distribution'' and have subsequently derived a set of desiderata and empirical tests that can be used to validate a model.
We hope that in the upcoming years the research community will establish a consensus for the task of ``out-of-distribution detection'', or perhaps specify a more precisely defined task.

Concerning adversarial attacks, we have noted that they only pose a risk in the presence of an adversary, and do not need to be considered when only natural perturbations are present.
If an adversary is present, we have seen that they pose a much greater risk if they have access to either the model or the training data.
We therefore suggest preventing potential adversarial attacks by simply prohibiting access to the model internals and the used data.
However, if this is not possible, we do consider adversarial attacks a substantial risk in safety-critical applications.

Finally, we have proposed a novel model structure that can make regression predictions of disentangled variables, predict uncertainty and detect out-of-distribution inputs, while being trained in a weakly-supervised way.
We claim that it exhibits important principles for inherently safe design and marks an example towards certifiable deep learning models.
However, in this work we have not yet established empirical evidence of the models efficacy, either as measured directly by the model's regression performance, or as evaluated by the tests developed in this work.
In the future we therefore look forward to evaluating the proposed model on different problem domains in order to verify it's role as an example towards certifiable deep learning.

\section{Future outlook for deep learning certification}
\label{sec:org70a81e7}
In the upcoming years, we are looking forward to the machine learning community making further progress in the direction of structured and symbolic models that can be used to certifiably represent structured knowledge about the data.
In particular, we are looking forward to progress made in the field of \emph{casuality in computer vision} and other work in causal discovery.

On the regulatory side, we will closely follow current progress made by various government agencies, including the FAA, EASA, and FDA, and also expect to see progress by space agencies like NASA or ESA.

In parallel to the more ``mechanical'' certification requirements arising in robotics applications, we are also looking forward to safety and certification regulation concerning \emph{Ethical AI}, \emph{Fairness}, and \emph{Information Security \& Privacy}.
As deep learning systems enter an increasing number of fields, including policy making, economic planning, and citizen governance, we expect these fields to become of crucial importance in the coming years, and suggest the expansion of certification frameworks to these three fields.

To this end, we welcome the current and future progress made by the European Union, proposing regulatory guidance\footnote{\url{https://digital-strategy.ec.europa.eu/en/policies/european-approach\%2Dartificial-intelligence}} for AI ethics, AI liability, and Human-centric AI.
Similarly, multiple universities have recently established research programs for Human-centered AI, and we are looking forward to seeing guidance and regulation that ensures a socially just integration of deep learning systems into our world.

\newpage
\addcontentsline{toc}{chapter}{Bibliography}
\printbibliography

\appendix
\chapter{Appendix}
\label{sec:orgd2b08d9}
\section{A  primer to the Julia language}
\label{sec:julia-crash-course}
In \cref{alg:julia-crash-course} we briefly present non-trivial language features of the Julia programming language, and assume a familiarity with the Python programming language, as well as the use of scientific and machine learning libraries like \texttt{numpy/scipy} and \texttt{tensorflow/} \texttt{pytorch/jax}.
\begin{listing}[htbp]
\caption{\label{alg:julia-crash-course}A crash-course for some non-trivial language features of the Julia programming language.}
\begin{Code}
\begin{Verbatim}
\color{EFD}\EFcd{\# }\EFc{We use these libraries for notation and ML functionality.}
\EFk{using} InvertedIndices, IntervalSets, Distributions, Flux
\EFcd{\# }\EFc{We use bold letters or plurals to denote vectors.}
xs = [\EFhn{1}, \EFhn{2}, \EFhn{3}, \EFhn{4}, \EFhn{5}]; preds = [pi, \EFhn{2}pi]
\EFcd{\# }\EFc{Julia is 1-indexed.}
\textcolor[HTML]{483d8b}{\textbf{@assert}} xs[\EFhn{1}] \EFt{==} \EFhn{1}
\EFcd{\# }\EFc{Appending a dot to a function name enables auto-vectorization.}
\EFf{f}(x) = \EFhn{2}\EFt{*}x
\textcolor[HTML]{483d8b}{\textbf{@assert}} f.(xs) \EFt{==} [\EFhn{2}, \EFhn{4}, \EFhn{6}, \EFhn{8}, \EFhn{10}]
\EFcd{\# }\EFc{We can use `Flux.jl` for machine learning.}
\EFk{using} Flux: Dense, relu
\EFcd{\# }\EFc{We store the data as a vector of features, instead of a tensor.}
Input\_t = Vector\{\EFt{<:Real}\}
\EFf{model}(x \EFt{::} \EFt{Input\_t}) = Dense(\EFhn{5}\EFt{=>}\EFhn{1}, relu)
samples = [rand(\EFhn{5}) \EFk{for} \_ \EFk{in} \EFhn{1}:\EFhn{100}]
model.(samples)
\EFcd{\# }\EFc{Unfortunately, this makes getting a specific feature}
\EFcd{\# }\EFc{of every sample a bit verbose.}
first\_features = getindex.(samples, \EFhn{1})  \EFcd{\# }\EFc{not samples[:, 1]}
\textcolor[HTML]{483d8b}{\textbf{@assert}} size(first\_features) \EFt{==} \EFhn{100}
\end{Verbatim}
\end{Code}
\end{listing}

\section{The disentanglement loss}
\label{subsec:impl-details-loss}
For completeness we provide a sample implementation of the disentanglement loss used in \cref{sec:proposed-model-and-dataset}.
We note that for numerical stability, the encoder network \(f\) computes the mean and the log-variance, such that we can recover a strictly positive standard deviation by computing \(\sigma = \exp(\log\!\mathit{variance} / 2)\).

\begin{listing}[h]
\caption{\label{alg:disentanglement-loss}Disentanglement loss, originally proposed by \textcite{locatelloWeaklySupervisedDisentanglementCompromises2020}.}
\begin{Code}
\begin{Verbatim}
\color{EFD}\EFcd{\# }\EFc{k, l = content and style dimensions}
\EFk{function} \EFf{disentanglement\_loss}((xs\_lhs, xs\_rhs) \EFt{::} \EFt{Tuple}\{Vector\{Input\_t\},
                                                      Vector\{Input\_t\}\},
                              S \EFt{::} \EFt{Vector}\{Int\},
                              ) \EFt{::} \EFt{Real}
    mus\_lhs, logvars\_lhs = model.encoder.(xs\_lhs)
    mus\_rhs, logvars\_rhs = model.encoder.(xs\_rhs)

    masks = [ones(k\EFt{+}l) \EFk{for} \_ \EFk{in} \EFhn{1}:length(S)]
    \EFk{for} (mask, i) \EFk{in} zip(masks, S) mask[i] \EFk{=} \EFhn{0.} \EFk{end}

    mus\_hat\_lhs = masks \EFt{.*} (mu\_lhs .\EFt{+} mu\_rhs)\EFt{/}\EFhn{2} .\EFt{+} (\EFhn{1} .\EFt{-} masks) \EFt{.*} mu\_lhs
    mus\_hat\_rhs = masks \EFt{.*} (mu\_lhs .\EFt{+} mu\_rhs)\EFt{/}\EFhn{2} .\EFt{+} (\EFhn{1} .\EFt{-} masks) \EFt{.*} mu\_rhs

    sigs\_hat\_lhs = masks \EFt{.*} (exp.(logvars\_lhs\EFt{/}\EFhn{2}) .\EFt{+} exp.(logvars\_rhs\EFt{/}\EFhn{2}))\EFt{/} \EFhn{2} \EFt{+}
                        (\EFhn{1} .\EFt{-} masks) \EFt{.*} exp.(logvars\_lhs\EFt{/}\EFhn{2})
    sigs\_hat\_rhs = masks \EFt{.*} (exp.(logvars\_lhs\EFt{/}\EFhn{2}) .\EFt{+} exp.(logvars\_rhs\EFt{/}\EFhn{2}))\EFt{/} \EFhn{2} \EFt{+}
                        (\EFhn{1} .\EFt{-} masks) \EFt{.*} exp.(logvars\_rhs\EFt{/}\EFhn{2})

    noise\_lhs, noise\_rhs = ([rand(Normal(), k\EFt{+}l) \EFk{for} \_ \EFk{in} \EFhn{1}:length(S)],
                            [rand(Normal(), k\EFt{+}l) \EFk{for} \_ \EFk{in} \EFhn{1}:length(S)])
    zs\_lhs = noise\_lhs \EFt{.*} sigs\_lhs .\EFt{+} mus\_hat\_lhs
    zs\_rhs = noise\_rhs \EFt{.*} sigs\_rhs .\EFt{+} mus\_hat\_rhs

    xs\_rec\_lhs = model.decoder(zs\_lhs)
    xs\_rec\_rhs = model.decoder(zs\_rhs)

    \EFk{return} mean(   bernoulli\_loss.(xs\_lhs, xs\_rec\_lhs)
                 \EFt{+} bernoulli\_loss.(xs\_rhs, xs\_rec\_rhs)
                 \EFt{+} gaussian\_kl\_divergence.(mus\_tilde\_lhs, sigs\_hat\_lhs)
                 \EFt{+} gaussian\_kl\_divergence.(mus\_tilde\_rhs, sigs\_hat\_rhs))
\EFk{end}
\EFk{function} \EFf{gaussian\_kl\_divergence}(mus, sigs; lambda=\EFhn{1.})
    \EFhn{0.5} \EFt{*} sum( \EFhn{1} \EFt{./}lambda \EFt{.*} (mus\EFt{.\char94{}}\EFhn{2} .\EFt{+} sigs\EFt{.\char94{}}\EFhn{2}) .\EFt{-} \EFhn{2}\EFt{*}log.(sigs) .\EFt{-} \EFhn{1} .\EFt{+} log.(lambda); dims=\EFhn{1})
\EFk{end}
\EFk{function} \EFf{bernoulli\_loss}(x, x\_rec)
    Flux.Losses.logitbinarycrossentropy(x\_rec, x; agg=x\EFt{->}sum(x; dims=[\EFhn{1},\EFhn{2},\EFhn{3}]))
\EFk{end}
\end{Verbatim}
\end{Code}
\end{listing}
\section{A illustrated example of calibration and sharpness}
\label{sec:calibration-illustration}
\begin{figure}[H]
\centering
\includegraphics[width=\linewidth]{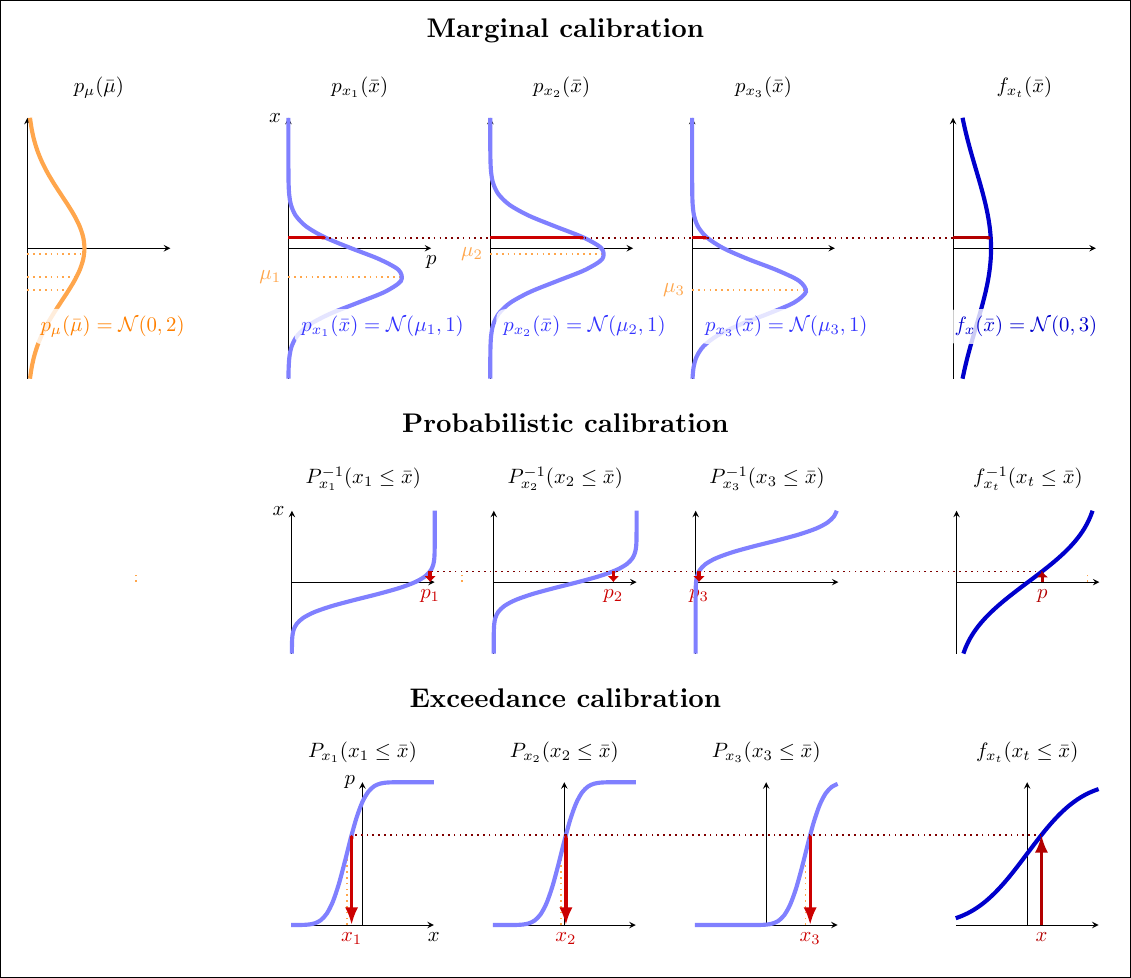}
% \includestandalone[width=\linewidth]{tikz-pictures/probabilistic-calibration/all-calibration}
\caption{\label{fig:marginal-calibration}An illustration of a slightly modified example from \textcite{gneitingProbabilisticForecastsCalibration2007} illustrating a forecast which is probabilistically and marginally calibrated, but not exeedance calibrated. The forecast is constructed as follows: For each \(t\), \(\mu_t\) is sampled from \(\mathcal{N}(0, 2)\) and \(G_t = \mathcal{N}(\mu_t, 1)\). The prediction forecast is always \(F_t(x) = \mathcal{N}(0, 3)\), i.e. it does not depend on the time step.}
\end{figure}
\section{Determining a minimum number of samples for conditional calibration.}
\label{sec:samples-for-conditional-calibration}
In the algorithm in \cref{alg:test-calibration-curve} we include a margin of error \(\epsilon\) such that the observed frequency must be greater than \(p \cdot (1 - \epsilon)\).
We address this question through a brief computational study:
We assume observations \(\mathcal{N}(0, 1)\) and perfect predictions \(\mathcal{N}(0, 1)\).
Then we randomly sample \(N\) observations and evaluate the test, which either passes or fails.
We repeat this for \(T\) trials and compute the frequency of test failures.
We compute this empirical frequency for different values of \(N\).
Finally we pick a value of \(N\) that has sufficiently low frequency of failure.
For example, with \(\epsilon=10\%\) and \(N=10'000\) we measure a false failure rate of approximately \(1\%\).
The algorithm is presented in \cref{alg:comp-study-failure-prob}.

\begin{listing}[htbp]
\caption{\label{alg:comp-study-failure-prob}Computational study on expected failure probability for conditional calibration.}
\begin{Code}
\begin{Verbatim}
\color{EFD}\EFk{function} \EFf{run\_trial}(n\_samples, eps)
    obs = randn(n\_samples);
    preds = [Normal(\EFhn{0}, \EFhn{1}\EFt{\char94{}}\EFhn{2}) \EFk{for} \_ \EFk{in} obs];
    trial\_failed = \EFo{false}
    \EFk{try}
        test\_calibration\_curve(preds, obs; eps=eps)
    \EFk{catch} AssertionError
        trial\_failed = \EFo{true}
    \EFk{end}
    \EFk{return} trial\_failed
\EFk{end}

n\_samples = \EFhn{10} \EFt{.\char94{}} (\EFhn{1}:\EFhn{5})
\EFk{function} \EFf{compute\_failure\_prob}(n\_samples, eps)
     n\_trials = \EFhn{100}
     \EFcd{\# }\EFc{Run trials in parallel with tcollect if `julia --threads=k`.}
     trials = (run\_trial(n\_samples, eps) \EFk{for} \_ \EFk{in} \EFhn{1}:n\_trials) \EFt{|>} tcollect
     mean(trials)
\EFk{end}

failure\_probs = Dict(
    [eps \EFt{=>} [compute\_failure\_prob(n\_samples, eps) \EFk{for} n\_samples \EFk{in} n\_samples]
     \EFk{for} eps \EFk{in} [\EFhn{0.05}, \EFhn{0.10}, \EFhn{0.20}]]
)
\end{Verbatim}
\end{Code}
\end{listing}

\begin{table}[htbp]
\label{tab:failure-probs}
\centering
\begin{tabular}{r|rrrrr|}
 & n = 10 & 100 & 1'000 & 10'000 & 100'000\\
\hline
\(\epsilon=5\%\) & 0.92 & 0.82 & 0.55 & 0.13 & 0.0\\
10\% & 0.88 & 0.64 & 0.31 & 0.02 & 0.0\\
20\% & 0.8 & 0.44 & 0.06 & 0.0 & 0.0\\
\end{tabular}
\end{table}

\section{Bi-Lipschitz constraints for different layers}
\label{subsec:bilipschitz-for-layers}
\subsection{On the bi-Lipschitz constraint for common activation functions}
\label{sec:org6fb015a}
Many activation functions have a useful upper Lipschitz constant \(\overline{C}\) (maximum absolute slope), usually with value 1 (relu, tanh) or lower (sigmoid).
If used in the residual function (i.e. \(f(x) = x + \sigma(Wx))\) this is sufficient, and leads to proper upper and lower Lipschitz constants for residual layers.
In dense layers however, the lower Lipschitz constant collapses to \(0\) for most activation functions, i.e. there are two distinct points \(x\) and \(x'\) such that \(|f(x) - f(x')| \to 0\).
Therefore, for activations after dense layers we suggest using the ``leaky relu'' activation function, i.e.
\begin{equation}
\sigma(x) = \begin{cases} \alpha x &\text{ if } x < 0 \\ \phantom{\alpha}x &\text{ if } x > 0 \end{cases}
\end{equation}
where \(\alpha \in (0, 1)\) is typically chosen to be \(0.01\) or \(0.02\), and the Lipschitz constants are \(\underline{C} = \alpha\) and \(\overline{C} = 1\).
In the case of feature collapse, larger values for \(\alpha\) may be useful as they have a larger Lipschitz constant.

\subsection{On the proof of bi-Lipschitz for residual networks}
\label{sec:org66e43e2}
Consider a model \(\mathit{model}(x) = (f_1\fatsemi f_2 \fatsemi \hdots \fatsemi f_L)(x)\) consisting purely of residual layers \(f(x) = x + f'(x)\) (we drop the index of \(f\) for convenience).
Assume for now that each \(f'(x)\) is \(\alpha\text{-Lipschitz}\), i.e. \(\|x-x'\|_X \leq \alpha \|f'(x) - f'(x')\|_F\) for some norms \(X\) and \(F\).
Then we can show
\begin{equation}
(1-\alpha) \|x - x'\|_X \leq \|f(x) - f(x') \|_F \leq (1+\alpha) \|x - x'\|_X
\end{equation}
and consequentially
\begin{equation}
(1-\alpha)^L \|x - x'\|_X \leq \| \mathit{model}(x) - \mathit{model}(x') \|_F \leq (1+\alpha)^{L} \|x - x'\|_X.
\end{equation}

First, to prove \(\| f(x) - f(x') \| \leq (1+\alpha) \|x-x'\|\) we proceed as follows:
\begin{equation}
\begin{aligned}
\|f(x) - f(x')\| &= \|x-x'+f'(x)-f'(x')\| \\
&\leq \|x-x'\|+\|f'(x)-f'(x')\|\\
&\leq \|x-x'\|+\alpha\|x-x'\|\\
&=(1+\alpha)\|x-x'\|
\end{aligned}
\end{equation}
Then, to prove \((1-\alpha)\|x-x'\| \leq \|f(x) - f(x')\|\) we proceed as follows:
\begin{equation}\begin{aligned}
\|x-x'\| &= \|(x-x') + (x+f'(x)) - (x+f'(x)) + (x'+f'(x')) - (x'+f'(x'))\|\\
&\leq \|(x-x')-(x+f'(x))+(x'+f'(x'))\|+\|f(x)-f(x')\|\\
&=\|f'(x)+f'(x')\|+\|f(x)-f(x')\|\\
&\leq \alpha\|x-x'\|+\|f(x)-f(x')\|
\end{aligned}\end{equation}
and we can conclude \(\Rightarrow (1-\alpha)\|x-x'\| \leq \|f(x)-f(x')\|\).

%\section{Declaration of originality}
%\label{sec:org8bc13ea}
%See final page.
% \includepdf{./figs/declaration-originality-1.pdf}
\end{document}